\documentclass[preprint,11pt,numbers,sort&compress]{elsarticle}

\usepackage[a4paper,margin=25mm]{geometry}
\usepackage{fancyhdr}
\usepackage{lastpage}

\fancypagestyle{fancyplain}{
	\fancyhf{} 
	\fancyhead[C]{\small Preprint – November 2025 – Rojahn and Grum – Green AI: A systematic review and meta-analysis of its definitions, lifecycle models, hardware and measurement attempts}
	\fancyfoot[C]{Page \thepage\ of \pageref{LastPage}}
}

\fancypagestyle{plain}{
	\fancyhf{}
	\fancyhead[C]{\small Preprint – November 2025 – Rojahn and Grum – Green AI: A systematic review and meta-analysis of its definitions, lifecycle models, hardware and measurement attempts}
	\fancyfoot[C]{Page \thepage\ of \pageref{LastPage}}

}

\makeatletter
\def\ps@pprintTitle{%
	\let\@oddhead\@empty
	\let\@evenhead\@empty
	\let\@oddfoot\@empty
	\let\@evenfoot\@oddfoot}
\makeatother

\usepackage{times} 
\usepackage{natbib} 
\usepackage{amsmath} 
\usepackage{hyperref} 
\usepackage{graphicx} 
\usepackage{longtable} 
\usepackage{booktabs} 
\usepackage{rotating} 
\usepackage{array} 
\usepackage{pdflscape} 
\usepackage{amssymb} 
\usepackage{multirow} 
\usepackage{enumitem} 
\usepackage{makecell} 
\usepackage{placeins} 
\usepackage{caption} 
\usepackage{subcaption} 
\usepackage{siunitx} 
\usepackage{tikz} 
\usetikzlibrary{positioning,arrows.meta} 
\usepackage{comment} 
\PassOptionsToPackage{table}{xcolor} 
\usepackage{xcolor} 
\usepackage{colortbl} 
\usepackage{array} 
\usepackage{ragged2e} 
\usepackage{csvsimple} 
\DeclareUnicodeCharacter{0301}{} 
\newcolumntype{L}[1]{>{\RaggedRight\arraybackslash}p{#1}} 

\usepackage{etoolbox}
\makeatletter
\patchcmd{\@author}
{}
{}
{}{}
\patchcmd{\@@author}
{}
{}
{}{}
\makeatother

\makeatletter
\def\ps@pprintTitle{%
	\let\@oddhead\@empty
	\let\@evenhead\@empty
	\def\@oddfoot{\hfill Preprint – November 2025\hfill}%
	\let\@evenfoot\@oddfoot}
\makeatother

\begin{document}

\title{Green AI: A systematic review and meta-analysis of its definitions, lifecycle models, hardware and measurement attempts}


\author[uni]{Marcel Rojahn \corref{cor1}}
\ead{marcel.rojahn@uni-potsdam.de}
\cortext[cor1]{Corresponding author. 
	ORCID: \href{https://orcid.org/0000-0002-2231-5690}{0000-0002-2231-5690}}

\author[uni]{Marcus Grum}
\ead{marcus.grum@uni-potsdam.de}

\address[uni]{University of Potsdam, Junior Chair of Business Information Systems, esp. AI-based Application Systems\\
	Karl-Marx-Str.~67, Potsdam 14482, Germany}

\date{November 11, 2025}


\begin{abstract}
	
	{\bf Background:}
	Across the Artificial Intelligence (AI) lifecycle - from hardware to development, deployment, and reuse - burdens span energy, carbon, water, and embodied impacts. Cloud-provider tools improve transparency but remains heterogeneous and often omit water and value-chain effects, limiting comparability and reproducibility. Addressing these multi-dimensional burdens requires a lifecycle approach linking phase-explicit mapping with system levers (hardware, placement, energy mix, cooling, scheduling) and calibrated measurement across facility, system, device, and workload levels.
	
	{\bf Objectives:}
	This article (i)~establishes a unified, operational definition of Green AI distinct from Sustainable AI; (ii)~formalizes a five-phase lifecycle mapped to Life Cycle Assessment (LCA) stages, making energy, carbon, water, and embodied impacts first-class; (iii)~specifies governance via Plan-Do-Check-Act (PDCA) cycles with decision gateways; (iv)~systematizes hardware and system-level strategies across the edge-cloud continuum to reduce embodied burdens; and (v)~defines a calibrated measurement framework combining estimator models with direct metering to enable reproducible, provider-agnostic comparisons.

	{\bf Methods:} 
	This review synthesizes 103 studies through a meta-analytical framework operationalizing 164 terms across the Green AI lifecycle. Using taxonomy-guided annotation, each source was situated along three analytical axes: lifecycle phase (e.g., design, training, end-of-life), impact vector (energy, water, material intensity), and system layer (hardware-software boundary).	Concepts were mapped to standardized LCA stages and consolidated into a task-level process model. Mechanisms from High-Performance Computing~(HPC) management were examined for their transferability to AI. Finally, measurement approaches were contrasting estimator-based tools with direct metering to evaluate calibration, reporting completeness.
	
	{\bf Results:}
	This article delivers four main contributions: (1)~a unified, operational definition of Green AI extending beyond training efficiency to energy, carbon, water, and embodied material footprints with standardized and transparent reporting; (2)~a comprehensive five-phase lifecycle mapped to LCA stages and structured by PDCA with decision gateways; (3)~hardware and system-level strategies across the edge-cloud continuum, informed by rigorously transferable HPC techniques; and (4)~a hybrid measurement architecture combining estimators with direct metering for reliable calibration and standardized reporting. Remaining gaps include inconsistent metrics, limited attention to water and Scope 3 impacts, provider-specific dashboards that hinder comparability, and a lack of reproducible, calibrated measurements across hardware tiers.
	
	{\bf Conclusions:}
	Combining definition, lifecycle processes, hardware strategies, and calibrated measurement, this article offers actionable, evidence-based guidance for researchers, practitioners, and policymakers. The framework improves green model/infrastructure choices, context-aware scheduling, transparent reporting, and resilient long-term circularity.
\end{abstract}

\maketitle

\newpage

\section{Introduction}

Artificial Intelligence (AI) accelerators, such as Graphics Processing Units (GPUs) and domain-specific Tensor/Neural Processing Units (TPUs/NPUs), have become the default compute substrate for training transformer-based and large-scale multimodal models \cite{achiam_gpt-4_2024,team_gemini_2024}. Realizing these capabilities has entailed an impressive scale-up in model size and training, from millions to billions of parameters in publicly reported models, together with operations such as matrix multiplications and attention mechanisms \cite{brown_language_2020,vaswani_attention_2017}. As a result, training deep neural networks and large language models requires several orders larger energy demand than conventional machine-learning approaches, with the factor driven by model size and training duration \cite{patterson_carbon_2021,strubell_energy_2019}. Beyond operational energy and carbon emissions, recent syntheses emphasize additional environmental burdens - including water consumption and embodied impacts across the hardware supply chain - that remain underrepresented in model-centric accounts \cite{bolon-canedo_review_2024,verdecchia_systematic_2023,li_making_2023,gupta_act_2022}. This escalation makes Green AI a first-class design maxim for modern AI systems.

At the same time, cloud providers expose emissions tooling (e.g., dashboards for location- and market-based accounting), signaling a shift toward operational transparency. Methodologies, however, differ across providers, often lack water and Scope~3 (value-chain) coverage, and impede cross-system comparability \cite{googlecloud_carbon_2024,microsoft_plan_2024,aws_machine_2024}. At the application level, end-to-end AI systems - combining perception, language, and decision components - can increase energy demand at deployment when subjected to large request volumes, long context windows, strict latency service-level objectives, and heterogeneous edge-cloud placement constraints \cite{luccioni_estimating_2023}.

Despite notable progress, the field lacks an AI-specific lifecycle-wide and calibrated standard that unifies energy, carbon, \emph{and} water measurement and reporting, integrates embodied impacts, and enables reproducible comparison across hardware tiers and deployment contexts. Existing efforts frequently optimize isolated segments - data centers, algorithms, or runtime kernels - without a consistent linkage across the AI lifecycle \cite{luccioni_estimating_2023,schwartz_green_2020}.

On the other side, accurate assessment of Green AI requires coordinated, multi-level measurement: facility-level metrics (e.g., Power Usage Effectiveness, PUE; and Water Usage Effectiveness, WUE), system consumption, hardware telemetry (e.g., Running Average Power Limit, RAPL; and Performance Monitoring Counters, PMCs), workload profiling (step/batch), and lifecycle-complete coverage (e.g., hardware, development, and end-of-life) \cite{au_identification_2017,luccioni_estimating_2023}. Indirect estimators must be calibrated against direct measurements to ensure valid reporting.

\textbf{Contributions.}
This article advances the literature through four key contributions: 
(1) a unified definition of \emph{Green AI} concise, comprehensive, delineated from \emph{Sustainable AI}; 
(2) a \emph{Green AI lifecycle} that maps five phases and subphases to Life Cycle Assessment (LCA) stages, exposing often-overlooked Scope~3 and water-related impacts alongside energy and carbon; 
(3) a hardware- and software-aware realization and governance framework that formalizes Plan-Do-Check-Act (PDCA) loops with decision gateways based on Phase Completion Criteria (PCC) and Performance-Environmental Thresholds (PET), transferring HPC techniques into AI pipelines; 
(4) a measurement architecture that combines indirect estimation (e.g., emissions models) with direct power metering to calibrate discrepancies and standardize reporting across energy, carbon, and water.

On this basis, the following research questions are addressed: 
\begin{itemize}[topsep=0pt]
	\item \textbf{RQ1:} How should \emph{Green AI} be defined and delimited relative to adjacent concepts, and which principles govern its lifecycle-wide application?
	\item \textbf{RQ2:} Which phases and subphases constitute the \emph{Green AI lifecycle}, and how do they map to Life Cycle Assessment (LCA) stages?
	\item \textbf{RQ3:} Which hardware components and system architectures enable improved performance per watt and reduced embodied impacts across the lifecycle?
	\item \textbf{RQ4:} How can environmental impacts (e.g. energy, carbon, water) be measured and reported consistently across the lifecycle, including calibration of indirect estimators with direct measurements?
	\item \textbf{RQ5:} Which open challenges and research priorities (benchmarks, reporting schemas, uncertainty handling, circularity, governance) are needed to advance lifecycle integration of Green AI?
\end{itemize}

Section~\ref{sec:Foundations} surveys the core Green AI literature and adjacent economic and social perspectives. Section~\ref{sec:MethodologicalApproachtoResearch} details the methodology, including taxonomy-driven coding and trend analysis. Section~\ref{sec:LiteratureAnalysis} reports findings addressing RQ1-RQ4. Section~\ref{sec:ChallengesInImplementingGreenAIAndFutureResearchDirections} synthesizes major challenges and outlines the research agenda (RQ5). Section~\ref{sec:Discussion} synthesizes implications for model and infrastructure selection, carbon- and context-aware scheduling, lifecycle governance, and PDCA-based decision gateways; it also delineates limitations and threats to validity, including systematic literature review (SLR) coverage bias, uncertainty from estimator-meter discrepancies, provider-specific dashboards that hinder comparability, and incomplete treatment of water and Scope~3 impacts. Section~\ref{sec:Conclusion} consolidates the contributions into concrete guidance and a prioritized agenda covering standardized benchmarks and reporting, calibrated measurement procedures, uncertainty disclosure, and circularity at end-of-life.

\section{Foundations and prior research in Green AI}
\label{sec:Foundations}

Green artificial intelligence (AI) is concerned with maintaining task performance while minimizing environmental impacts across the full lifecycle, including operational energy use and carbon emissions, cooling-related water use, and material (embodied) footprints in hardware supply chains \cite{schwartz_green_2020,li_making_2023,gupta_act_2022}. Early evidence on the training-related footprint of large models \citeyearpar{strubell_energy_2019} motivated subsequent studies along complementary levers: algorithmic efficiency, hardware and placement choices, compiler/runtime optimizations, and data and storage management - each targeting energy and carbon reduction under explicit quality and latency constraints. Within the algorithmic family, pruning and quantization, together with related compression methods (e.g., knowledge distillation, low-rank adaptation, and sparsity), reduce compute and energy demand while controlling accuracy loss \cite{zhou_opportunities_2023}.

\textbf{Operationalizing Green AI.}
Operationalization spans multiple system layers. On the hardware side, energy-efficient accelerators and context-aware placement across the edge-cloud continuum enable improved performance per watt under regional energy mixes and cooling technologies, including cooling-water intensity \cite{bolon-canedo_review_2024,dodge_measuring_2022}. At the runtime level, dynamic voltage and frequency scaling, power capping, queue-level energy budgeting, and locality-aware scheduling translate established high-performance computing (HPC) practices into modern AI pipelines. On the data and storage side, de-duplication, tiered storage, retention policies, and automated data-quality gates curb unnecessary data movement and redundant compute \cite{calero_green_2015,calero_introduction_2021}. Frameworks from green software engineering (e.g., GAISSA; \cite{martinez-fernandez_towards_2023}) articulate concrete development and inference practices that can be transferred to AI-centric systems, while noting the additional hardware- and model-induced impacts specific to AI deployments.

\textbf{Holistic perspectives and lifecycle alignment.}
Green-ability models propose quality attributes and metrics - such as energy consumption, resource usage, and system longevity - that support architecture-level Green AI assessments \cite{calero_green_2015,calero_introduction_2021}. Complementary studies argue for end-to-end integration of energy-aware practices across lifecycle stages \cite{haakman_ai_2021} and for aligning AI processes with standardized LCA concepts, ensuring that upstream (raw-material extraction, component manufacturing, transport) and downstream (inference, maintenance, reuse) effects are captured alongside model-centric impact metrics.

\textbf{Sustainability dimensions and access.}
Sustainable AI entails environmental, economic, and social dimensions \cite{van_wynsberghe_sustainable_2021}. Incorporating efficiency and footprint metrics into research evaluation can realign incentives toward lower-impact solutions. Lower resource requirements can broaden participation by reducing entry costs for under-resourced institutions and enabling more equitable access to experimentation \cite{schwartz_green_2020,dodge_measuring_2022}.

\textbf{Positioning.}
Prior studies deliver important building blocks but remains fragmented: some strands target algorithms or hardware in isolation, others focus on software practices or socio-economic aspects. What is still missing is an integrated, lifecycle-explicit synthesis that (i)~clearly distinguishes \emph{Green AI} from broader \emph{Sustainable AI}, (ii)~maps AI phases and subphases to standardized LCA stages, (iii)~formalizes governance via Plan-Do-Check-Act (PDCA) loops with decision gateways based on phase completion criteria (PCC) and Performance-Environmental Thresholds (PET), and (iv)~couples estimator-based reporting with calibrated direct measurements to consistently cover energy, carbon, and water. The remainder of this article addresses these gaps.

\section{Methodological approach to this research}
\label{sec:MethodologicalApproachtoResearch}

\begin{figure}[ht]
	\centering
	\includegraphics[width=\textwidth]{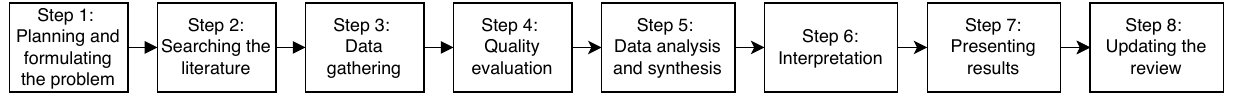}
	\caption{Eight-step process used in this systematic literature review (SLR) (adapted from \cite{thome_conducting_2016}).}
	
	\label{fig:EightStepApproachtoConductaSLR}
\end{figure}

This systematic literature review (SLR) follows established guidance for evidence synthesis in information systems and software engineering and reports according to the Preferred Reporting Items for Systematic Reviews and Meta-Analyses (PRISMA 2020) \cite{tranfield_towards_2003,thome_conducting_2016,kitchenham_systematic_2009,page_prisma_2021}.
All raw export files are available in the replication package. Figure~\ref{fig:EightStepApproachtoConductaSLR} outlines the eight-step process adapted from Thom\'e et~al.\ \citeyear{thome_conducting_2016}; the following sections detail database coverage and search strings (Phase~1), eligibility criteria (Phase~2), screening and de-duplication procedures with inter-rater agreement reporting (Phase~3), and extraction, appraisal, and synthesis methods (Phase~4). To strengthen validity, the protocol was pre-specified, with dual coding and inter-rater agreement.

\noindent
The remainder of this section details 
(i)~the review focus and taxonomy (section~\ref{subsec:SLRReviewFocus}), 
(ii)~information sources and search strategy (section~\ref{subsec:LiteratureSearchProcess}), 
and (iii)~eligibility, selection, extraction, appraisal, formal risk-of-bias assessment, and synthesis procedures (Sections~\ref{subsubsec:Phase2}--\ref{subsubsec:Phase4}) as well as 
(iv)~yearly trends and article sources (section~\ref{subsubsec:YearlyTrendsandArticleSourcesAnalysis}).

\subsection{Systematic literature review focus}
\label{subsec:SLRReviewFocus}

The review scope is structured using the framework by vom Brocke et~al.\ \citeyearpar{vom_brocke_reconstructing_2009} and Cooper's taxonomy \citeyearpar{cooper_organizing_1988} (Table~\ref{tab:TaxonomyFrameworkoftheSLR}). Specifically: 
(1)~\textbf{Focus - research outcomes, research methods, and theories}; 
(2)~\textbf{Goal - integration} across environmental assessment, lifecycle coverage mapped to LCA stages, and benchmarking of reporting practices; 
(3)~\textbf{Perspective - neutral representation}, avoiding bias across methodological and epistemic stances; 
(4)~\textbf{Coverage - representative} across computer science, environmental science, and engineering, with targeted inclusion of top-tier AI venues (e.g., NeurIPS, ICML, ICLR via proceedings indices, with arXiv preprints flagged accordingly) and domain journals (e.g., \emph{Communications of the ACM}, \emph{Journal of Machine Learning Research}, \emph{Journal of Cleaner Production}); 
(5)~\textbf{Organization - conceptual and lifecycle-explicit}, along the Green AI lifecycle with impact metrics (energy in J or kWh; carbon footprint in kg~CO$_2$e; water use; embodied/material impacts) and 
(6)~\textbf{Audience - general scholars, practitioners, and policymakers}.

\begin{table}[ht]
	\centering
	\caption{Taxonomy framework of the SLR. Bold entries denote applied categories.}
	\includegraphics[width=\textwidth]{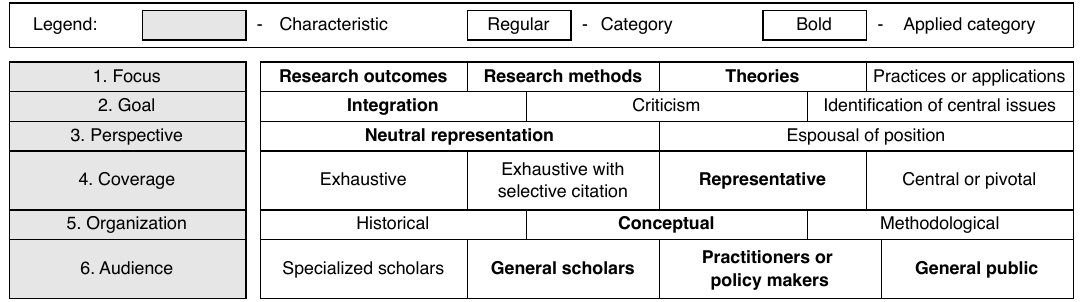}
	\label{tab:TaxonomyFrameworkoftheSLR}
\end{table}

\subsection{Literature search process}
\label{subsec:LiteratureSearchProcess}

The literature search proceeded in four phases: 
(1)~identifying databases and iteratively refining field-restricted search strings (section~\ref{subsubsec:Phase1});
(2)~specifying transparent inclusion and exclusion criteria (section~\ref{subsubsec:Phase2}); 
(3)~executing the queries and systematically screening titles/abstracts and full texts (section~\ref{subsubsec:Phase3}; PRISMA in Fig.~\ref{fig:PRISMA}); and 
(4)~consolidating the final article set and preparing the structured extraction corpus (section~\ref{subsubsec:Phase4}). 

\subsubsection{Phase 1 -- Choosing databases and refining keywords iteratively}
\label{subsubsec:Phase1}

\begin{figure}[ht]
	\centering
	\includegraphics[width=\textwidth]{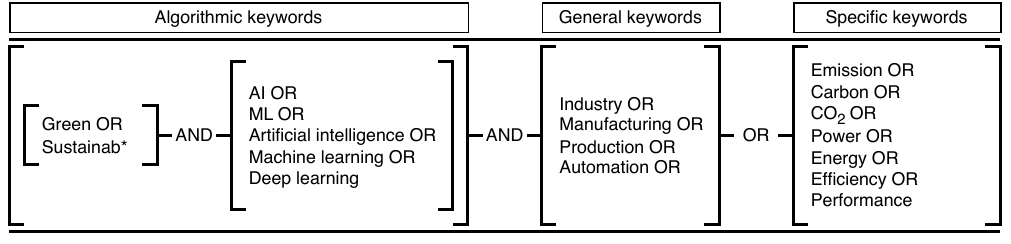}
	\caption{Core keyword groups and terms used for the database searches (domain/context, impact dimensions, AI scope).}
	\label{fig:DefinedKeywordsfortheSLR}
\end{figure}

Primary databases were \emph{Web of Science} (Topic: Title, Abstract, Author Keywords, Keywords Plus), \emph{ScienceDirect} (Title, Abstract, Author Keywords), \emph{IEEE Xplore} (journals and conferences). To capture leading AI venues not consistently indexed - specifically NeurIPS, ICML, ICLR - \emph{arXiv} was queried; preprints were flagged and handled as specified in Phase~2. Field-restricted Boolean search strings were iteratively refined across three concept groups: 
(i)~domain and context (industry, manufacturing, production, automation); 
(ii)~impact dimensions (emissions/carbon, CO$_2$, power/energy, efficiency/performance); and 
(iii)~AI scope (AI, machine learning, deep learning) (Fig.~\ref{fig:DefinedKeywordsfortheSLR}). 
During screening, lifecycle/measurement terminology (e.g., LCA; Scope~3/embodied impacts; Water Usage Effectiveness, WUE; Power Usage Effectiveness, PUE; dynamic voltage and frequency scaling, DVFS; running average power limit, RAPL; direct power metering; estimator calibration; location- vs.\ market-based emission factors; carbon-aware scheduling; edge-cloud placement; uncertainty disclosure) was systematically used for backward and forward citation chaining to capture articles on water, value-chain impacts, and calibration/uncertainty well beyond purely energy- and carbon-centric work. This targeted chaining was pre-specified in the protocol and did not alter selection decisions; instead, it served to verify comprehensive coverage across impact dimensions and to enable critical cross-checks between provider dashboards and metering-based articles.

\subsubsection{Phase 2 -- Inclusion and exclusion criteria}
\label{subsubsec:Phase2}

Consistent with standards for systematic reviews \cite{light_summing_1984,durach_new_2017}, \textit{eligible sources} were restricted to 
(i)~peer-reviewed journal articles (retrieved via \emph{Web of Science}, \emph{ScienceDirect}, \emph{IEEE Xplore}) and (ii)~peer-reviewed conference proceedings indexed in \emph{IEEE Xplore} or \emph{ACM Digital Library}, as well as proceedings series accessible via \emph{ScienceDirect} (paper-level peer review indicated in the record), 
(iii)~peer-reviewed book chapters in edited volumes with documented external review, as well as 
(iv)~scholarly monographs from academic presses with documented external peer review and an explicit methods section on energy/measurement or LCA (section~\ref{subsubsec:Phase4}). 
Preprints (e.g., from \emph{arXiv}) were flagged and used systematically for mapping, rigorous citation chaining, and transparency; where a journal version existed, it superseded the preprint.

In addition to outlet-level peer review, \textit{methodological eligibility} required an AI focus and at least one of the following: (a)~quantitative reporting of energy/power or compute intensity, or of carbon/water/embodied impacts; 
(b)~lifecycle boundary specification or mapping to LCA stages; or 
(c)~explicit measurement/telemetry or calibration procedures (e.g., RAPL; direct power metering; emission factor specification).

\textit{Exclusions} covered grey literature (e.g., theses, white papers, blogs, patents), non-English texts (for terminology consistency and database coverage), and articles prior to 2014 (onset of the modern deep-learning era and associated measurement/efficiency literature). Eligibility criteria were explicitly aligned with the research questions (RQ1-RQ5) and the review taxonomy (Table~\ref{tab:TaxonomyFrameworkoftheSLR}), thereby minimizing scope drift and selection bias. Review articles were retained for mapping and citation chaining but not treated as primary evidence unless they introduced standardized measurement assets (e.g., emission factor models with documented provenance, metering calibration protocols, or LCA boundary definitions) or unique, otherwise unavailable benchmark datasets.

\subsubsection{Phase 3 -- Executing the literature search and screening}
\label{subsubsec:Phase3}

\begin{figure}[ht]
	\centering
	\includegraphics[width=\textwidth]{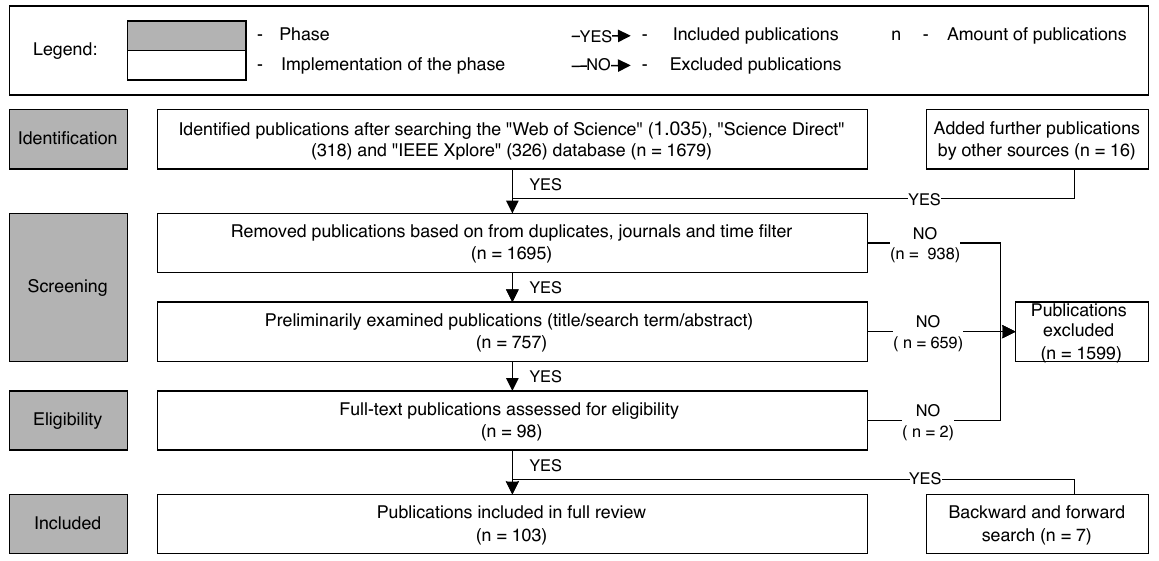}
	\caption{PRISMA 2020 flow diagram of article selection (Oct~2024-Aug~2025) (Adapted from \cite{page_prisma_2021}).}
	\label{fig:PRISMA}
\end{figure}

The literature search commenced in October 2024, with final data extraction completed in August 2025. The PRISMA flow (Fig.~\ref{fig:PRISMA}) documents records identified through database searching - \emph{Web of Science} (1{,}035), \emph{ScienceDirect} (318), and \emph{IEEE Xplore} (326) - for a total of 1{,}679 items, plus 16 records from other sources (e.g., provider dashboards and tool repositories), yielding 1{,}695 items before comprehensive eligibility screening and systematic de-duplication procedures.

After automated de-duplication that first collapsed records by digital object identifier (DOI) and otherwise by normalized title-first-author-year tuples (case-folded; diacritics and punctuation removed; fuzzy title matching using a pre-specified token-set ratio threshold; see replication package), and after applying pre-specified filters (English language; year $\geq 2014$; document type: journal or conference; preprints permitted and flagged; journal versions superseding preprints), \textbf{757} records remained for title/abstract screening by two independent assessors. In line with Thom\'e et~al.\ \citeyear{thome_conducting_2016}, titles, keywords, and abstracts were independently assessed against predefined inclusion/exclusion criteria and the research questions, thereby narrowing the dataset from \textbf{757} to \textbf{98} articles for full-text assessment. Of the \textbf{757} records screened, \textbf{659} were excluded at this stage.

At full text, \textbf{2} articles were excluded for lack of Green AI relevance or insufficient methodological transparency in measurement/reporting, leaving \textbf{96} database-sourced articles. Backward and forward citation chaining then contributed a further \textbf{7} articles, resulting in a final total of \textbf{103} included articles. 

\subsubsection{Phase 4 -- Information collection}
\label{subsubsec:Phase4}

A structured extraction schema captured: bibliographic data; lifecycle phases and mapping to LCA stages; impact dimensions (energy, carbon, water, embodied/material); measurement approach (estimator vs.\ direct metering; device- vs.\ facility-level); hardware and deployment context (central processing units, CPU; graphics processing units ,GPUs; domain-specific tensor/neural processing units, TPUs/NPUs; edge-cloud placement; region/energy mix; cooling); algorithmic/system levers (compression, DVFS, power capping, carbon-aware scheduling); reporting completeness (International System of Units, emission factor provenance, reproducibility artifacts); and auxiliary elements such as provider-specific dashboards and end-of-life treatment of hardware. Beyond these categories, the schema enforced traceability between lifecycle stages and reported indicators, with dual coding and inter-rater agreement ensuring reproducibility and minimizing subjective bias. Linkage fields mapped AI tasks to standardized LCA stages, and detailed measurement provenance (e.g., market- vs.\ location-based carbon factors, water withdrawal vs.\ actual consumption) was explicitly captured.

\subsection{Yearly trends and article sources analysis}
\label{subsubsec:YearlyTrendsandArticleSourcesAnalysis}

\begin{figure}[ht]
	\centering
	\begin{subfigure}[ht]{0.58\textwidth}
		\centering
		\includegraphics[width=\textwidth]{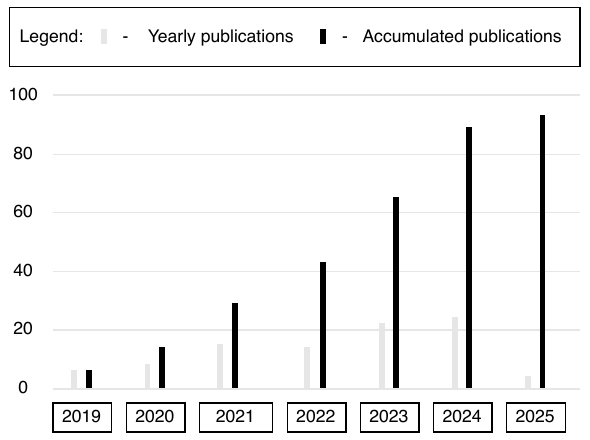}
		\caption{Annual and cumulative Green AI articles (2019--2025).}
		\label{fig:YearlyandAccumulatedGreenAIArticles}
	\end{subfigure}
	\hfill
	\begin{subfigure}[ht]{0.39\textwidth}
		\centering
		\includegraphics[width=\textwidth]{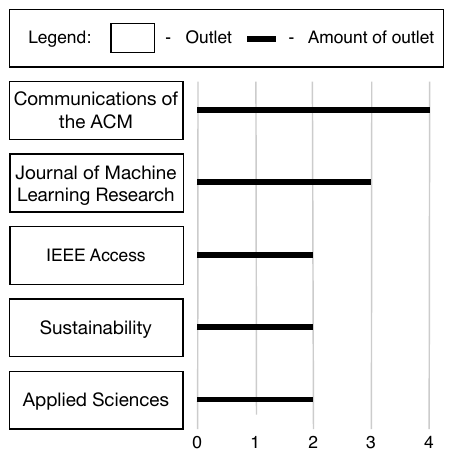}
		\caption{Number of articles per outlet (journals and conferences only).}
		\label{fig:NumberofArticlessperOutlet}
	\end{subfigure}
	\caption{Comparison of Green AI artcile trends, 2019-2025.}
	
	\label{fig:ComparisonofGreenAIArticleTrends}
\end{figure}

Fig.~\ref{fig:YearlyandAccumulatedGreenAIArticles} shows a sustained rise in Green AI articles that are explicitly energy- and measurement-focused since 2019. Based on the curated corpus (cut-off date: 7~Aug~2025), yearly counts are 6 (2019), 8 (2020), 15 (2021), 14 (2022), 22 (2023), 24 (2024), and 4 (2025), for a total of 93 items over 2019-2025. This trajectory closely coincides with rapidly growing attention to AI's environmental footprint in the literature \citep{strubell_energy_2019}. 

To distinguish consistently active venues from sporadic contributors, Fig.~\ref{fig:NumberofArticlessperOutlet} displays outlets with at least two peer-reviewed articles in 2019--2025: \emph{Communications of the ACM} (4), \emph{Journal of Machine Learning Research} (3), \emph{IEEE Access} (2), \emph{Sustainability} (2), and \emph{Applied Sciences} (2). By article type (2019--2025), journals account for 40 items (43.0\%), conference papers for 21 (22.6\%), preprints for 14 (15.1\%), web resources for 10 (10.8\%), books for 3 (3.2\%), book chapters for 2 (2.2\%), and other types for 1 (1.1\%). Item types follow \texttt{journalArticle}, \texttt{conferencePaper}, \texttt{preprint}, \texttt{webpage}, \texttt{book}, \texttt{bookSection}, and \texttt{document}. For the outlets figure, only rigorously peer-reviewed journals and established conferences are included (preprints and web resources excluded).

\section{Literature analysis}
\label{sec:LiteratureAnalysis}

\begin{figure}[ht]
	\centering
	\includegraphics[width=\textwidth]{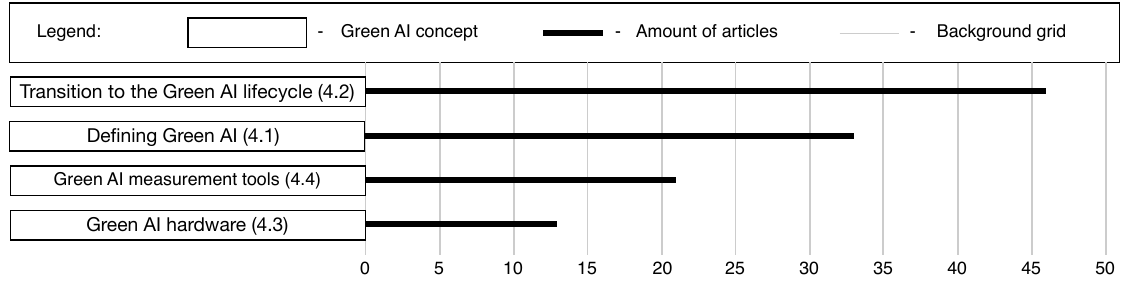}
	\caption{Distribution of articles across Green AI concepts (multi-label; base $n=103$).}
	\label{fig:DistributionofArticlesacrossGreenAIConcepts}
\end{figure}

Figure~\ref{fig:DistributionofArticlesacrossGreenAIConcepts} summarizes the thematic categories derived during the synthesis phase (iterative coding) and maps them to the sections that address the research questions (RQ1-RQ5).
The corpus is dominated by lifecycle-explicit contributions: 
The \textit{first stream} is based on 46 out of 103 total articles, which refers to about 44.7\%.
It defines stages, maps to LCA, specifies system boundaries, or proposes stage-specific levers.
These will be captured in section~\ref{subsec:TransitiontotheGreenAILifecycle}. 
A \textit{second stream} (33 out of 103 articles, or rather 32.0\%) synthesizes or proposes formal definitions. These will be captured in section~\ref{subsec:DefiningGreenAI}. 
A \textit{third stream} concentrates on measurement assets (21/103 articles or 20.4\%) including metered and calibrated-estimator approaches, telemetry, and emission-factor provenance (see~section~\ref{subsec:GreenAIMeasurementTools}). 
A smaller but growing \textit{fourth stream} targets green hardware and infrastructure (see~section~\ref{subsec:GreenAIHardware}): About 13/103 articles (12.6\%) are covering embodied/material accounting, energy-proportional architectures, dynamic voltage and frequency scaling (DVFS); Running Average Power Limit (RAPL) with power capping, and carbon-aware scheduling. Since an article may belong to multiple categories, the presented counts reflect multi-label coding.

The subsequent analysis proceeds stream-wise to synthesize evidences on Green AI research step by step. 
By this, and remaining gaps relative to the Green AI lifecycle model and the guiding research questions are systematically identified.

\subsection{Definition of Green AI}
\label{subsec:DefiningGreenAI}

This section addresses RQ1 - ``How should \emph{Green AI} be defined and delimited relative to adjacent concepts, and which principles govern its lifecycle-wide application?'' 
A precise definition is required because research and practice increasingly report non-trivial environmental burdens (e.g. energy, carbon, water, and embodied impacts) across training and inference, with magnitudes that depend strongly on hardware choice, data-center region, scheduling, workload characteristics, and reporting practices \cite{strubell_energy_2019,kaack_aligning_2022,verdecchia_systematic_2023,bolon-canedo_review_2024}.

In response to RQ1, the section proceeds in three steps: 
first, a synthesis and critique of existing definitions and adjacent concepts (Green AI vs.\ Sustainable AI) (section~\ref{subsubsec:ExistingDefinitionsOfGreenAI}); 
second, a unified definition - in both concise and operational form - aligned with the Green AI lifecycle model introduced in section~\ref{subsec:TransitiontotheGreenAILifecycle} (section~\ref{subsubsec:ProposalOfAGreenAIDefinition}); and 
third, limitations and open challenges that delimit scope, specify boundaries (Scope~1-3; end-of-life), and systematically identify unresolved open measurement and reporting issues (section~\ref{subsubsec:LimitationsAndOpenChallenges}).

A structured approach is necessary because current usage varies along three fault lines: 
(i)~\emph{scope} (narrow efficiency during training vs.\ a system-level, lifecycle perspective across hardware, software, and interconnected data pipelines); 
(ii)~\emph{boundaries and metrics} (compute-centric proxies vs.\ impact-based indicators with emission-factor transparency, water accounting, and uncertainty disclosure); and 
(iii)~\emph{governance principles} (design-time vs.\ run-time commitments, thresholds, and explicit trade-off handling across lifecycle phases) \cite{schwartz_green_2020,xu_survey_2021,wu_sustainable_2022,kaack_aligning_2022}.

\subsubsection{Existing definitions of Green AI}
\label{subsubsec:ExistingDefinitionsOfGreenAI}

Addressing RQ1 begins with clarifying how \emph{Green AI} has been defined and used in recent scholarship. Early accounts conceptualize Green AI as \emph{efficiency under performance constraints} - maintaining accuracy while reducing computational cost \cite{schwartz_green_2020,xu_survey_2021}. Subsequent studies broaden this view from training-time efficiency to \emph{system-level and lifecycle} considerations, incorporating reporting standards, carbon-aware inference, and alignment with LCA stages \cite{patterson_carbon_2022,chen_survey_2023}. Several contributions move beyond energy and CO$_2$e to \emph{multi-impact} framings that include water and embodied/material impacts \cite{alzoubi_green_2024,bolon-canedo_review_2024}.
For instance, \cite{dodge_measuring_2022} define \textit{Scope 1} emissions as direct company emissions (e.g., from vehicles or AIs), \textit{Scope 2} emissions as indirect emissions from purchased energy (e.g. electricity, heating, cooling of operating AI systems), and \textit{Scope 3} emissions as other indirect emissions, such as those from investments or product use downstream (e.g. those associated with the upstream raw
materials extraction for manufacturing AI hardware, the use of AI-capable
IT asset infrastructure such as servers from suppliers to be used in
a cloud provider's datacenter for realizing local AI needs).

Across the literature, Schwartz's formulation is widely cited as the definitional anchor for Green AI. Numerous articles reference or operationalize that framing without introducing new definitional criteria \cite{strubell_energy_2019,strubell_energy_2020,henderson_towards_2020,trebaol_ecole_2020,bannour_evaluating_2021,cao_irene_2021,lannelongue_green_2021,li_full-cycle_2021,parcollet_energy_2021,shaikh_energyvis_2021,budennyy_eco2ai_2022,anzt_high_2022,dodge_measuring_2022,georgiou_green_2022,ligozat_unraveling_2022,li_making_2023,xu_energy_2023,yarally_uncovering_2023,yokoyama_investigating_2023,zhou_opportunities_2023}. This uptake underscores the influence of the efficiency-centric perspective but also reveals recurring omissions in boundary specification (Scope~1-3), impact coverage (water, embodied), measurement stance (compute proxies vs.\ metered/LCA-based metrics), and uncertainty disclosure.

Table~\ref{tab:Definitions} consolidates \emph{explicit} Green AI definitions. 
The left column gives verbatim excerpts; the middle classifies each definition by dominant emphasis 
(\emph{efficiency-centric}, \emph{impact-centric: energy/carbon}, \emph{multi-impact}, \emph{lifecycle-explicit}, \emph{governance/principles}); the right lists sources. 
Classification follows a deliberately conservative coding scheme (replication package) to show overlaps while transparently highlighting underlying conceptual intent and methodological consistency. 
This consolidation ultimately motivates the unified definition in section~\ref{subsubsec:ProposalOfAGreenAIDefinition}, integrating efficiency with lifecycle scope, multi-impact accounting, and governance principles.

\begin{table}[ht]
	\centering
	\caption{Definitions of Green AI across the literature, with dominant emphasis and source.}
				\begin{tabular}{p{9,8cm} p{2,8cm} p{2,2cm}}
				\toprule
				Definition & Dominant emphasis & Source \\
				\midrule
				\midrule
				The term Green AI refers to AI research that yields novel results while taking into account the computational cost, encouraging a reduction in resources spent. & Energy consumption / carbon footprint & Schwartz et~al. \citeyear {schwartz_green_2020} \cite{schwartz_green_2020} \\
				\hline
				Green AI, appeals to researchers to obtain novel results without increasing computational cost rather, ideally reducing it. & Energy efficiency & Xu et~al. \citeyear{xu_survey_2021} \cite{xu_survey_2021} \\
				\hline
				Green AI where the focus is on computing efficiency as well as model quality. & Energy efficiency / carbon footprint & Patterson et~al. \citeyear{patterson_carbon_2022} \cite{patterson_carbon_2022} \\
				\hline
				The research efforts of Green AI also reveal the critical need to take environmental impacts into major consideration when developing AI. & Energy consump. / carbon footprint & Chen et~al. \citeyear{chen_survey_2023} \cite{chen_survey_2023} \\
				\hline
				Approaches to build AI models that are environmentally friendly and inclusive. & Energy efficiency / carbon footprint & Mart\'inez-Fern\'andez et~al. \citeyear{martinez-fernandez_towards_2023} \cite{martinez-fernandez_towards_2023} \\
				\hline
				Advocating to consider the energy efficiency of AI algorithms during their development. & Energy consumption & Tornede et~al. \citeyear{tornede_towards_2023} \cite{tornede_towards_2023} \\
				\hline
				Green AI regards practices aimed at utilizing AI to mitigate the impact that humans have on the natural environment in terms of natural resources utilized, and/or mitigating the impact that AI itself can have on the natural environment. & Energy consumption, energy efficiency / carbon footprint & Verdecchia et~al. \citeyear{verdecchia_systematic_2023} \cite{verdecchia_systematic_2023} \\
				\hline
				The application of AI technology with an emphasis on energy consumption, CO$_2$e reduction, and environmental sustainability. & Energy efficiency / carbon footprint & Alzoubi and Mishra \citeyear{alzoubi_green_2024} \cite{alzoubi_green_2024} \\
				\hline
				A new paradigm, called Green AI, which incorporates sustainable practices and techniques in model design, training, and deployment that aim to reduce the associated environmental cost and carbon footprint. & Energy consump., energy efficiency, carbon footprint / water consumption & Bol\'on-Canedo et~al. \citeyear{bolon-canedo_review_2024} \cite{bolon-canedo_review_2024} \\
				\bottomrule
			\end{tabular}
	\label{tab:Definitions}
\end{table}

\clearpage

The scope of Green AI has expanded to encompass social sustainability and inclusivity \cite{martinez-fernandez_towards_2023}, differentiated lifecycle phases such as design, training, and deployment \cite{bolon-canedo_review_2024}, and environmental measures including water use \cite{bolon-canedo_review_2024}. Algorithmic efficiency is increasingly prominent \cite{tornede_towards_2023}, and explicit carbon reduction remains a priority \cite{alzoubi_green_2024}. Emerging strands operationalize the environmental scope through 
(i)~responsible hardware sourcing with embodied-impact accounting and end-of-life traceability \cite{gupta_act_2022,tongyodkaew_ai_2025,bolon-canedo_review_2024}, 
(ii)~circular-economy interventions (reuse, repair, remanufacture, recycle) with quantified footprint effects where reported \cite{tongyodkaew_ai_2025}, 
(iii)~infrastructure siting and regional equity as determinants of impact via grid carbon intensity and water scarcity \cite{kaack_aligning_2022,li_making_2023}, and 
(iv)~human-centered design choices that measurably alter energy and material use (e.g., latency/accuracy targets, accessibility constraints) \cite{bolon-canedo_review_2024,verdecchia_systematic_2023}. These developments position Green AI as a holistic, lifecycle-integrated approach.

An appraisal of Green AI's thematic scope reveals a concentration on energy, carbon, efficiency, and water, leaving hardware, complexity, duration, data management underexamined \cite{schwartz_green_2020,chen_survey_2023}. Lifecycle stages beyond training and inference - including data collection, labeling, and maintenance - contribute substantially to ecological costs yet remain understudied \cite{patterson_carbon_2022,verdecchia_systematic_2023}. Furthermore, trade-offs between algorithmic fairness, privacy, and environmental impact remain uncharted, despite direct implications for deployment choices \cite{wu_sustainable_2022,kaack_aligning_2022}. The human dimension is equally critical; evidence on how academic reward systems, industrial KPIs, and cultural norms bias actors against Green AI priorities remains sparse. Insights from organizational psychology and behavioral economics could illuminate barriers and motivators for embracing Green AI principles \cite{lv_unseen_2025}. Furthermore, current Green AI metrics tend to rely on static benchmarks. Developing adaptive, context-aware metrics that consider regional energy mixes, temporal fluctuations, and specific use cases would enable precise, actionable assessments \cite{eilam_towards_2023}. Exploring synergies with federated learning and neuromorphic hardware may reveal Sustainable AI pathways via distributed, energy-efficient paradigms \cite{malviya_neuromorphic_2024}.

\emph{Sustainable AI} designates a broad socio-technical program that combines environmental goals with requirements for fairness in algorithmic decisions, accessibility of digital services, and labor rights in global supply chains \cite{verdecchia_systematic_2023,wu_sustainable_2022,kaack_aligning_2022}. 
\emph{Green AI}, in contrast, narrows the lens to measurable ecological performance of AI systems. It targets energy demand during model development and inference, CO$_2$ emissions relative to regional grid mixes, water consumption from cooling, and embodied impacts arising from semiconductor fabrication and hardware logistics \cite{schwartz_green_2020,patterson_carbon_2022,bolon-canedo_review_2024}. 
This distinction underscores that Sustainable AI addresses normative governance challenges, whereas Green AI develops engineering-level levers for reducing the physical footprint of computation. 

\subsubsection{Proposal of a Green AI definition}
\label{subsubsec:ProposalOfAGreenAIDefinition}

Building on section \ref{subsubsec:ExistingDefinitionsOfGreenAI}, this section introduces a consolidated and citable definition of Green AI. It frames Green AI as the systematic pursuit of environmental sustainability throughout all lifecycle phases, anchored in verifiable artefacts and enforceable decision mechanisms. Core requirements include (i)~transparent boundary delineation across hardware, software, and data processes, ensuring reproducibility and accountability within environmental performance evaluation, (ii)~hybrid measurement combining direct metering (e.g., RAPL; performance monitoring counters, PMCs; rack-level power meters) and calibrated estimators with uncertainty intervals, and 
(iii)~decision criteria for targets and trade-offs expressed through \textit{Performance-Environmental Thresholds} (PET) and \textit{Phase Completion Criteria} (PCC). 
The proposed definition captures embodied-resource, water, and carbon impacts, thus transcending a narrow focus on runtime efficiency. It institutionalizes governance mechanisms - including carbon-aware scheduling and explicit uncertainty disclosure - within the Green AI lifecycle model. Implementation relies on three linked artefacts: (1)~phase demarcation, (2)~actionable levers per phase (e.g., DVFS, power capping, pruning, carbon-aware routing, reuse), and (3)~Plan-Do-Check-Act (PDCA) gateways enforcing PET and PCC criteria. This architecture ensures that conceptual rigor is paired with operational reproducibility across codebases, datasets, and metering evidence.
\\
\\
\noindent\textbf{Def.:}
\emph{
	Green AI refers to a state of best AI system realization 
	considering all research approaches, common practices and techniques
	(this includes algorithms, process context, hardware used, ...) to achieve, 
	within the declared system and its AI lifecycle boundaries, 
	the minimum of the AI's cumulative negative environmental impacts, 
	such as operational and embodied energy, greenhouse-gas (CO$_2$) emissions, water consumption, material footprints, and further resources (time, costs, capacities, ...),
	and the AI's maximum of its cumulative positive environmental impacts, 
	such as human labor improvement, work acceleration, product quality increase, production cost saving, 
	in AI-related scenarios (training, testing, validation, inference, refinement, transfer, ...)
	for AI-task fulfillment throughout the corresponding life-cycle of an AI,
	is evaluated based on measured or calibrated evidence 
	in regard with published targets or thresholds with transparent trade-off rules with documented uncertainty,
	and is going along with economical, society and governmental constraints. 
	}
\\

The definition treats Green AI as a lifecycle-explicit specification \textit{process} covering phases from raw-material extraction and hardware/software provisioning through design, development, deployment, inference, maintenance, and end-of-life. 
It \textit{requires} 
(i)~explicit boundary setting and phase mapping,
which includes for instance assets, phases and scope 1-3; 
(ii)~impact quantification with units and provenance (e.g., location- or market-based emission factors and time horizon); 
(iii)~published targets and trade-off rules that relate environmental indicators to task performance and economic constraints; 
(iv)~phase-appropriate levers (e.g., siting and hardware selection; dynamic voltage and frequency scaling; power capping; pruning/quantization/distillation; batch sizing/placement and carbon-aware scheduling; repair/reuse/recycling) 
governed by continuous monitoring and iterative adjustment; 
as well as (v)~transparent documentation and artifacts for reproducibility. 

Further, the definition asks for \textit{boundaries}, that are to be declared at system and phase levels (assets, activities, Scope~1-3). 
\textit{Impacts} are quantified with units and provenance: energy (J/kWh), greenhouse-gas emissions (kg~CO$_2$e with location- or market-based factors and stated vintage/horizon), water (L and water usage effectiveness, WUE), and embodied/material footprints (e.g., via Environmental Product Declarations, EPDs). 
\textit{Measurement} relies on direct metering (e.g., RAPL/PMCs, rack-level meters, facility telemetry) or \textit{calibrated estimators} with documented error and uncertainty. 
The \textit{scope} concerns the environmental impacts of AI systems themselves; 
applications that use AI to mitigate external environmental burdens
or would like to benchmark with the Green AI's performance (in the sense of concurrent approach),
are considered only when their effects are quantified within the same declared boundaries.
Allocation of \textit{embodied impacts} follows amortization over hardware lifetime, including replacement cycles and secondary reuse. 
To reduce reporting heterogeneity and guarantee \textit{measurement standards}, emission factors (e.g. grid carbon intensity, cooling water coefficients) and resource proxies (e.g. water, material footage) should state regional and temporal granularity, provenance, and uncertainty. Where proxies are used (e.g., GPU-hours, FLOPs), calibration against the same qualified and standardized metered data is required to avoid bias. 
Finally, \textit{uncertainty} is treated as a first-class attribute: confidence intervals, propagation of estimator error, and disclosure of measurement limits are mandatory for reproducible comparison across studies and providers.

Comparative evaluation is performed against baselines under a stated functional unit and service-level objectives: per training run, per inference, or per task at target accuracy/latency. 
\textit{Valid comparators} include 
(i) an internal predecessor/version, 
(ii) an external best-practice or state-of-the-art configuration, and 
(iii) a non-AI baseline. 
\textit{Results} are normalized to the functional unit and lifecycle stage, with disclosure of allocation rules for shared infrastructure (Power Usage Effectiveness, PUE; WUE assumptions, regional grid mix, cooling). 
\textit{Decision rules} use PET and PCC gates to relate environmental indicators to performance and cost, with carbon-aware siting/dispatch considered where applicable. 
All benchmarks report hardware class, dataset/workload, configuration, emission-factor sources, and uncertainty, enabling auditable, lifecycle-explicit comparisons.

The ``\textit{Green AI balance}'' is a lifecycle-explicit evaluation construct that aggregates measured environmental indicators (energy, CO$_2$e, water, embodied/material footprints) across declared system and phase boundaries (assets, activities, Scope~1-3) for a given AI system.
It situates AI solutions along a continuum from ``\textit{less green}'' to ``\textit{more green}'' using validated ecological and material indicators, explicit weighting schemes, and transparent evaluation protocols. 
This framework facilitates systematic trade-off analysis between residual impacts, performance, and efficiency, while explicitly accounting for methodological uncertainties and context-dependent value judgments. In doing so, it provides the basis for evidence-driven comparison, iterative refinement, and future algorithmic designs that reduce ecological and resource footprints.

\subsubsection{Limitations and open challenges}
\label{subsubsec:LimitationsAndOpenChallenges}

While the proposed Green AI definition anchors lifecycle boundaries and impact categories, it still lacks operational specificity. Current criteria often conflate device-level efficiency with system-wide Green AI, preventing reproducible benchmarking across platforms and scales. 
Unified testbeds that co-measure carbon, water, and embodied resources would provide the empirical ground for verifiable Green AI evaluation \cite{strubell_energy_2019}. Green AI should remain an ecologically delimited concept, distinct from broader sustainability narratives, ensuring analytical clarity. Consequently, standardization of metrics and protocols for both harmful and beneficial impacts - including emergent indicators such as circular-material reuse or AI-for-environment feedback - is indispensable for methodological alignment and large-scale adoption.

Robust comparison currently hinges on unresolved choices in 
(i)~boundary conventions - system assets (compute, storage, networking), lifecycle stages (cradle-to-grave), organizational scopes (Scope~1-3), and allocation rules for shared infrastructure (virtualization, multi-tenancy, PUE, and WUE assumptions) \cite{clemm_towards_2024}; 
(ii)~data and provenance - availability of bill-of-materials/EPDs for embodied impacts, emission-factor sources (location- vs.\ market-based) with temporal granularity and vintage, and regional water-scarcity weighting \cite{henderson_towards_2020}; and 
(iii)~uncertainty treatment - meter accuracy classes and sampling rates, estimator calibration error, and propagation of uncertainty into decision rules (PET, PCC gates). Without specifications for functional units (per training run, per inference, per task at target accuracy/latency), baselines, amortization of embodied impacts over service life/duty cycle, and disclosure of allocation and emission-factor choices, cross-study assessments remain fragile \cite{patterson_carbon_2022}.

Accordingly, constructs such as the \emph{Green AI balance} and phase-specific \emph{algorithmic processing} levers (e.g., pruning, quantization, distillation, early-exit) should be treated as candidate evaluation tools until validated. Immediate priorities include 
(i)~reference workloads with fixed datasets/targets and canonical hardware classes, 
(ii)~cross-meter calibration protocols (device PMCs/RAPL vs.\ rack-level meters vs.\ facility telemetry) with acceptance bounds, 
(iii)~inter-site round-robin studies to test reproducibility under differing grid mixes and cooling, and 
(iv)~standardized reporting checklists covering units, provenance, allocation rules, and uncertainty disclosure \cite{schwartz_green_2020,henderson_towards_2020}. 
These methodological issues, together with governance aspects directly tied to environmental performance (targets, thresholds, auditability, uncertainty disclosure), are examined in detail in section~\ref{sec:Discussion}.

\subsection{Transition to the Green AI lifecycle}
\label{subsec:TransitiontotheGreenAILifecycle}

This section operationalizes the unified Green AI definition into a phase model with system and lifecycle boundaries, phase-specific levers, and decision gates. The model spans provisioning and siting, model development, training, deployment/inference, maintenance, and end-of-life, with measurable environmental indicators (energy, CO$_2$e, water, embodied/material) per phase. Each phase defines artefacts (configuration, workload, metering/calibration logs) to enable auditable reductions in environmental impact at preserved performance.

To characterise the state of the art, a four-step procedure was applied, namely 
(1)~terminology extraction with keyword-in-context and consolidation of synonyms to obtain a harmonized vocabulary; 
(2)~quantitative term analysis using length-normalised frequencies and a composite salience ranking, 
(3)~coverage coding to construct a binary source-criterion matrix and corresponding heatmaps, and 
(4)~hierarchical clustering of term co-occurrence to induce a stage hierarchy. Details for each step and the resulting artefacts are provided in the steps below.

Addressing RQ2 - ``Which phases and subphases constitute the \emph{Green AI lifecycle}, and how do they map to Life Cycle Assessment (LCA) stages?'' - is executed as an evidence-ordered exposition, since it 
(i)~reports the lifecycle stages attested in the corpus,
(ii)~synthesizes them into a unified scheme of five phases and 33 subphases aligned with the dendrogram leaf order, 
(iii)~quantifies coverage and imbalances at phase level granularity, and
(iv)~consolidates the results into a process-oriented lifecycle model for implementation.

For this, section~\ref{subsubsec:StateoftheArtGreenAILifecycleStages} enumerates the lifecycle stages attested in the corpus using the harmonized vocabulary, salience ranking, and the chronological heatmap; the dendrogram (Fig.~\ref{fig:DendrogramAILifecycle}) fixes a semantically consistent column order.
These signals are consolidated into a unified scheme of five phases with 33 subphases in section~\ref{subsubsec:SynthesizingGreenAILifeCyclePhases}, that are aligned to the dendrogram leaf order and the coding rules used for coverage for systematic comparison.
Thereafter, section~\ref{subsubsec:QuantitativeAssessmentofLifecyclePhaseRepresentation} quantifies phase-level coverage and imbalance (criterion prevalence, mean coverage per phase, overall breadth), identifying over- and under-represented areas based on the matrix in Table~\ref{tab:Lifecycle}.
Finally, section~\ref{subsubsec:ProposalofaGreenAILifecycleProcessModel} instantiates a process-oriented Green AI lifecycle with phase-specific levers as PET and decision gates as PCC, providing artefacts, indicators, and metrics for practical implementation.

This staged exposition connects term-level evidence to phase-level synthesis and quantified coverage, yielding a reproducible lifecycle model with declared system and phase boundaries, phase-specific levers, and measurable indicators (units and provenance) that can be applied directly for implementation and audit.

\subsubsection{State-of-the-art of Green AI lifecycle stages}
\label{subsubsec:StateoftheArtGreenAILifecycleStages}

To quantify how Green AI is distributed across the AI lifecycle, the analysis moves from term-level evidence to phase-level evidence in six bounded steps. 
Step~1 produces a harmonized vocabulary from the 43-article corpus (164 lifecycle-related term groups via keyword-in-context extraction and synonym consolidation \cite{frantzi_automatic_2000}). 
Step~2 computes length-normalised frequencies and a composite salience score to separate head from tail. 
Step~3 builds a chronological heatmap for the 34 highest-salience terms. 
Step~4 derives a dendrogram from term co-occurrence to fix a stable, semantically consistent column order. 
The step~5 aggregates terms into phases/subphases. 
Finally, step~6 can quantify coverage (criterion prevalence, mean per phase, overall breadth) \cite{egghe_h-index_2008,donthu_how_2021}. 
This pipeline replaces narrative impressions with auditable, reproducible, and methodologically transparent identification of over- and under-represented lifecycle areas and can be found in the following.

\emph{Step~1 - Identification and grouping of lifecycle-related terms.}\\
From the 43 articles in the SLR, keyword-in-context analysis of titles, abstracts, and author keywords extracted 164 lifecycle-relevant terms. Frequency thresholds removed singletons, and systematic consolidation of morphological variants (e.g., train vs. training) and technical synonyms (e.g., deployment vs. operationalization) produced the harmonized set:
(e.g., \emph{training}; \emph{model training}; \emph{deployment}; \emph{operationalization}) \cite{frantzi_automatic_2000}. The candidate set was generated from both author-provided keywords and automated term extraction from titles and abstracts, ensuring comprehensive representation of domain-specific vocabulary across all AI lifecycle stages. 

To maximise conceptual precision and eliminate redundancy, synonymous or closely related expressions were consolidated into unified term clusters (e.g., \emph{training};\emph{model training}; \emph{deployment}; \emph{operationalization}) following established principles of terminological harmonization \cite{temmerman_towards_2000}. The resulting harmonized vocabulary spans all major lifecycle phases and includes both high-frequency head terms (e.g., \emph{development} \cite{masrour_ai_2023}, \emph{inference} \cite{hilty_rebound_2015}, \emph{production} \cite{georgiou_green_2022}, \emph{power consumption} \cite{garcia-martin_estimation_2019}) and strategically important low-frequency tail terms occurring only twice in the corpus (e.g., \emph{approximate computing}, \emph{green data centers}, \emph{on device learning}). 

Quantitatively, the consolidated set exhibits a median occurrence of four articles per term, indicating substantial variability in corpus-wide coverage and the presence of a pronounced long-tail distribution. This harmonized vocabulary serves as the methodological foundation for all subsequent quantitative analyses, including frequency distributions, lifecycle coverage evaluation, and hierarchical clustering of concept co-occurrence (see Table~\ref{tab:ConsolidatedTermList} for the excerpt term list, grouping rationale, and statistical profiles). It also enables the computation of composite salience scores and lifecycle phase balance metrics, as described in step~2.

\emph{Step~2 - Quantitative analysis of lifecycle terms.}\\
Normalized term frequencies (per 1,000 words) were calculated for all lifecycle-related vocabulary items extracted in step~1. 
To prioritize terms, a composite salience score was derived that combines three factors: length-adjusted frequency, dispersion across the corpus (Juilland's~D), and temporal growth (the standardized slope of yearly frequency trends). 
The separation between high- and low-frequency terms was determined using a piecewise regression ``elbow'' on the rank-frequency curve, thereby avoiding arbitrary thresholds.
Two comparative bar charts clearly illustrate the resulting distribution: head terms (more than eight occurrences; Fig.~\ref{fig:BarchartLifeCyclePhasesOver8}) and tail terms (eight or fewer occurrences; Fig.~\ref{fig:BarchartLifeCyclePhasesUnder9}).

The head panel is dominated by training- and deployment concepts (e.g., \emph{model training}, \emph{deployment}, \emph{inference}, \emph{power consumption}). A Gini coefficient of 0.74 captures the concentration of attention in few lifecycle terms, while the Intervention Bias Ratio of 3.6 confirms a persistent overemphasis on compute-focused interventions at the expense of infrastructure and end-of-life phases.

In contrast, the long tail emphasizes infrastructure and end-of-life topics: \emph{dynamic voltage and frequency scaling, DVFS}/power capping and carbon-aware scheduling (inference), renewable-energy integration and green data centers (infrastructure), supplier transparency and conflict minerals (provisioning), and remanufacturing or e-waste pathways (end-of-life). Although sparse, their composite salience is amplified by cross-source recurrence and visible temporal growth. The Phase Balance Score confirms a systematic skew: training and deployment dominate above baseline, while supply-chain and end-of-life fall well below.This imbalance highlights that existing studies focus on computational efficiency while failing to capture material sourcing and post-use recovery.

\begin{figure}[ht!]
	\centering
	\includegraphics[width=\textwidth,height=0.95\textheight,keepaspectratio]{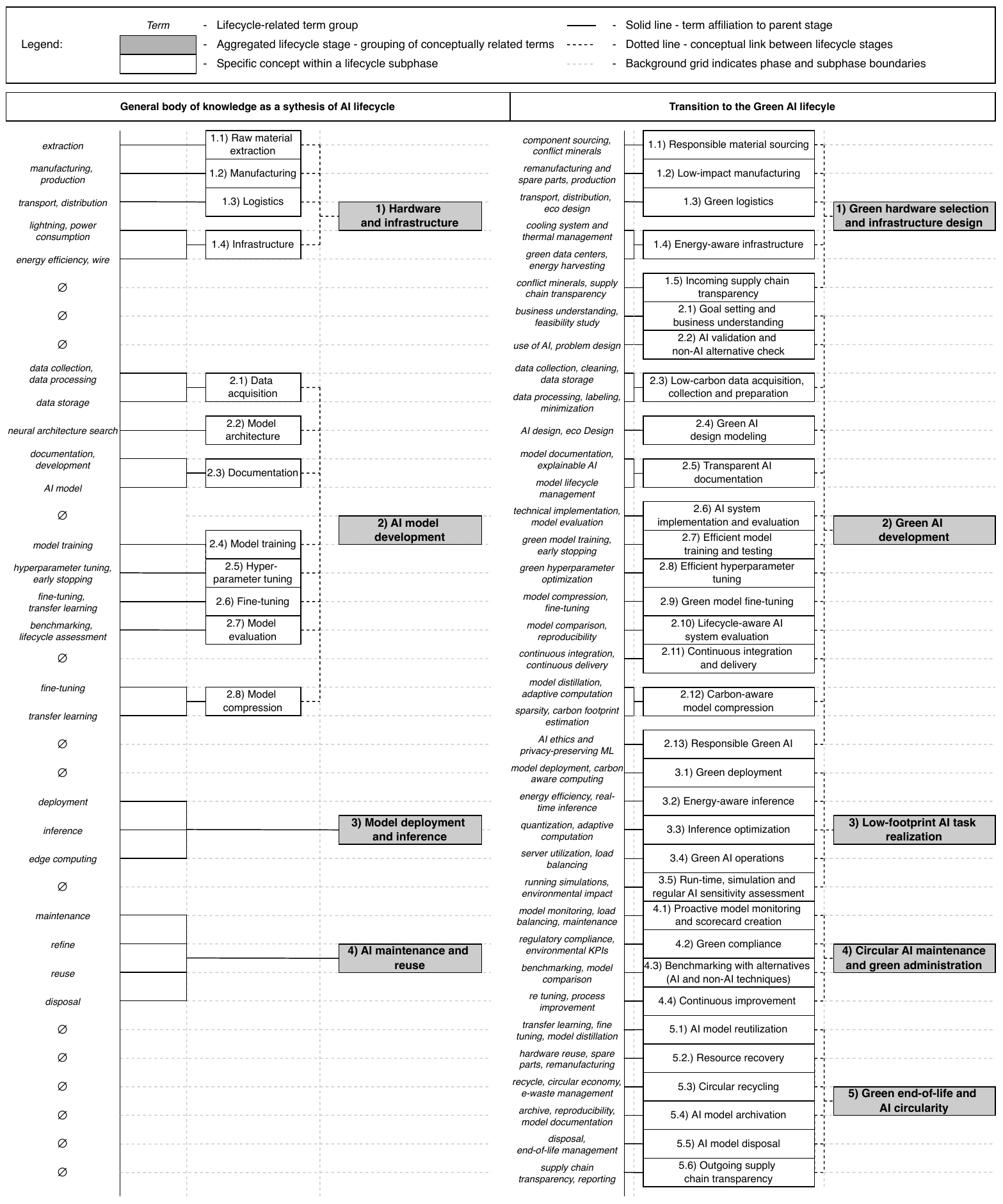}
	\caption{Comparison of gernal body of knowledge of AI lifecycle and transition to the Green AI lifecycle as dendrograms.}
	\label{fig:DendrogramAILifecycle}
\end{figure}

\emph{Step~3 - Visualization of lifecycle terms.}\\
Building on the previous step, step~3 adds a temporal perspective via a chronological heatmap. 
The 34 most frequent harmonized lifecycle terms were coded as present/absent per article under the step~1 rules and arranged with rows strictly ordered by article publication year (Table~\ref{tab:Heatmap}). 
To preserve semantic neighbourhoods and enable direct comparative analysis across related figures, the column order follows the dendrogram leaf order in Fig.~\ref{fig:DendrogramAILifecycle} and aligns consistently with the structured lifecycle phase aggregation in Table~\ref{tab:Lifecycle}.

This dual ordering preserves temporal comparability (rows) and semantic proximity (columns), allowing for longitudinal trend detection while maintaining the structural relationships revealed in step~4. 
Three main patterns emerge from the heatmap. 
First, a \emph{compute-centric} focus dominates early contributions (pre-2020), with the most frequent columns being \emph{development} (\(39\) sources), \emph{inference} (\(35\)), and \emph{production} (\(33\)), while upstream phases (e.g., \emph{raw material extraction}) and downstream impacts (e.g., \emph{disposal}, \(11\)) remain underrepresented. 
Second, coverage breadth increases noticeably in recent years: several 2024-2025 articles reach per-source coverage values of \(\omega \in [14,20]\), compared to \(\omega \leq 8\) for many pre-2020 articles. 
This reflects a gradual broadening from narrow, compute-heavy scopes toward more holistic lifecycle considerations. 
Third, columns for measurement-governance indicators with direct bearing on environmental accounting (e.g., emission-factor disclosure, supplier traceability/auditability) and end-of-life levers (reuse, repair, recycling) exhibit late-onset, low-persistence patterns - long runs of zeros punctuated by isolated activations in 2024-2025. In Table~\ref{tab:Heatmap}, the matrix contains \(\sum=640\) filled cells, yielding a fill rate of \(\frac{640}{34 \times N_{\text{sources}}}\); sparsity is concentrated in upstream (provisioning/supply chain) and downstream (end-of-life) columns, whereas training/inference columns maintain continuous coverage over time. 
Recent contributions increase per-source coverage \(\omega\) and add operations/hardware facets, but these gains remain column-local and do not close the upstream/downstream gap \cite{eilam_towards_2023,barbierato_toward_2024}. The chronological mapping therefore corroborates the compute-centric overhang; step~5 aggregates these columns into subphases where prevalence remains low relative to training/deployment and quantifies the shortfall at phase level.

\emph{Step~4 - Hierarchical structuring via dendrograms.}\\
To address the identified gap between isolated term usage and holistic lifecycle conceptualization, the dendogram-based structuring advances from binary co-occurrence data to a process-oriented ontology. 
Leveraging agglomerative hierarchical clustering \cite{rokach_clustering_2005} with a Jaccard distance metric, conceptual proximities between the harmonized lifecycle terms are extracted and visualized as dendrograms. 
Unlike flat frequency tables, this method preserves hierarchical term relations and exposes structural clusters even in sparsely represented lifecycle phases. The first dendrogram reconstructs the general body of knowledge as a synthesis of AI lifecycle phases from the literature corpus, while the second extends this scaffold into the proposed Green AI lifecycle (Fig.~\ref{fig:DendrogramAILifecycle}), integrating additional upstream (e.g., \emph{responsible resource sourcing}) and downstream (e.g., \emph{circular AI maintenance}) phases, as well as green end-of-life and AI cicularity topics. 

Juxtaposing the two dendrograms surfaces two concrete shifts. 
(1)~Measurement- and governance-related terms with direct impact-accounting roles (e.g., emission-factor disclosure, metering/calibration logs, Power Usage Effectiveness, PUE; Water Usage Effectiveness, WUE) no longer sit as peripheral annotations to training/inference; in the Green-AI structure they align as a cross-cutting spine linking provisioning/siting to deployment/operation, thereby bridging hardware/infrastructure with model-centric phases. 
(2)~Upstream and downstream subphases that are terminal leaves in the traditional map - \emph{disposal}, \emph{remanufacturing}, \emph{supply-chain transparency} - co-cluster with provisioning phases (\emph{responsible material sourcing}, \emph{low-impact manufacturing}, \emph{green logistics}) in the Green-AI dendrogram. 
This co-clustering identifies supply-chain and end-of-life as quantifiable leverage points. Environmental Product Declarations (EPDs) and supplier audits (e.g., cobalt and tantalum traceability) translate procurement into verifiable inventories; downstream, structured take-back programs, component-level repair, and certified recycling demonstrably cut embodied CO$_2$ emission and reduce dependence on virgin extraction. 
Training/inference columns show sustained coverage, while provisioning/supply-chain and end-of-life clusters remain sparse and late-emerging. The shared leaf order furnishes a fixed column basis for the phase aggregation in Table~\ref{tab:Lifecycle} and for quantifying the resulting imbalances. However, both perspectives remain rooted in individual term occurrences or clustered concepts. 
To move from micro-level lexical analysis to a macro-level assessment of lifecycle comprehensiveness, it is necessary to aggregate terms into their corresponding phases and systematically measure coverage breadth. 
This is addressed in step~5 and step~6. 
So, step~5 acts as the bridge between qualitative term coding and quantitative phase-level assessment, consolidating the fragmented term-level findings into a structured lifecycle map.

\emph{Step~5 - Phase-level mapping and coverage matrix.}\\
To enable a more holistic assessment, all harmonized terms from step~1 were aggregated into their corresponding subphase and phase categories within the proposed Green AI lifecycle ontology (Fig.~\ref{fig:DendrogramAILifecycle}). This aggregation collapses heterogeneous terminology into a structured set of 33 subphases nested under five high-level phases, ensuring that coverage analysis reflects substantive lifecycle dimensions rather than isolated lexical choices and explicitly accommodates upstream supply-chain and downstream end-of-life processes.

For each of the 43 reviewed articles, a binary coding was applied at the subphase level (\checkmark\ = criterion explicitly addressed; empty cell = not addressed), based on explicit mention and operational treatment of at least one associated term within the article's scope. This coding yields a rectangular coverage matrix (\emph{Source}~$\times$~\emph{Subphase}), where columns represent the 33 subphases (grouped under their five parent phases) and rows represent individual articles ordered chronologically. The resulting phase-level heatmap (Tab.~\ref{tab:Lifecycle}) is fully aligned with the dendrogram leaf order from step~4, maintaining visual and semantic continuity across term-, cluster-, and phase-level views while enabling systematic longitudinal comparison across the entire corpus.

A \checkmark\ is recorded only when a subphase is evidenced by \emph{operational content}: a concrete method/lever (e.g., dynamic voltage and frequency scaling, DVFS/power capping, carbon-aware routing), a quantitative metric with unit and declared scope/provenance (e.g., energy in J/kWh; CO$_2$ emission in kg using stated emission factors), or an implemented procedure with reproducible configuration/results. 

\clearpage

Pure narrative mentions, problem statements, or policy aspirations without method/measurement are not coded; ambiguous cases are treated as missing (no \checkmark) and documented in the audit log with detailed explanatory rationale provided.
The rightmost column ($\Sigma$) counts fulfilled subphases per source; the bottom row (\emph{Total per Criterion}) counts sources per subphase. This makes coverage breadth comparable across studies and phases, explicitly highlighting systematic gaps, biases, and persistent imbalances in lifecycle phase attention.

\emph{Step~6 - Statistical analysis of lifecycle coverage.}\\ 
To quantify coverage, the phase-level source-subphase matrix (Table~\ref{tab:Lifecycle}) was evaluated using three pre-specified metrics: \emph{criterion prevalence} (percentage of sources explicitly addressing a subphase), \emph{per-phase mean coverage} (average share of subphases addressed within a phase), and \emph{overall lifecycle breadth} \((\omega)\) (mean number of fulfilled criteria per reviewed source).

(1) \emph{Criterion prevalence} - the percentage of sources consistently addressing a given subphase, corresponding to the \emph{Total per Criterion} row normalized by the corpus size; 
(2) \emph{Per-phase mean coverage} - the average share of distinct subphases addressed per phase across all sources (\emph{Mean per Phase \%}); and 
(3) \emph{Overall lifecycle breadth} - the mean number of distinct fulfilled criteria per source, reported in the rightmost column ($\omega$). 

The aggregated results reveal strong imbalances. Deployment-related phases (\emph{efficient model training and testing}, \emph{energy-aware inference}) exhibit the highest mean coverage rates, reaching up to 39.5\%. 
Hardware-oriented phases - \emph{responsible material sourcing} and \emph{green logistics} - achieved 36.7 \% coverage, exceeding \emph{Green AI development} (24.0 \%) and \emph{circular AI maintenance} (22.1 \%), while \emph{green end-of-life and AI circularity} remained lowest at 17.1 \%. These metrics corroborate the term-frequency analysis from step~2, confirming that lexical underrepresentation translates into structural lifecycle blind spots. Low-salience yet strategically crucial subphases - such as \emph{circular recycling} and \emph{green compliance} - emerge as priority levers for future Green AI research and policy alignment.

Taken together, steps~5 and~6 close the methodological loop: from raw terminology to structured, phase-level coverage analysis, enabling defensible, evidence-based prioritisation of underrepresented lifecycle phases and providing clear targets for advancing a genuinely end-to-end Green AI agenda.

\subsubsection{Synthesizing Green AI lifecycle phases}
\label{subsubsec:SynthesizingGreenAILifeCyclePhases}

Building on the clustered terms, the harmonized vocabulary was mapped onto a five-phase Green AI lifecycle that spans hardware, software, and operational stages, using the dendrogram leaf order to assign terms to phase-specific subphases: 
(1)~\emph{Green hardware selection and infrastructure design} (provisioning, siting, data-centre operations), 
(2)~\emph{Green AI development} (architecture/design, training setup, evaluation), 
(3)~\emph{Low-footprint AI task realization} (placement, batching, serving), 
(4)~\emph{Circular AI maintenance and green administration} (monitoring, updates, retraining triggers), 
(5)~\emph{Green end-of-life and AI circularity} (take-back, repair/reuse, recycling). 
This structure improves targeted assessment: training- and inference-oriented improvements are well represented \cite{garcia-martin_estimation_2019,patterson_carbon_2021}, whereas end-of-life flows such as hardware recycling remain sparsely evidenced \cite{alzoubi_green_2024}. 
The following section details each phase, beginning with process synthesis and system definition as the basis for subsequent steps.
\\
\\
\noindent \emph{4.2.2.1 System definition}\\
The main phases synthesized from the literature corpus are clarified in the following first.
Their subphase tasks are clarified thereafter.

\begin{table}[t]
	\centering
	\caption{Phase-level heatmap of the Green AI lifecycle. The heatmap ranges from blue-white-red.
	A check mark (\checkmark) indicates that a criterion is addressed in the corresponding article. 
	The rightmost column ($\Sigma$) reports the number of fulfilled criteria per source, 
	while the bottom row (\emph{total per criterion}) shows how many sources address each criterion. 
	Background shading highlights aggregate rows for readability. 
	Columns follow the dendrogram leaf order (Fig.~\ref{fig:DendrogramAILifecycle}) and the term-level heatmap (Table~\ref{tab:Heatmap}), 
	with consistent colour codes and phase patterns to ensure cross-figure comparability. 
	In addition to coverage per phase (\%), it reports the mean number of fulfilled criteria per phase and per article.}
	\scalebox{0.5}{
		\begin{tabular}{@{} |l
				| *{5}{>{\centering\arraybackslash}m{0.33cm}} |
				*{13}{>{\centering\arraybackslash}m{0.33cm}} |
				*{5}{>{\centering\arraybackslash}m{0.33cm}} |
				*{4}{>{\centering\arraybackslash}m{0.33cm}} |
				*{6}{>{\centering\arraybackslash}m{0.33cm}} | m{0.51cm} |@{}}
			\toprule
			\textbf{Synthesis} & \multicolumn{5}{l|}{\textbf{1) Green hardware}} & \multicolumn{13}{l|}{\textbf{2) Green AI development}} & \multicolumn{5}{l|}{\textbf{3) Low-footprint}} & \multicolumn{4}{l|}{\textbf{4) Circular AI }} & \multicolumn{6}{l|}{\textbf{5) Green end-of-life}} & \\
			& \multicolumn{5}{l|}{\textbf{selection and}} & \multicolumn{13}{l|}{} & \multicolumn{5}{l|}{ \textbf{AI task}} & \multicolumn{4}{l|}{\textbf{maintenance}} & \multicolumn{6}{l|}{\textbf{and AI circularity}} & \\
			& \multicolumn{5}{l|}{\textbf{infrastructure}} & \multicolumn{13}{l|}{} & \multicolumn{5}{l|}{ \textbf{realization}} & \multicolumn{4}{l|}{\textbf{and green}} & \multicolumn{6}{l|}{ \textbf{}} & \\
			\textbf{} & \multicolumn{5}{l|}{\textbf{design}} & \multicolumn{13}{l|}{\textbf{}} & \multicolumn{5}{l|}{\textbf{}} & \multicolumn{4}{l|}{\textbf{administration}} & \multicolumn{6}{l|}{\textbf{}} & \\
			\cmidrule(lr){1-1}\cmidrule(lr){2-6}\cmidrule(lr){7-19}\cmidrule(lr){20-24}\cmidrule(lr){25-28}\cmidrule(lr){29-34}\cmidrule(l){35-35}
			Source & \rotatebox{90}{\shortstack{1.1) Responsible material sourcing}} & 
			\rotatebox{90}{\shortstack{1.2) Low-impact manufacturing}} & 
			\rotatebox{90}{\shortstack{1.3) Green logistics}} & 
			\rotatebox{90}{\shortstack{1.4) Energy-aware infrastructure}} & 
			\rotatebox{90}{\shortstack{1.5) Incoming supply chain transpar.}} & 
			\rotatebox{90}{\shortstack{2.1) Goal setting, business understanding}} & 
			\rotatebox{90}{\shortstack{2.2) AI validation and non-AI alternative}} & 
			\rotatebox{90}{\shortstack{2.3) Low-carbon data acquisition}} & 
			\rotatebox{90}{\shortstack{2.4) Green AI design modeling}} & 
			\rotatebox{90}{\shortstack{2.5) Transparent AI documentation}} & 
			\rotatebox{90}{\shortstack{2.6) AI system implementation and eval.}} & 
			\rotatebox{90}{\shortstack{2.7) Efficient model training and test.}} & 
			\rotatebox{90}{\shortstack{2.8) Efficient hyperparameter tuning}} & 
			\rotatebox{90}{\shortstack{2.9) Green model fine-tuning}} & 
			\rotatebox{90}{\shortstack{2.10) Lifecycle-aware AI system eval.}} & 
			\rotatebox{90}{\shortstack{2.11) Continuous integration and delivery}} & 
			\rotatebox{90}{\shortstack{2.12) Carbon-aware model compression}} &
			\rotatebox{90}{\shortstack{2.13) Responsible Green AI}} &
			\rotatebox{90}{\shortstack{3.1) Green deployment}} &
			\rotatebox{90}{\shortstack{3.2) Energy-aware inference}} &
			\rotatebox{90}{\shortstack{3.3) Inference optimization}} &
			\rotatebox{90}{\shortstack{3.4) Green AI operations}} & 
			\rotatebox{90}{\shortstack{3.5) Run-time, simulation and reg. AI sens.}} & 
			\rotatebox{90}{\shortstack{4.1) Proactive model monitoring}} & 
			\rotatebox{90}{\shortstack{4.2) Green compliance}} &
			\rotatebox{90}{\shortstack{4.3) Benchmarking with alternatives}} &
			\rotatebox{90}{\shortstack{4.4) Continuous improvement}} & 
			\rotatebox{90}{\shortstack{5.1) AI model reutilization}} &
			\rotatebox{90}{\shortstack{5.2) Ressource recovery}} &
			\rotatebox{90}{\shortstack{5.3) Circular recycling}} &
			\rotatebox{90}{\shortstack{5.4) AI model archivation}} &
			\rotatebox{90}{\shortstack{5.5) Ai model disposal}} &
			\rotatebox{90}{\shortstack{5.6) Outgoing supply chain transpar.}} & 
			$\sum$ \\
			\midrule
			Gossart \citeyear{hilty_rebound_2015} \cite{hilty_rebound_2015} &  & \checkmark & \checkmark &  &  &  &  & \checkmark &  &  &  &  &  &  &  &  &  &  & \checkmark & \checkmark &  &  & \checkmark &  &  &  &  &  &  &  & \checkmark &  &  & \cellcolor[HTML]{AAAAFF} 7 \\
			Batra et~al. \citeyear{batra_artificial-intelligence_2018} \cite{batra_artificial-intelligence_2018} &  &  &  &  &  &  &  &  &  &  &  &  &  &  &  &  &  &  &  & \checkmark &  &  &  &  &  &  &  &  &  &  &  &  &   & \cellcolor[HTML]{9595FF}1 \\
			Lacoste et~al. \citeyear{lacoste_quantifying_2019} \cite{lacoste_quantifying_2019} &  & \checkmark & \checkmark &  &  &  &  &  &  &  &  & \checkmark & \checkmark &  &  &  &  &  &  & \checkmark &  & \checkmark &  & \checkmark &  &  &  &  &  &  &  &  &  & \cellcolor[HTML]{AAAAFF} 7 \\
			Wu et~al. \citeyear{wu_machine_2019} \cite{wu_machine_2019} &  & \checkmark & \checkmark &  &  &  &  &  &  &  &  &  &  &  &  &  & \checkmark &  & \checkmark & \checkmark & \checkmark &  &  &  &  & \checkmark &  &  &  &  &  &  &  & \cellcolor[HTML]{AAAAFF} 7 \\
			Henderson et~al. \citeyear{henderson_towards_2020} \cite{henderson_towards_2020} &  & \checkmark & \checkmark &  &  &  &  & \checkmark &  &  &  & \checkmark &  &  & \checkmark &  &  &  & \checkmark & \checkmark &  &  &  &  &  & \checkmark &  &  &  &  & \checkmark &  &  & \cellcolor[HTML]{EEEEFF}9 \\
			Anthony et~al. \citeyear{anthony_carbontracker_2020} \cite{anthony_carbontracker_2020} &  & \checkmark &  &  &  &  &  &  &  &  &  & \checkmark & \checkmark &  & \checkmark &  &  &  &  & \checkmark & \checkmark &  &  &  &  & \checkmark &  &  &  &  & \checkmark &  &  & \cellcolor[HTML]{CCCCFF} 8 \\
			Schwartz et~al. \citeyear{schwartz_green_2020} \cite{schwartz_green_2020} &  & \checkmark &  &  &  &  &  &  &  &  &  & \checkmark & \checkmark & \checkmark &  &  &  &  &  & \checkmark &  &  &  &  &  &  &  & \checkmark &  &  &  &  &  & \cellcolor[HTML]{9595FF} 6 \\
			Tr\'ebaol \citeyear{trebaol_ecole_2020} \cite{trebaol_ecole_2020} &  & \checkmark & \checkmark &  &  &  & \checkmark &  &  &  &  &  &  &  &  &  & \checkmark &  & \checkmark &  &  &  &  &  &  & \checkmark &  &  &  &  &  &  &  & \cellcolor[HTML]{9595FF} 6 \\
			Strubell et~al. \citeyear{strubell_energy_2020} \cite{strubell_energy_2020} &  &  & \checkmark &  &  &  &  & \checkmark &  &  &  &  & \checkmark &  &  &  &  &  &  & \checkmark &  &  &  &  &  &  &  &  &  &  &  &  &  & \cellcolor[HTML]{9595FF}4 \\
			Hernandez and Brown \citeyear{hernandez_measuring_2020} \cite{hernandez_measuring_2020} &  &  &  &  &  &  &  &  &  &  &  &  &  &  &  &  & \checkmark &  &  & \checkmark &  &  &  &  &  &  &  & \checkmark &  &  &  &  &  & \cellcolor[HTML]{9595FF}3 \\
			Haakman et~al. \citeyear{haakman_ai_2021} \cite{haakman_ai_2021} &  & \checkmark & \checkmark &  &  & \checkmark & \checkmark & \checkmark &  & \checkmark &  & \checkmark &  &  & \checkmark & \checkmark &  &  & \checkmark & \checkmark &  &  &  & \checkmark &  &  & \checkmark &  &  &  & \checkmark & \checkmark &  & \cellcolor[HTML]{FF7474}15 \\
			Xu et~al. \citeyear{xu_survey_2021} \cite{xu_survey_2021} &  & \checkmark & \checkmark &  &  &  &  & \checkmark &  &  &  & \checkmark &  & \checkmark &  &  & \checkmark &  & \checkmark & \checkmark & \checkmark &  &  &  &  &  & \checkmark & \checkmark &  &  &  &  &  & \cellcolor[HTML]{FFCCCC}11 \\
			Li et~al. \citeyear{li_full-cycle_2021} \cite{li_full-cycle_2021} &  &  & \checkmark & \checkmark &  &  &  &  &  &  &  & \checkmark &  & \checkmark &  &  & \checkmark &  & \checkmark & \checkmark & \checkmark &  &  &  &  & \checkmark &  & \checkmark &  &  &  &  &  & \cellcolor[HTML]{FFEEEE} 10 \\
			Patterson et~al. \citeyear{patterson_carbon_2021} \cite{patterson_carbon_2021} &  & \checkmark & \checkmark &  &  &  &  &  &  &  &  & \checkmark &  & \checkmark &  &  & \checkmark &  &  & \checkmark & \checkmark & \checkmark &  &  &  & \checkmark &  & \checkmark &  &  &  &  &  & \cellcolor[HTML]{FFEEEE} 10 \\
			Van Wynsberghe \citeyear{van_wynsberghe_sustainable_2021} \cite{van_wynsberghe_sustainable_2021} &  & \checkmark & \checkmark & \checkmark &  &  & \checkmark &  &  &  &  & \checkmark &  &  &  &  &  & \checkmark &  &  &  &  &  &  &  &  & \checkmark &  &  &  &  &  &  & \cellcolor[HTML]{AAAAFF} 7 \\
			Bannour et~al. \citeyear{bannour_evaluating_2021} \cite{bannour_evaluating_2021} &  & \checkmark & \checkmark &  &  &  &  & \checkmark &  &  &  &  &  &  &  &  &  &  & \checkmark & \checkmark &  &  &  &  &  & \checkmark &  &  &  &  &  & \checkmark &  & \cellcolor[HTML]{AAAAFF} 7 \\
			Lannelongue et~al. \citeyear{lannelongue_green_2021} \cite{lannelongue_green_2021} &  & \checkmark & \checkmark &  &  &  &  & \checkmark &  &  &  & \checkmark &  &  &  &  &  &  &  &  &  &  &  & \checkmark &  &  &  &  &  &  & \checkmark &  &  & \cellcolor[HTML]{9595FF} 6 \\
			Cao et~al. \citeyear{cao_irene_2021} \cite{cao_irene_2021} &  &  & \checkmark &  &  &  &  &  &  &  &  &  & \checkmark &  &  &  & \checkmark &  & \checkmark & \checkmark &  &  &  &  &  &  &  &  &  &  &  &  &  & \cellcolor[HTML]{9595FF}5 \\
			Patterson et~al. \citeyear{patterson_carbon_2022} \cite{patterson_carbon_2022} &  & \checkmark &  &  &  &  &  &  &  &  &  & \checkmark &  &  &  &  & \checkmark &  & \checkmark & \checkmark &  &  &  &  &  &  &  &  &  &  &  &  &  & \cellcolor[HTML]{9595FF}5 \\
			Shaikh et~al. \citeyear{shaikh_energyvis_2021} \cite{shaikh_energyvis_2021} &  & \checkmark &  &  &  &  &  &  &  &  &  & \checkmark &  &  &  &  &  &  & \checkmark &  &  &  &  &  &  &  &  &  &  &  &  &  &  & \cellcolor[HTML]{9595FF}3 \\
			Wu et~al. \citeyear{wu_sustainable_2022} \cite{wu_sustainable_2022} &  & \checkmark & \checkmark &  &  &  & \checkmark & \checkmark &  &  &  & \checkmark & \checkmark &  &  &  & \checkmark &  & \checkmark & \checkmark & \checkmark & \checkmark &  &  &  & \checkmark &  &  &  & \checkmark &  &  &  & \cellcolor[HTML]{FF8888}13 \\
			Kaack et~al. \citeyear{kaack_aligning_2022} \cite{kaack_aligning_2022} &  & \checkmark & \checkmark & \checkmark &  &  &  & \checkmark &  &  &  & \checkmark & \checkmark & \checkmark &  &  &  & \checkmark & \checkmark & \checkmark &  &  &  & \checkmark &  &  &  & \checkmark &  &  &  & \checkmark &  & \cellcolor[HTML]{FF8888}13 \\
			Ligozat et~al. \citeyear{ligozat_unraveling_2022} \cite{ligozat_unraveling_2022} &  & \checkmark & \checkmark &  &  &  &  & \checkmark &  &  &  &  &  &  &  &  & \checkmark &  & \checkmark & \checkmark &  &  & \checkmark &  &  &  &  &  &  & \checkmark & \checkmark & \checkmark &  & \cellcolor[HTML]{FFEEEE} 10 \\
			Dodge et~al. \citeyear{dodge_measuring_2022} \cite{dodge_measuring_2022} &  & \checkmark & \checkmark &  &  &  &  &  &  &  &  & \checkmark & \checkmark &  &  &  &  &  & \checkmark & \checkmark &  &  & \checkmark & \checkmark &  &  &  & \checkmark &  &  &  &  &  & \cellcolor[HTML]{EEEEFF}9 \\
			Robbins et~al. \citeyear{robbins_our_2022} \cite{robbins_our_2022} &  & \checkmark & \checkmark &  &  &  & \checkmark & \checkmark &  & \checkmark &  &  &  &  &  &  &  & \checkmark &  &  &  &  &  &  &  &  & \checkmark &  &  &  &  &  &  & \cellcolor[HTML]{AAAAFF} 7 \\
			Georgiou et~al. \citeyear{georgiou_green_2022} \cite{georgiou_green_2022} &  &  & \checkmark &  &  &  &  &  &  &  &  & \checkmark &  &  & \checkmark &  &  &  &  & \checkmark &  &  &  & \checkmark &  &  &  &  &  &  & \checkmark &  &  & \cellcolor[HTML]{9595FF} 6 \\
			Gupta et~al. \citeyear{gupta_act_2022} \cite{gupta_act_2022} &  & \checkmark & \checkmark &  &  &  &  &  &  &  &  &  &  &  &  &  &  &  &  & \checkmark &  &  &  &  &  &  &  &  &  & \checkmark &  & \checkmark &  & \cellcolor[HTML]{9595FF}5 \\
			Verdecchia et~al. \citeyear{verdecchia_systematic_2023} \cite{verdecchia_systematic_2023} &  & \checkmark & \checkmark &  &  &  & \checkmark & \checkmark &  &  &  & \checkmark & \checkmark &  & \checkmark &  & \checkmark &  & \checkmark & \checkmark & \checkmark &  &  & \checkmark &  & \checkmark &  & \checkmark &  & \checkmark & \checkmark &  &  & \cellcolor[HTML]{FF7474}16 \\
			Zhou et~al. \citeyear{zhou_opportunities_2023} \cite{zhou_opportunities_2023} &  & \checkmark & \checkmark &  &  &  &  & \checkmark &  &  &  & \checkmark & \checkmark & \checkmark &  &  & \checkmark &  & \checkmark & \checkmark & \checkmark &  &  & \checkmark &  &  &  & \checkmark &  & \checkmark & \checkmark &  &  & \cellcolor[HTML]{FF7474}14 \\
			Menghani \citeyear{menghani_efficient_2023} \cite{menghani_efficient_2023} &  & \checkmark & \checkmark &  &  &  &  & \checkmark &  &  &  & \checkmark & \checkmark & \checkmark &  &  & \checkmark &  & \checkmark & \checkmark & \checkmark &  &  &  &  & \checkmark &  & \checkmark &  &  &  &  &  & \cellcolor[HTML]{FFAAAA}12 \\
			Chen et~al. \citeyear{chen_survey_2023} \cite{chen_survey_2023} &  &  & \checkmark &  &  &  &  & \checkmark & \checkmark & \checkmark &  & \checkmark &  & \checkmark &  &  & \checkmark & \checkmark & \checkmark & \checkmark & \checkmark &  &  &  &  &  &  & \checkmark &  &  &  &  &  & \cellcolor[HTML]{FFAAAA}12 \\
			Tornede et~al. \citeyear{tornede_towards_2023} \cite{tornede_towards_2023} &  & \checkmark & \checkmark &  &  &  &  &  &  &  &  & \checkmark & \checkmark & \checkmark & \checkmark &  &  & \checkmark &  & \checkmark &  &  &  &  &  & \checkmark &  & \checkmark & \checkmark &  & \checkmark &  &  & \cellcolor[HTML]{FFAAAA}12 \\
			Luccioni et~al. \citeyear{luccioni_estimating_2023} \cite{luccioni_estimating_2023} &  & \checkmark & \checkmark & \checkmark &  &  &  & \checkmark &  &  &  & \checkmark &  &  &  &  & \checkmark &  & \checkmark & \checkmark &  &  &  & \checkmark &  & \checkmark &  &  &  &  &  &  &  & \cellcolor[HTML]{FFEEEE} 10 \\
			Eilam et~al. \citeyear{eilam_towards_2023} \cite{eilam_towards_2023} &  &  & \checkmark &  &  &  &  & \checkmark &  &  &  & \checkmark &  & \checkmark &  &  & \checkmark &  & \checkmark & \checkmark &  &  & \checkmark & \checkmark &  &  &  & \checkmark &  &  &  &  &  & \cellcolor[HTML]{FFEEEE} 10 \\
			Li et~al. \citeyear{li_making_2023} \cite{li_making_2023} &  &  & \checkmark & \checkmark &  &  &  & \checkmark &  &  &  & \checkmark &  &  &  &  &  &  &  & \checkmark &  & \checkmark &  & \checkmark &  &  &  &  &  &  &  &  &  & \cellcolor[HTML]{AAAAFF} 7 \\
			Bouza et~al. \citeyear{bouza_how_2023} \cite{bouza_how_2023} &  & \checkmark & \checkmark &  &  &  &  & \checkmark &  &  &  &  &  &  &  &  & \checkmark &  &  &  &  &  &  &  &  &  &  &  &  &  &  & \checkmark &  & \cellcolor[HTML]{9595FF}5 \\
			Grum et~al. \citeyear{masrour_ai_2023} \cite{masrour_ai_2023} &  & \checkmark & \checkmark &  &  &  &  & \checkmark &  &  & \checkmark &  &  &  &  &  &  &  &  &  &  &  &  &  &  &  &  &  &  &  &  &  &  & \cellcolor[HTML]{9595FF}4 \\
			Barbierato and Gatti \citeyear{barbierato_toward_2024} \cite{barbierato_toward_2024} &  & \checkmark & \checkmark & \checkmark &  &  &  & \checkmark &  &  &  & \checkmark &  & \checkmark & \checkmark &  & \checkmark & \checkmark & \checkmark & \checkmark & \checkmark &  &  & \checkmark &  & \checkmark &  & \checkmark &  &  &  & \checkmark &  & \cellcolor[HTML]{FF7474}16 \\
			Bol\'on Canedo et~al. \citeyear{bolon-canedo_review_2024} \cite{bolon-canedo_review_2024} &  & \checkmark & \checkmark & \checkmark &  &  & \checkmark & \checkmark &  & \checkmark &  &  & \checkmark & \checkmark &  &  & \checkmark &  & \checkmark & \checkmark & \checkmark & \checkmark &  & \checkmark &  &  &  & \checkmark &  &  &  &  &  & \cellcolor[HTML]{FF7474}15 \\
			Longpre et~al. \citeyear{longpre_responsible_2024} \cite{longpre_responsible_2024} &  & \checkmark & \checkmark &  &  &  &  & \checkmark &  & \checkmark &  & \checkmark &  & \checkmark & \checkmark &  &  &  & \checkmark & \checkmark &  &  &  & \checkmark &  & \checkmark &  & \checkmark & \checkmark &  & \checkmark &  &  & \cellcolor[HTML]{FF7474}14 \\
			Alzoubi and Mishra \citeyear{alzoubi_green_2024} \cite{alzoubi_green_2024} &  & \checkmark & \checkmark & \checkmark &  &  & \checkmark & \checkmark &  &  &  & \checkmark &  & \checkmark &  &  & \checkmark &  & \checkmark & \checkmark & \checkmark &  &  & \checkmark &  & \checkmark &  &  &  &  &  &  &  & \cellcolor[HTML]{FF8888}13 \\
			Grum and Gronau \citeyear{grum_meaningfulness_2024} \cite{grum_meaningfulness_2024} &  & \checkmark & \checkmark &  &  &  &  & \checkmark & \checkmark &  & \checkmark &  &  &  &  &  &  &  &  & \checkmark &  &  &  & \checkmark &  & \checkmark & \checkmark & \checkmark &  &  &  &  &  & \cellcolor[HTML]{FFEEEE} 10 \\
			Wright et~al. \citeyear{wright_efficiency_2025} \cite{wright_efficiency_2025} &  & \checkmark & \checkmark &  &  &  &  &  &  &  &  & \checkmark &  &  &  &  &  & \checkmark & \checkmark & \checkmark & \checkmark &  &  &  &  & \checkmark &  &  &  & \checkmark &  & \checkmark &  & \cellcolor[HTML]{FFEEEE} 10 \\
			\midrule
			\textbf{Total per Phase} &
			\multicolumn{5}{c|}{\textbf{\cellcolor[HTML]{EAEAFF}79}} &
			\multicolumn{13}{c|}{\textbf{\cellcolor[HTML]{FF7474}134}} &
			\multicolumn{5}{c|}{\textbf{\cellcolor[HTML]{EAEAFF}85}} &
			\multicolumn{4}{c|}{\textbf{\cellcolor[HTML]{9595FF}38}} &
			\multicolumn{6}{c|}{\textbf{\cellcolor[HTML]{CFCFFF}44}} &
			\textbf{380} \\
			\textbf{Mean per Phase (in \%)} &
			\multicolumn{5}{c|}{\textbf{\cellcolor[HTML]{EAEAFF}36.7\%}} &
			\multicolumn{13}{c|}{\textbf{\cellcolor[HTML]{FF7474}24.0\%}} &
			\multicolumn{5}{c|}{\textbf{\cellcolor[HTML]{EAEAFF}39.5\%}} &
			\multicolumn{4}{c|}{\textbf{\cellcolor[HTML]{9595FF}22.1\%}} &
			\multicolumn{6}{c|}{\textbf{\cellcolor[HTML]{CFCFFF}17.1\%}} &
			\textbf{--} \\
			\midrule
			\textbf{Total per Criterion} 
			& \cellcolor[HTML]{9595FF} \textbf{0}
			& \cellcolor[HTML]{FF7474} \textbf{34} 
			& \cellcolor[HTML]{FF7474} \textbf{37} 
			& \cellcolor[HTML]{CFCFFF} \textbf{8} 
			& \cellcolor[HTML]{9595FF} \textbf{0} 
			& \cellcolor[HTML]{9595FF} \textbf{1} 
			& \cellcolor[HTML]{CFCFFF} \textbf{8} 
			& \cellcolor[HTML]{FF8888} \textbf{25} 
			& \cellcolor[HTML]{9595FF} \textbf{2} 
			& \cellcolor[HTML]{9595FF} \textbf{5} 
			& \cellcolor[HTML]{9595FF} \textbf{2} 
			& \cellcolor[HTML]{FF8888} \textbf{28} 
			& \cellcolor[HTML]{EAEAFF} \textbf{13} 
			& \cellcolor[HTML]{EAEAFF} \textbf{14} 
			& \cellcolor[HTML]{CFCFFF} \textbf{8} 
			& \cellcolor[HTML]{9595FF} \textbf{1} 
			& \cellcolor[HTML]{FFEEEE} \textbf{20} 
			& \cellcolor[HTML]{CFCFFF} \textbf{7} 
			& \cellcolor[HTML]{FF8888} \textbf{26} 
			& \cellcolor[HTML]{FF7474} \textbf{36} 
			& \cellcolor[HTML]{EAEAFF} \textbf{14} 
			& \cellcolor[HTML]{9595FF} \textbf{5} 
			& \cellcolor[HTML]{9595FF} \textbf{4} 
			& \cellcolor[HTML]{EAEAFF} \textbf{16} 
			& \cellcolor[HTML]{9595FF} \textbf{0} 
			& \cellcolor[HTML]{EAEAFF} \textbf{17} 
			& \cellcolor[HTML]{9595FF} \textbf{5} 
			& \cellcolor[HTML]{EAEAFF} \textbf{17} 
			& \cellcolor[HTML]{9595FF} \textbf{2} 
			& \cellcolor[HTML]{9595FF} \textbf{6} 
			& \cellcolor[HTML]{CFCFFF}\textbf{11} 
			& \cellcolor[HTML]{CFCFFF} \textbf{8} 
			& \cellcolor[HTML]{9595FF} \textbf{0} 
			& \textbf{380} \\
			\small Coverage per Phase (in \%) 
			& \cellcolor[HTML]{9595FF} \small 0 
			& \cellcolor[HTML]{FF7474} \small9
			& \cellcolor[HTML]{FF7474} \small10
			& \cellcolor[HTML]{CFCFFF} \small 2 
			& \cellcolor[HTML]{9595FF} \small 0 
			& \cellcolor[HTML]{9595FF} \small 0 
			& \cellcolor[HTML]{CFCFFF} \small 2 
			& \cellcolor[HTML]{FF8888} \small7
			& \cellcolor[HTML]{9595FF} \small 1
			& \cellcolor[HTML]{9595FF} \small 1
			& \cellcolor[HTML]{9595FF} \small 1
			& \cellcolor[HTML]{FF8888} \small7
			& \cellcolor[HTML]{EAEAFF} \small 3
			& \cellcolor[HTML]{EAEAFF}\small 4
			& \cellcolor[HTML]{CFCFFF} \small 2 
			& \cellcolor[HTML]{9595FF} \small 0 
			& \cellcolor[HTML]{FFEEEE} \small6
			& \cellcolor[HTML]{CFCFFF} \small 2 
			& \cellcolor[HTML]{FF8888} \small7
			& \cellcolor[HTML]{FF7474} \small9
			& \cellcolor[HTML]{EAEAFF} \small 4
			& \cellcolor[HTML]{9595FF} \small 1
			& \cellcolor[HTML]{9595FF} \small 1
			& \cellcolor[HTML]{EAEAFF}\small 4
			& \cellcolor[HTML]{9595FF} \small 0 
			& \cellcolor[HTML]{EAEAFF}\small 4
			& \cellcolor[HTML]{9595FF} \small 1
			& \cellcolor[HTML]{EAEAFF} \small 4
			& \cellcolor[HTML]{9595FF} \small 1
			& \cellcolor[HTML]{9595FF} \small 2 
			& \cellcolor[HTML]{CFCFFF} \small 3
			& \cellcolor[HTML]{CFCFFF} \small 2
			& \cellcolor[HTML]{9595FF} \small 0 
			& \small 100 \\
			\bottomrule
		\end{tabular}
	}
	\label{tab:Lifecycle}
\end{table}

\clearpage

\textbf{1. Green hardware selection and infrastructure design.}
The \emph{green hardware selection and infrastructure design} phase addresses upstream impacts often hidden in model-centric discussions but dominate full LCAs. It covers responsible sourcing of raw materials (e.g., cobalt, lithium, rare earth elements for batteries and GPUs), low-impact semiconductor manufacturing processes (e.g., extreme ultraviolet lithography with reduced water and energy intensity), and the carbon footprint of chip packaging and assembly. Empirical evidence shows that embodied emissions from mining, water fabrication, and board integration can exceed the operational footprint of smaller-scale AI deployments if amortized over short hardware lifetimes \cite{henderson_towards_2020,ligozat_unraveling_2022}. Large-scale open models such as BLOOM demonstrate that pre-use stages account for several hundred tons of CO$_2$e, underlining the non-triviality of hardware supply chains \cite{luccioni_estimating_2023}. Production concentrated in few regions (e.g., Taiwan, South Korea) creates supply-chain fragility and carbon-intensive logistics. Water-intensive fabrication in scarcity regions (e.g., Arizona, Taiwan) links hardware to local stressors beyond carbon. Green logistics targets low-carbon transport and optimized data-center siting to cut shipping distances and grid emissions \cite{gupta_act_2022}. Finally, energy-aware infrastructure levers - such as dynamic voltage and frequency scaling (DVFS), and renewable-powered data-center siting - connect embodied and operational perspectives by extending hardware lifetimes and reducing emissions during use \cite{alzoubi_green_2024}. These mechanisms anchor hardware considerations as a first-order determinant of AI's ecological footprint.

\textbf{2. Green AI development.} 
The \emph{Green AI development} phase targets Green AI in algorithmic design and training workflows, complementing hardware-focused measures. It covers low-carbon data acquisition, green model architectures, transparent documentation, and lifecycle-aware evaluation, alongside efficient training, hyperparameter tuning, fine-tuning, and carbon-conscious model compression. These activities dominate the energy footprint of AI because dataset preparation, iterative experimentation, and large-scale parameter searches are computationally intensive \cite{chen_survey_2023,zhou_opportunities_2023}. Recent work stresses the importance of embedding Green AI constraints directly into development pipelines, e.g., through energy-proportional architectures, carbon-aware scheduling of training jobs, and standardized reporting of dataset and model choices \cite{grum_meaningfulness_2024,schwartz_green_2020,longpre_responsible_2024,henderson_towards_2020}. Traditional lifecycle models often ignore the cumulative impact of repeated trials and ablation studies, although these can exceed the footprint of final model deployment. Evidence from application domains such as fintech illustrates that fast iteration cycles amplify this burden \cite{haakman_ai_2021}. Explicitly integrating Green AI principles at the development stage is therefore essential to balance accuracy gains with measurable reductions in environmental cost.

\textbf{3. Low-footprint AI task realization.} 
The \emph{low-footprint AI task realization} phase involves transferring AI models from the development environment into real-world production settings where they serve diverse end-users or operational systems. This phase covers deployment, real-time inference, on-device learning, edge computing, and fog computing, emphasizing strategies that reduce operational energy consumption and carbon emissions. During deployment, model pipelines must be integrated with monitoring infrastructures that track latency, accuracy drift, and energy consumption in real time. This operational telemetry enables adaptive scaling, targeted retraining, and selective activation of components, directly linking performance management with environmental cost control. Recent work demonstrates that coupling monitoring with energy-aware schedulers can reduce inference-related energy consumption by 15--25\% in production systems \cite{haakman_ai_2021, grum_construction_2022}. Studies highlight that such hybrid architectures yield measurable reductions in water and carbon footprints while maintaining Service Level Agreement (SLA) compliance, making deployment a decisive Green AI lever \cite{verdecchia_systematic_2023}.

\textbf{4. Circular AI maintenance and green administration.} 
The \emph{circular AI maintenance and green administration} phase emphasizes retraining models with updated data, replacing only degraded components, and reusing accelerators across projects to reduce embodied emissions. Core activities include continuous performance and drift monitoring, targeted retraining to counter dataset shift, and refactoring of inefficient model components to reduce computational demand. Maintenance also covers patching vulnerabilities, pruning redundant parameters, and recalibrating models under new workload or regulatory conditions, thereby avoiding premature redevelopment. Empirical studies show that lifecycle costs can be dominated by repeated retraining and inefficient update cycles if models are not systematically maintained \cite{patterson_carbon_2021}. Circular maintenance instead emphasizes resource-aware updates that preserve accuracy while reducing marginal energy use, e.g., selective fine-tuning on critical subsets or modular replacement of underperforming components. Moreover, transparent logging of monitoring metrics and maintenance actions enables auditable traceability of improvements, facilitating replication and compliance \cite{alzoubi_green_2024}. By maximizing reuse and adaptability, this phase mitigates the environmental burden of frequent redeployment and aligns AI practice with broader circular economy strategies.

\textbf{5. Green end-of-life and AI circularity.} 
The \emph{green end-of-life and AI circularity} phase covers retirement and material recovery of AI infrastructure, including servers, accelerators, and networking equipment. Central tasks involve standardized decommissioning, component-level sorting for rare earths, cobalt, and palladium, and refurbishment routes documented through certification schemes. Yet, empirical evidence shows that roughly 80\% of electronic waste bypasses formal recycling, handled instead in informal markets with severe ecological consequences \cite{ligozat_unraveling_2022}. AI-specific barriers include missing disclosure of bill-of-materials, inconsistent component provenance, and absent inventory systems for tracking reuse cycles. Policy-driven measures such as producer-responsibility mandates, carbon- and material-labeled devices, and integration with ISO-compliant LCA reporting could close these gaps and make embodied impacts auditable at scale.
\\
\\
\noindent \emph{4.2.2.2 Subphase tasks}\\
The following details the five high-level phases of the lifecycle model presented before.

\textbf{1.1) Responsible material sourcing.} 
Unlike traditional AI lifecycle models, which often omit upstream embodied emissions, this subphase \emph{responsible material sourcing} explicitly incorporates the environmental and carbon impacts inherently associated with resource procurement (addressing a critical gap emphasized by Gupta (\citeyear{gupta_act_2022}) and Henderson et~al. (\citeyear{henderson_towards_2020}). Embedding responsible sourcing principles at this stage enables early and proactive mitigation of the ecological footprint of AI hardware supply chains, representing a novel and necessary extension of existing frameworks.

\textbf{1.2) Low-impact manufacturing.} 
This phase involves transforming raw materials into hardware components through energy- and emission-aware production processes \cite{ligozat_unraveling_2022}. It emphasizes manufacturing innovations that substantially reduce embodied carbon and waste generation (factors that traditional AI lifecycle models frequently overlook). 
Recent LCA studies \cite{luccioni_estimating_2023, gupta_act_2022} quantify the disproportionately high Scope~3 emissions concentrated in semiconductor foundries and assembly lines, conclusively demonstrating that upstream manufacturing frequently dominates overall device footprints. These findings collectively underscore the necessity of integrating verifiable low-impact manufacturing strategies - such as renewable-powered fabrication, material efficiency, and closed-loop recycling - directly within Green AI paradigms.

\textbf{1.3) Green logistics.} 
The \emph{green logistics} subphase highlights that most accelerators and server boards are produced in East Asia and shipped to North America or Europe, creating long supply chains with concentrated carbon hotspots in maritime and aviation links. Beyond intercontinental flows, regional distribution and repeated packaging of components amplify embodied impacts that remain poorly documented in Green AI assessments \cite{ligozat_unraveling_2022}. Green practices include optimized multimodal routing, modal shifts from air to sea or rail, shipment consolidation, and the use of electric or hydrogen-powered fleets. Empirical assessments show that transport can represent a non-trivial share of embodied emissions, especially when production is concentrated in East Asia with long supply chains to North America or Europe \cite{zhou_opportunities_2023}. Incorporating logistics into lifecycle models therefore closes a persistent blind spot in Green AI analysis and highlights siting strategies that reduce both distances and grid-related operational emissions.

\textbf{1.4) Energy-aware infrastructure.} 
This phase captures the design and inference of infrastructure supporting AI hardware (such as data centers and network facilities), prioritizing renewable energy use, high energy efficiency, advanced cooling technologies, and intelligent resource management. It complements the hardware lifecycle by addressing ongoing operational impacts and Green AI integrating infrastructure-level, a critical yet insufficiently treated dimension in conventional AI lifecycle models \cite{alzoubi_green_2024, patterson_carbon_2021}.

\textbf{1.5) Incoming supply cahin transparency.} 
The \emph{Incoming Supply Chain Transparency} subphase addresses the upstream visibility of material flows, component origins, and supplier Green AI hardware manufacturing. It extends beyond standard procurement audits by enforcing traceability of raw materials such as rare earths, semiconductors, and high-carbon alloys across multi-tier supplier networks. Compliance regimes prescribe pre-production disclosure of environmental and governance data. Companies merge supplier carbon and water metrics with logistics-related Scope~3 data into life-cycle inventories and deploy digital product passports to tie components to provenance information. These linked datasets detect high-impact inputs early and supply the empirical backbone for later-stage circularity evaluation \cite{gupta_act_2022}.
\\

\textbf{2.1) Goal setting and business understanding.} 
At the \emph{goal-setting and business understanding} phase - projects define their technical and ethical scope. Infrastructure readiness, stakeholder roles, and Green AI targets are determined in concert, while potential risk, fairness, and compliance concerns are systematically addressed \cite{ligozat_unraveling_2022, mathworks_what_2024}. Embedding these guardrails from inception stabilizes downstream lifecycle execution and facilitates transparent monitoring.

\textbf{2.2) AI validation and non-AI alternative check.} 
The \emph{AI validation and non-AI alternative check} phase typically involves analyzing and evaluating how traditional systems or simpler algorithms (which may not involve ML or AI) handle the problem at hand. It is about gathering empirical evidence on the feasibility, efficiency, and limitations of existing, non-AI approaches to solving the problem, and using this information to identify where AI might be applied more effectively \cite{grum_meaningfulness_2024, ligozat_unraveling_2022}.

\textbf{2.3) Low-carbon data acquisition, collection and preparation.} 
Recognizing that data acquisition and processing contribute significantly to AI's carbon footprint \cite{verdecchia_future_2022, chen_survey_2023}, this phase focuses on sustainable practices such as selective sampling, noise reduction, and de-duplication to reduce dataset size and processing energy. Standard AI lifecycle models often treat data acquisition superficially; by embedding carbon-awareness into data pipelines, Green AI development effectively curtails unnecessary computational burdens early in the lifecycle \cite{henderson_towards_2020}.

\textbf{2.4) Green AI design modeling.} 
Designing Green AI models involves balancing computational complexity and performance, adopting energy-efficient architectures, and leveraging modular designs to facilitate reuse and compression \cite{grum_meaningfulness_2024, longpre_responsible_2024}. This architectural approach extends beyond typical AI design by embedding lifecycle emissions considerations at the conceptual stage.

\textbf{2.5) Transparent AI documentation.} 
This phase mandates comprehensive documentation of design decisions, energy consumption, and environmental impacts, thereby enabling reproducibility, accountability, and informed auditing \cite{grum_meaningfulness_2024, alzoubi_green_2024}. Transparency fosters shared community standards and supports regulatory compliance (dimensions that classical lifecycle models insufficiently emphasize).

\textbf{2.6) AI system implementation and evaluation.} 
The \emph{AI system implementation and evaluation} phase involves taking the theoretical model
and implementing it in code, creating the necessary software infrastructure, and integrating the model into the
larger technical environment. The implementation phase ensures that the AI model works effectively within its
intended operational setting. This includes setting up the required infrastructure, such as choosing the appropriate
hardware, configuring servers, and connecting the model to databases or APIs it will interact with \cite{grum_construction_2022, ligozat_unraveling_2022}.

\textbf{2.7) Efficient model training and testing.} 
Model training is computationally intensive and a major source of AI's carbon emissions \cite{chen_survey_2023, patterson_carbon_2021}. Green AI integrates training improvements such as dynamic resource allocation, adaptive batch sizing, and early stopping to reduce energy use without compromising accuracy \cite{garcia-martin_estimation_2019}. This refinement addresses a critical environmental challenge absent in traditional lifecycle definitions.

\textbf{2.8) Efficient hyperparameter tuning.} 
Hyperparameter tuning can require extensive training iterations, leading to high energy costs \cite{verdecchia_systematic_2023, zhou_opportunities_2023}. Green AI proposes approaches including multi-fidelity optimization and surrogate modeling to minimize redundant computations, representing a distinct methodological layer of Green AI integration within the AI development process.

\textbf{2.9) Green model fine-tuning.} 
Fine-tuning pre-trained models is a cost-effective alternative to full retraining but still contributes to emissions. Green fine-tuning techniques (including pruning, quantization, and task-adaptive retraining) reduce energy use while maintaining or improving performance \cite{longpre_responsible_2024, alzoubi_green_2024}.

\textbf{2.10) Lifecycle-aware AI system evaluation.} 
Evaluating AI models extends beyond accuracy metrics to include environmental impact assessments across the model's lifecycle \cite{zhou_opportunities_2023, ligozat_unraveling_2022}. This holistic evaluation facilitates informed trade-offs between Green AI considerations and performance, an innovation absent from classical AI evaluation frameworks.

\textbf{2.11) Continuous integration and delivery.} 
In the \emph{continuous integration and delivery} phase, ML models
and code are regularly integrated into shared repositories, with automated tests ensuring that updates do not
disrupt existing functionality or degrade system performance. This process helps identify issues early, reducing the risk of errors during the later stages of model deployment. In continuous delivery, once models pass automated
testing, they are automatically deployed to production environments \cite{mathworks_what_2024}.

\textbf{2.12) Carbon-aware model compression.} 
This subphase addresses pruning, quantization, and knowledge distillation techniques that cut parameter counts and FLOPs during inference \cite{verdecchia_systematic_2023,chen_survey_2023}. Empirical benchmarks show that quantization-aware training can reduce energy per inference by up to 60\% without accuracy loss, while structured pruning lowers DRAM access and interconnect traffic, reducing both power draw and latency. Compression also delays the need for high-end accelerators, decreasing embodied emissions from frequent hardware refresh. Embedding compression systematically into the lifecycle therefore couples design-level interventions with measurable carbon savings across deployment and retraining cycles.

\textbf{2.13) Responsible Green AI.} 
\emph{Responsible Green AI} integrates Green AI with broader principles of responsible AI, including fairness, privacy, accountability, transparency, human oversight, and social inclusion \cite{schwartz_green_2020}. This cross-cutting phase ensures that ecological objectives are pursued without compromising ethical obligations, recognizing that genuine Green AI requires balancing environmental, societal, and technical considerations in AI design, deployment, and governance. By embedding these values into Green AI strategies, responsible Green AI highlights that ecological gains must align with trustworthiness and human-centered AI, reinforcing compliance with emerging global guidelines such as the OECD AI Principles and the EU AI Act.
\\

\textbf{3.1) Green deployment.} 
The \emph{green deployment} phase covers the transition of AI models from development into production while explicitly minimizing the environmental footprint of integration and inference. It involves carefully selecting infrastructures (e.g., edge devices, hybrid cloud-edge) based on lifecycle-aware energy and carbon performance, optimizing software-hardware matching, and enabling adaptive resource scaling under variable loads \cite{verdecchia_systematic_2023}. Embedding Green AI criteria into deployment planning addresses a key gap in conventional lifecycles, where deployment is often reduced to a purely technical decision.

\textbf{3.2) Energy-aware inference.} 
The \emph{energy-aware inference} phase focuses on reducing the energy intensity of prediction workloads without sacrificing model accuracy. While inference typically consumes less energy per inference than training, its high execution frequency in real-world systems can result in substantial cumulative impacts \cite{verdecchia_future_2022}. Inference energy can be reduced by applying mixed-precision kernels on GPUs, dynamically deactivating unused attention heads, and adjusting batch sizes to match accelerator memory bandwidth without exceeding latency constraints. Comparative analyses of inference energy versus training emissions remain an open research challenge, requiring harmonized metrics to enable consistent benchmarking.

\textbf{3.3) Inference optimization.} 
The \emph{inference Optimization} phase minimizes energy and latency during model serving through quantization, pruning, and adaptive computation \cite{ligozat_unraveling_2022, henderson_towards_2020}. Hardware-software co-tuning leverages carbon-aware scheduling and dynamic voltage and frequency scaling to align inference workloads with low-carbon power availability \cite{patterson_carbon_2022, gupta_act_2022}. As shown by Luccioni et al. \citeyear{luccioni_estimating_2023}, sustained inference phases often dominate lifecycle energy use, reinforcing the need for continuous efficiency monitoring within deployed architectures \cite{schwartz_green_2020, patterson_carbon_2021}.

\textbf{3.4) Green AI operations.} 
\emph{Green AI operations} span runtime practices that extend beyond initial deployment, including accuracy preservation under shifting data distributions, adaptive retraining triggers, and hardware-level telemetry for thermal and power anomalies \cite{grum_construction_2022}. Central mechanisms involve lightweight monitoring pipelines coupled with DVFS-based throttling, queue-aware load balancing across heterogeneous nodes, and carbon-aware job scheduling linked to grid mix variability. By explicitly integrating these operational levers, the subphase connects reliability engineering with ecological accountability, ensuring that inference workloads remain both performant and resource-efficient across the system's full service horizon.

\textbf{3.5) Run-time, simulation and regular AI sensitivity assessment.} 
In the \emph{run-time simulation and regular AI sensitivity assessment} phase, the focus is on verifying that
the model sustains its accuracy and reliability over time in practical settings. This involves monitoring for
performance decline, detecting shifts in data distribution, and managing hardware issues. Regular evaluations
help maintain system stability and improve overall performance \cite{grum_construction_2022}.
\\

\textbf{4.1) Proactive model monitoring and scorecard creation.} 
The \emph{proactive model monitoring and scorecard creation} subphase extends conventional monitoring pipelines by coupling accuracy and latency metrics with energy, carbon, and thermal telemetry \cite{verdecchia_systematic_2023}. Runtime dashboards integrate readings from hardware counters (e.g., Running Average Power Limit, RAPL; performance monitoring counters, PMCs), rack-level meters, and carbon-intensity APIs to detect efficiency drifts under real workloads. Early-warning triggers initiate corrective actions such as adaptive batching, or workload migration to greener nodes, thereby linking operational reliability with continuous Green AI assurance throughout the inference lifecycle.

\textbf{4.2) Green compliance.} 
The \emph{green compliance} subphase operationalizes binding and voluntary frameworks by embedding environmental requirements into procurement, design, and reporting workflows \cite{alzoubi_green_2024}. 
Compliance mechanisms include third-party verified carbon disclosure protocols (e.g., Carbon Disclosure Project, CDP; Greenhouse Gas, GHG Protocol), lifecycle reporting aligned with ISO~50001, and audit trails linking AI workloads to renewable energy purchase agreements (Power Purchase Agreement, PPAs; and Renewable Energy Certificates, RECs). For data centers, verification may extend to Power Usage Effectiveness (PUE), Water Usage Effectiveness (WUE) metrics and regional grid mix disclosure. By documenting boundary choices, and allocation rules, Green Compliance establishes auditability and comparability, while reducing exposure to greenwashing claims.

\textbf{4.3) Benchmarking with alternatives (AI and non-AI techniques).} 
The focus of model benchmarking lies in identifying the most efficient models, whether AI-based or non-AI, by comparing their performance on standardized benchmarks (such as evaluating their energy consumption or training methods) \cite{grum_construction_2022}.

\textbf{4.4) Continuous improvement.} 
The \emph{continuous improvement} subphase establishes structured feedback loops where monitoring results trigger targeted interventions such as selective retraining on drifting data, parameter-efficient fine-tuning, or modular reconfiguration of components \cite{grum_construction_2022}. Improvements are validated not only against accuracy but also Performance-Environmental Thresholds (PET) / Phase Completion Criteria (PCC) thresholds, ensuring environmental objectives remain binding constraints. Techniques like low-rank adaptation, dynamic sparsification, and lifecycle-aware retraining schedules reduce overhead while preserving service quality, thereby embedding Green AI directly into the system's optimization cycle.
\\

\textbf{5.1) AI model reutilization.} 
The \emph{AI model reutilization} subphase promotes extending model lifetimes through transfer learning, fine-tuning, and distillation, thereby avoiding redundant training and embodied carbon from new model development \cite{schwartz_green_2020, patterson_carbon_2021}. Reusing pretrained architectures across tasks leverages existing compute investments while preserving empirical performance under reduced energy budgets \cite{ligozat_unraveling_2022, henderson_towards_2020}. Lifecycle-explicit accounting further links model reuse to avoided upstream emissions in data collection and hardware provisioning, establishing reutilization as a core lever for sustainable AI deployment \cite{gupta_act_2022, patterson_carbon_2022}.

\textbf{5.2) Resource recovery.} 
The \emph{resource recovery} phase targets the systematic extraction and reuse of valuable components and materials from decommissioned AI hardware \cite{gupta_act_2022}. This includes extending component lifespans through refurbishment, reclaiming rare earth elements, and reintroducing recovered parts into new systems. By reducing the need for virgin material extraction, resource recovery directly offsets upstream embodied emissions.

\textbf{5.3) Circular recycling.} 
\emph{Circular recycling} extends traditional recycling by incorporating closed-loop material flows in AI hardware manufacturing \cite{ligozat_unraveling_2022}. Component dismantling entails selective separation optimized for purity-driven material recovery, facilitating closed-loop reintegration into comparable-quality components. This reduces landfill dependency and supports lifecycle continuity by buffering production systems against primary-resource constraints.

\textbf{5.4) AI model archivation.} 
In the \emph{AI model archivation phase}, decommissioned models and associated devices are catalogued, documented, and retained within traceable repositories to preserve reproducibility and institutional memory. Following Grum, such archival practice maintains access to historic configurations and underpins sustainability through reuse-oriented resource stewardship rather than hardware obsolescence \cite{grum_construction_2022}.

\textbf{5.5) AI model disposal.} 
The \emph{AI model disposal} phase marks the terminal lifecycle stage in which obsolete hardware is eliminated, frequently through non-regulated recycling, incineration, or landfilling. Such unmanaged disposal pathways negate circularity gains and perpetuate embodied-carbon and toxicity leakage. A shift toward certified dismantling and material recovery processes is critical to integrate this phase into the broader Green-AI governance architecture. According to Luccioni et al. \citeyear{luccioni_estimating_2023}, such disposal methods pose significant environmental risks due to hazardous emissions and resource loss, making sustainable disposal strategies critical \cite{luccioni_estimating_2023}. 

\textbf{5.6) Outgoing supply chain transparency.} 
The \emph{outgoing supply chain transparency} subphase complements the upstream focus by ensuring that sustainability information propagates forward through the distribution and recovery network. After AI systems or hardware leave the manufacturer, downstream partners - distributors, operators, refurbishers, recyclers - must retain access to environmental and social performance data attached to each product instance.
This includes verified documentation of embodied carbon, repairability scores, and recyclability indices, which support responsible take-back schemes. Blockchain-anchored material passports and supplier scorecards enable traceable reverse logistics, ensuring that devices and components re-enter secondary material cycles instead of contributing to e-waste streams \cite{gupta_act_2022}.
\\

\noindent
Building on the five Green AI phases and 33 subphase tasks, section~\ref{subsubsec:ProposalofaGreenAILifecycleProcessModel} formalizes them into a process model-oriented Green AI lifecycle with explicit gateways (PCC/PET) and traceable task dependencies.

\subsubsection{Quantitative assessment of lifecycle phase representation}
\label{subsubsec:QuantitativeAssessmentofLifecyclePhaseRepresentation}

From the five Green AI phases and 33 subphase tasks, section~\ref{subsubsec:ProposalofaGreenAILifecycleProcessModel} derives a process model that encodes task sequences, boolean decision logic, and iterative Plan-Do-Check-Act (PDCA) feedback loops, providing a methodological backbone for LCA. The analysis exposes overemphasis on training and inference while provisioning, logistics, and recycling remain underexplored.

Table~\ref{tab:Lifecycle} consolidates 43 reviewed contributions, mapping upstream activities (e.g., raw material sourcing, manufacturing), core compute phases (e.g., model training, inference), and downstream activities (e.g., deployment, end-of-life). While training and inference dominate coverage across nearly all studies, upstream phases - particularly raw material extraction, component manufacturing, and transport - receive limited attention despite disproportionate Scope~3 impacts. Cross-cutting governance dimensions, including supply chain traceability, compliance auditing, and Green AI principles, emerge only sporadically, underscoring the fragmented and uneven integration of Green AI across the reviewed literature.

To assess these patterns systematically, all harmonized lifecycle terms from step~1 were aggregated into \textbf{33 subphases} grouped under \textbf{five} phases (Fig.~\ref{fig:DendrogramAILifecycle}). A binary coverage matrix (\emph{source}~$\times$~\emph{subphase}) coded whether each reviewed article addressed a subphase explicitly. This phase-level mapping (Tab.~\ref{tab:Lifecycle}) enables visual comparison of coverage breadth across studies, revealing a strong bias toward compute- and deployment-oriented subphases.

Quantitative analysis (step~6) confirms these imbalances. The \emph{mean per phase (\%)} values are:
\emph{low-footprint AI task realization}~39.5\%, 
\emph{green hardware selection and infrastructure design}~36.7\%, 
\emph{Green AI development}~24.0\%, 
\emph{circular AI maintenance and green administration}~22.1\%, and \emph{green end-of-life and AI circularity}~17.1\%.
\emph{Low-footprint AI task realization} and \emph{green hardware selection and infrastructure design} are most frequently analyzed, \emph{Green AI development} and \emph{circular AI maintenance and green administration} moderately, while \emph{green end-of-life and AI circularity} attracts little coverage. Both the \emph{total per criterion} row and the $\omega$ column confirm that impactful subphases like \emph{green compliance} and \emph{outgoing supply chain transparency} remain overlooked leverage points.

Results demonstrate entrenched focus on compute phases, contrasted by neglect of embodied impacts, recycling pathways, and governance. Advancing this from efficiency narratives to a full-spectrum Green AI framework.

The distribution of energy and material burdens concentrates on training and inference, while upstream activities such as water fabrication and transoceanic shipping, and downstream stages such as recycling or landfill disposal, remain almost entirely excluded from impact assessment.

\emph{Upstream phases} - from bauxite mining and silicon water production to accelerator board assembly - are dominated by embodied emissions tied to electricity-intensive smelting, ultrapure water use, and globalized supply chains. Case studies show that these impacts, especially rare-earth extraction and chip packaging, can outweigh runtime energy when hardware replacement intervals fall below three years \cite{patterson_carbon_2021, schwartz_green_2020}.

\emph{Core computing phases} - notably multi-terabyte dataset staging, GPU-intensive training of transformer architectures, large-scale hyperparameter sweeps, and latency-critical inference pipelines - account for the majority of direct electricity use in AI systems, with training runs often measured in megawatt-hours \cite{patterson_carbon_2021,schwartz_green_2020}. While training remains the single most energy-intensive peak activity (particularly for frontier-scale transformer models), sustained inference across heterogeneous environments (cloud clusters, embedded edge devices, and end-user hardware) often exceeds training in cumulative consumption over deployment lifetimes \cite{wu_sustainable_2022,wright_efficiency_2025,patterson_carbon_2022}. Upstream data-handling processes, especially massive preprocessing and archival storage of high-volume datasets, also impose substantial environmental burdens that are frequently underestimated in carbon accounting \cite{henderson_towards_2020,schwartz_green_2020}.

\emph{Downstream and cross-cutting phases} - including deployment pipelines, runtime inference services, continuous monitoring infrastructures, and end-of-life management of hardware - show lower energy intensity per transaction but accumulate substantial impacts due to high query volumes and persistent operation in production. End-of-life activities such as component reuse, rare-earth recovery, and certified recycling are rarely quantified with methodological rigor, yet remain decisive for mitigating embodied carbon and material scarcity \cite{ligozat_unraveling_2022,grum_meaningfulness_2024}.

\subsubsection{Proposal of a Green AI lifecycle process model}
\label{subsubsec:ProposalofaGreenAILifecycleProcessModel}

\begin{figure}[ht]
	\centering
	\includegraphics[width=0.65\textheight, angle=0]{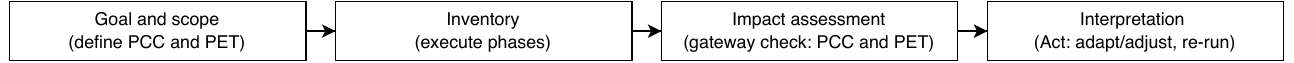}
	\caption{Level-0 LCA-style overview aligned with PCC and PET logic.}
	\label{fig:GreenAILifecycleL0}
\end{figure}

\begin{figure}[t]
	\centering
	\includegraphics[width=0.65\textheight, angle=0]{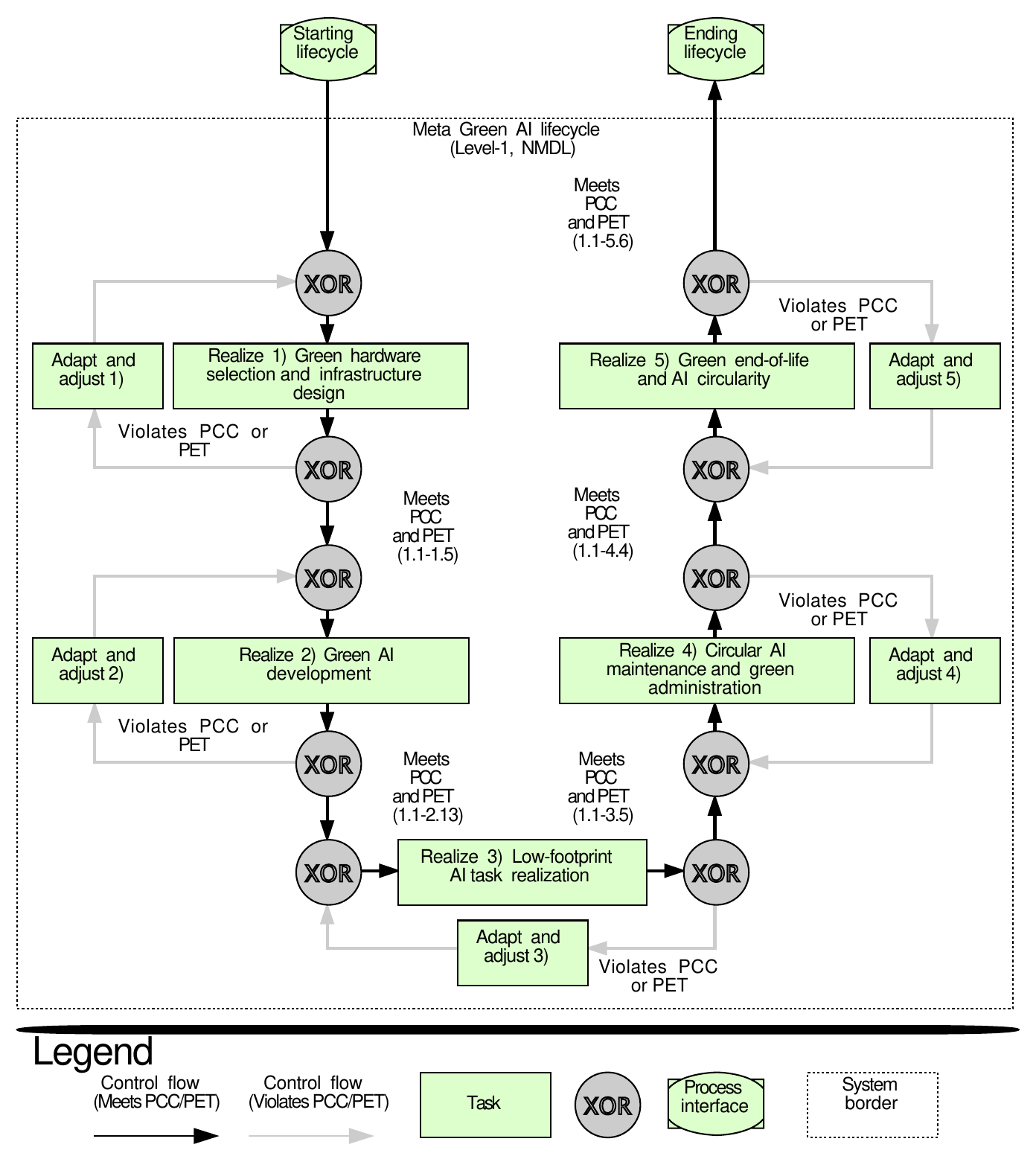}
	\caption{Meta Green AI lifecycle (Level-1, NMDL variation). Decision gateways use PCC and PET; adapt/adjust actions follow the PDCA pattern. XOR guards explicitly encode \emph{[meets PCC and PET]} or \emph{[violates PCC or PET]}.}
	\label{fig:MetaGreenAILifecycle}
\end{figure}

Section~\ref{subsubsec:StateoftheArtGreenAILifecycleStages} revealed an uneven coverage of lifecycle phases in the literature, a \emph{process-oriented Green AI lifecycle model} is proposed that integrates upstream, core compute, and downstream activities into a coherent, actionable framework. This model operationalizes the five high-level Green AI phases and 33 subphases introduced in Table~\ref{tab:LCAMap}, linking them into a structured view of the \emph{Green AI lifecycle process}.
Two complementary visualisations are provided: (1) a level-0 overview (Fig.~\ref{fig:GreenAILifecycleL0}) for a high-level LCA perspective, (2) a formal level-1 representation in Neuronal Modeling and Description Language (NMDL) (Fig.~\ref{fig:MetaGreenAILifecycle}) highlighting the top tasks (green) with the numbering from Table~\ref{tab:Lifecycle}, and (3) a formal level-2 representation in NMDL highlighting the detailed tasks (green) with the numbering from Table~\ref{tab:Lifecycle} (see Figs.~\ref{fig:RealizeGreenHardwareSelectionAndInfrastructureDesign}--\ref{fig:RealizeGreenEndoflifeAndAICircularity}).

\textbf{Need for a process-oriented Green AI lifecycle model.}
Process orientation encodes Green AI thresholds as decision gateways, where task outcomes determine adaptation of AI workflows. By formalizing lifecycle checkpoints through boolean logic, the model operationalizes decision gates that integrate technical levers - from chip-level dynamic voltage and frequency scaling (DVFS) to infrastructure siting - into a unified governance framework. This logical formalization transforms isolated optimization techniques into verifiable control structures linking environmental metrics with procedural accountability. This enables systematic evaluation of trade-offs between performance, carbon, water, and embodied emissions under methodological rules.

A key enabler for implementing such a model is the adoption of a process description language that is both formal and interpretable, capable of encoding Phase Completion Criteria (PCC) / Performance-Environmental Thresholds (PET) decision gates, phase-specific levers, and uncertainty annotations in a machine-readable structure. For this purpose, the NMDL is adopted, providing the notation for all subsequent diagrams (Fig.~\ref{fig:MetaGreenAILifecycle}).

\textbf{Background and notation of the Neuronal Modeling and Description Language (NMDL)}
The NMDL is a domain-specific, machine-readable language for modeling AI processes, their control flow, decision logic, and interfaces to organizational and technical contexts \cite{grum_construction_2022}. In this article, a minimal and interpretable subset of NMDL constructs is applied, restricted to task blocks, boolean gateways, and PET/PCC decision markers, to ensure operational usability while avoiding unnecessary notation overhead. This selection balances formal rigor with practical accessibility for Green AI applications, where reproducibility and lifecycle traceability are critical.

\textbf{Semantics.} 
Control flow in the model follows directed edges between lifecycle nodes, where transitions encode sequencing of Green AI levers (e.g., hardware sourcing, training efficiency, deployment optimization) and decision gateways route execution based on PET/PCC thresholds \cite{grum_construction_2022}. In this model, XOR gateways determine phase progression by evaluating the \emph{PCC} and, \emph{conditional on PCC being satisfied}, the \emph{Performance-Environmental Thresholds (PET)} for the current phase. 
Inputs comprise (i)~completion flags for all mandatory subphases and (ii)~measured indicators for energy, carbon, and performance. An XOR gateway has one incoming and \emph{one or more} outgoing flows. If multiple flows are present, they are labeled with mutually exclusive guard conditions (e.g., \emph{[meets PCC and PET]}, \emph{[violates PCC or PET]}), and exactly one outgoing branch is taken per execution based on the evaluation of these conditions. If only a single outgoing flow exists, the XOR functions as a controlled pass-through with an explicit guard condition. Gateways evaluate \emph{PCC first} and, only if PCC is satisfied, \emph{PET second}. Tasks consume defined inputs, produce outputs, and are evaluated against the success criteria.

\textbf{Notation rule.}
Wherever a decision point has only one valid pathway and no alternative branch is needed, the XOR gateway may be omitted and replaced by a simple direct sequence flow. In the proposed Green AI model, all guard labels are consistently standardized to the phase decision outcomes \emph{[meets PCC and PET]} and \emph{[violates PCC or PET]}, ensuring interpretability and machine-checkable conformance.

\textbf{Illustrative micro-example.} 
To lower the conceptual entry barrier for readers unfamiliar with NMDL, Fig.~\ref{fig:NMDLBasics} presents a minimal but representative example illustrating the execution logic. Here, a single task is followed by an XOR gateway that first evaluates PCC and then PET. If both are satisfied, the flow continues; otherwise, an improvement action (\emph{adapt} or \emph{adjust}) is taken before the phase is re-checked. While this example shows the simplest possible configuration, later diagrams (e.g., Fig.~\ref{fig:MetaGreenAILifecycle}) extend this logic to the full Green AI lifecycle, incorporating multiple decision points and improvement pathways.

\textbf{Why Neuronal Modeling and Description Language (NMDL)?}
While a conventional LCA process diagram can depict static relationships between phases, it lacks the ability to encode and execute conditional decision logic. NMDL extends this capability by combining formal, machine-readable semantics with practical applicability to Green AI management in AI systems \cite{grum_construction_2022}. For example, during model deployment in an environment with fluctuating grid carbon intensity, the NMDL process can automatically trigger redeployment to greener regions or rescheduling to low-carbon time windows when PET are violated \cite{kaack_aligning_2022,wright_efficiency_2025}. Such dynamic, data-driven adaptations cannot be captured or executed by static LCA diagrams, underscoring the added value of NMDL.

For readers who prefer a higher-level and more aggregated representation, a level-0 LCA-style overview (Fig.~\ref{fig:GreenAILifecycleL0}) is presented alongside the formal level-1 NMDL model (Fig.~\ref{fig:MetaGreenAILifecycle}).\emph{The correspondence between Level-0 LCA phases and level-1 Green AI tasks is summarised in Table~\ref{tab:LCAMap}.} Such dynamic, data-driven adaptations cannot be represented or operationalized through static LCA diagrams, underscoring the added value of NMDL.

Given that the overarching objective of the process model is to reduce energy consumption and CO$_2$ emissions, each detailed Green AI task functions as a measurable lever for continuous optimisation within a recursive \emph{Plan-Do-Check-Act} (PDCA) cycle \cite{kaack_aligning_2022,verdecchia_systematic_2023}. 
By spanning the entire lifecycle (from raw material extraction to governance and cross-cutting compliance), each task contributes explicitly to lowering embodied and operational impacts while preserving functional accuracy and service quality of AI systems \cite{patterson_carbon_2022,wright_efficiency_2025}.

\textbf{Phase completion and decision criteria (PCC and PET).}
In the NMDL-based Green AI lifecycle model, gateway decisions are driven by two complementary success measures: PCC and PET. Together, these criteria jointly determine the execution path after each phase and ensure that both procedural completeness and Green AI objectives are rigorously met before progressing to the next stage.

\textbf{Phase Completion Criteria (PCC).}
A phase is considered \emph{completed} when all of its mandatory subphases have been fulfilled. For example, the phase \emph{1) Green hardware selection and infrastructure design} is only deemed complete once the following subphases are successfully executed: 1.1~\emph{Responsible material sourcing}, 1.2~\emph{Low-impact manufacturing}, 1.3~\emph{Green logistics}, 1.4~\emph{Energy-aware infrastructure}, and1.5~\emph{Incoming supply chain transparency}. While these mandatory elements must be fulfilled to meet the PCC.

\textbf{Performance-Environmental Thresholds (PET).}
PET is evaluated \emph{only after} PCC is satisfied. In addition to fulfilling PCC, each task must meet predefined environmental and performance conditions. Environmental aspects include energy use, carbon footprint, and water footprint; performance aspects refer to the primary utility of the AI system (e.g., accuracy, latency, throughput) and its service quality \cite{schwartz_green_2020,henderson_towards_2020}. These conditions can be assessed qualitatively or quantitatively, depending on data availability and project maturity.

\textbf{Decision rule.}
At each gateway, the \emph{[meets PCC and PET]} branch is taken if the phase is complete according to PCC \emph{and} the PET conditions are satisfied. If either criterion is not met, the \emph{[violates PCC or PET]} branch is followed, triggering the relevant improvement actions before the phase is re-evaluated.

\textbf{Worked example (PCC/PET).}
\emph{Phase 3 - Low-footprint model deployment and inference.} 
In this example, the deployment phase of the Green AI lifecycle is evaluated according to both the Process Compliance Criteria (PCC) and the Performance Evaluation Thresholds (PET). 
All subphases - 3.1~\emph{Green deployment}, 3.2~\emph{Energy-aware inference}, 3.3~\emph{Inference optimization}, 3.4~\emph{Green AI operations}, and 3.5~\emph{Run-time, simulation and regular AI sensitivity assessment} - achieved procedural completion under the Green-AI governance cycle. 
The PET specified $E_{\mathrm{op}}\!\le\!\SI{0.50}{\joule\per\text{inference}}$ and a latency constraint of 
$\le\!\SI{100}{\milli\second}$. 
Initial measurements ($E_{\mathrm{op}}\!=\!\SI{0.78}{\joule\per\text{inference}}$, 
$\SI{95}{\milli\second}$) satisfied latency but exceeded the energy limit, violating the PET while keeping 
the PCC intact. 
In accordance with PDCA logic, a corrective mechanism was initiated through model 
\emph{quantization}, reducing arithmetic and memory overhead 
\cite{patterson_carbon_2022, wright_efficiency_2025}. 
Re-assessment verified restored compliance ($E_{\mathrm{op}}\!=\!\SI{0.48}{\joule\per\text{inference}}$, 
$\SI{97}{\milli\second}$).

\begin{table}[ht]
	\centering
	\caption{Mapping Green AI tasks to LCA phases (Level-0 vs. Level-1)}
	\label{tab:LCAMap}
	\begin{tabular}{p{0.25\linewidth}p{0.65\linewidth}}
		\toprule
		\textbf{LCA phase (Level-0)} & \textbf{Green AI task cluster (Level-1)} \\
		\midrule
		Goal and scope & Define PCC/PET; set constraints \\
		Inventory & Execute data collection, training, deployment; and subphase completion \\
		Impact assessment & Check gateways with guards on PCC and PET \\
		Interpretation & Act: adapt (external) or adjust (internal); re-run \\
		\bottomrule
	\end{tabular}
\end{table}

\clearpage

\textbf{Process pattern: PDCA with external and internal improvements.}
Building on the PCC and PET criteria in section~\ref{subsubsec:ProposalofaGreenAILifecycleProcessModel}, each Green AI lifecycle phase is implemented as a recursive PDCA cycle. This pattern embeds decision logic into an iterative improvement loop, ensuring that Green AI and performance goals are continuously monitored and achieved before advancing to the next stage. The PDCA pattern ensures each phase can be repeatedly executed and improved until both PCC and PET are satisfied. Its four steps are interpreted as follows:
\begin{enumerate} [topsep=0pt]
	\item \textbf{Plan} - Define the PCC and PET relevant for the current phase (e.g., mandatory subphases to be completed, maximum permissible energy/carbon/performance thresholds).
	\item \textbf{Do} - Execute the phase-specific tasks, producing outputs and logging performance, energy ($E$), carbon ($C$), and utilization ($U$) indicators.
	\item \textbf{Check} - Evaluate PCC and PET at a decision gateway: if both are met, take the \emph{[meets PCC and PET]} branch to the next phase; otherwise take \emph{[violates PCC or PET]} and proceed to \emph{Act}.
	\item \textbf{Act} - Apply targeted improvements: \emph{Adapt (external)} (e.g., renewable sourcing, low-carbon scheduling, greener regions) or \emph{Adjust (internal)} (e.g., pruning, quantization, hyperparameter tuning, operator fusion) \cite{kaack_aligning_2022,wright_efficiency_2025}. After the improvement, re-execute \emph{Do} and re-check PCC/PET.
\end{enumerate}

\textbf{Adapt vs. adjust by phase realization.}
This explicit separation between \emph{adapt} and \emph{adjust} ensures consistent semantics across all lifecycle phases and enables the benefits of improvements to be traced reproducibly. The PDCA structure facilitates iterative optimisation without compromising AI system performance.
As a general design decision, all process steps or rather Green AI tasks have been labeled with one of three verbs,
namely \textit{realize} as well as \textit{adapt} and \textit{adjust}.
Therefore, these are define in the following.
\begin{itemize}[topsep=0pt]
	\item \textbf{Realize.}
	The verb \textit{realizing} a task was chosen because it clarifies that
	activities to achieve or make something happen are carried out.
	For instance, when \textit{realizing raw material extraction}, (cf. Fig.~\ref{fig:RealizeGreenHardwareSelectionAndInfrastructureDesign}),
	the extraction of the raw material is achieved.
	\item \textbf{Adapt.}
	The verb \textit{adapt} a task was chosen because it clarifies that
	initial parameters, probably environmental conditions, are either changed or integrated in
	realizing a task so that the task realization works better in a new iteration.
	For instance, when the task \textit{realizing raw material extraction} has not been successful because 
	the silicon mining for hardware board production needs to much energy,
	the extraction method adapts to their environments to survive (such as by using renewable energy techniques).
	\item \textbf{Adjust.} 
	The verb \textit{adjust} a task was chosen because it clarifies that
	the own behavior is changed when realizing as task
	so that the task realization works better in a new iteration.
	The model accentuates reflexivity between tasks and their systemic context. When the task raw material extraction exceeds acceptable energy consumption - e.g., through inefficient silicon processing - production processes are recalibrated to reduce energy demand.	
\end{itemize}

The identified triad of verbs within the process model yielded a reproducible process pattern (see Fig.~\ref{fig:MetaGreenAILifecycle}), encapsulating the cyclical logic of realization, correction, and verification. 

\textbf{Illustrative execution paths.} 
The following clarifies selected execution paths to illustratively clarify how to proceed in the one or other process state.\\
Let's assume to be in \emph{Phase 1 - Green hardware selection and infrastructure design:} If subphase~1.3 \emph{green logistics} is incomplete (PCC not satisfied), the gateway takes the \emph{[violates PCC or PET]} branch. An \emph{adapt (external)} action is applied, e.g., switching to lower-emission transport or optimising routes. Once 1.3 complete and PET conditions are satisfied, the phase re-runs, transitioning via \emph{[meets PCC and PET]} to the next stage.\\
Let's assume to be in \emph{Phase 3 - Low-footprint AI task realization:} If PET is violated because the operational energy intensity $E_{\mathrm{op}}$ of deployed inference exceeds the permitted threshold, while all PCC subphases are fulfilled, the gateway takes the \emph{[violates PCC or PET]} branch. An \emph{adjust (internal)} action is applied, such as quantization or pruning to reduce load. After re-evaluation, if $E_{\mathrm{op}}$ meets the threshold and PCC remains satisfied, the process continues along the \emph{[meets PCC and PET]} branch.\\
Let's assume to be in \emph{Phase 4 - Circular AI maintenance and green administration:} If PCC is not satisfied because subphase~4.4 \emph{Continuous improvement} is incomplete, the gateway takes the \emph{[violates PCC or PET]} branch. An \emph{adapt (external)} improvement is applied by scheduling a renewable-powered maintenance window. Once the subphase complete and PET is met, the phase re-runs and transitions via \emph{[meets PCC and PET]} to subsequent lifecycle stages.

\begin{table}[ht]
	\centering
	\caption{Examples of \emph{adapt (external)} vs. \emph{adjust (internal)} across phases}
	\label{tab:adaptadjustexamples}
	\begin{tabular}{p{0.28\linewidth}p{0.32\linewidth}p{0.32\linewidth}}
		\toprule
		\textbf{Phase} & \textbf{Adapt (external)} & \textbf{Adjust (internal)} \\
		\midrule
		1) Green hardware selection and infrastructure design & Switch to renewable-powered suppliers; optimise logistics windows & Redesign PCB for low-power states; firmware power caps \\
		2) Green AI development & Schedule data pre-processing in low-carbon windows & Architecture simplification; pruning; quantization-aware training \\
		3) Low-footprint AI task realization & Carbon-aware autoscaling; placement on greener regions & quantization; operator fusion; batch-size tuning \\
		4) Circular AI maintenance and green administration & Renewable-powered maintenance windows & Retraining with early stopping; model distillation \\
		5) Green end-of-life and AI circularity & Certified recycler contracts; reverse logistics & Component reuse planning; modular design for disassembly \\
		\bottomrule
	\end{tabular}
\end{table}

\textbf{Establishing the focus on the details of the Green AI lifecycle stages}
Connected via the overview life cycle model (cf.~Fig.~\ref{fig:MetaGreenAILifecycle}),
the detailed steps and lifecycle phases can be found in the models as follows:
Fig.~\ref{fig:RealizeGreenHardwareSelectionAndInfrastructureDesign} focuses on \textit{Green hardware selection and infrastructure design}.
The \textit{Green AI development} is presented in Fig.~\ref{fig:RealizeGreenAIDevelopment1} and Fig.~\ref{fig:RealizeGreenAIDevelopment2}.
The Fig.~\ref{fig:RealizeLowFootprintAITaskRealization} draws attention to the \textit{Green AI development}.
Fig.~\ref{fig:RealizeCircularAIMaintenanceAndGreenAdministration} depicts the \textit{Circular AI maintenance and green administration}, highlighting continuous monitoring and retraining for green efficiency.
Finally, Fig.~\ref{fig:RealizeGreenEndoflifeAndAICircularity} shows the \textit{Green end-of-life and AI circularity}.

\textbf{Process interfaces connecting the models.}
Although the detailed process models (Fig.~\ref{fig:MetaGreenAILifecycle}-Fig.~\ref{fig:RealizeCircularAIMaintenanceAndGreenAdministration}) 
are connected via the model overview (Fig.~\ref{fig:RealizeCircularAIMaintenanceAndGreenAdministration}),
which becomes clear due to the same naming of the process steps of the overview model
and the respective same system border's name of the detailed process models,
neighboring detailed models are connected with each other via the \textit{process interface} modeling form.
For instance, the \textit{realize Green AI deployment} model at Fig.~\ref{fig:RealizeCircularAIMaintenanceAndGreenAdministration}
is connected with the preceded model of the \textit{realize Low-footprint AI task realization} at Fig.~\ref{fig:RealizeLowFootprintAITaskRealization}
via the two process interfaces shown on the very left.
Further, this model is connected with the downstream model labeled with \textit{realize Circular AI maintenance and green administration} at Fig.~\ref{fig:RealizeCircularAIMaintenanceAndGreenAdministration}
via the two process interfaces shown on the very right.

\textbf{Recursive wording and standardization.}
As the very same three verbs, namely (1) realize, (2) adapt and (3) adjust, 
and process pattern, namely the fix combination of boolean operators and control flow, 
have been implemented at the detailed process models
than it was used at the overview model and presented in the previous section, 
the iterative interplay of progressing or regressing detailed process steps can be found on the detailed modeling level, too. 
One can state the following:

Since this process pattern accounts for all the connected process models 
in Fig.~\ref{fig:MetaGreenAILifecycle}, Fig.~\ref{fig:RealizeGreenHardwareSelectionAndInfrastructureDesign}, Fig.~\ref{fig:RealizeGreenAIDevelopment1}, Fig.~\ref{fig:RealizeGreenAIDevelopment2}, Fig.~\ref{fig:RealizeLowFootprintAITaskRealization}, Fig.~\ref{fig:RealizeCircularAIMaintenanceAndGreenAdministration}, Fig.~\ref{fig:RealizeGreenEndoflifeAndAICircularity},
the process pattern is implemented recursively at all modeling levels.
Thus, as a \textit{design principle}, 
the five abstract Green AI phases and 33 more detailed Green AI tasks generally have been positioned at the top line of the models
that are to be \textit{realized}.
The corresponding \textit{adaption} or an \textit{adjustment} tasks can be found at the bottom line of the models next to the respective task to be realized.
These are connected via a fix control flow and boolean operator setting.

\textbf{Validation of Green AI lifecycle process model.} 
Following the formalisation of the Green AI lifecycle model using NMDL and the integration of PCC/PET-driven PDCA cycles, the proposed process model is validated through structural conformance to lifecycle boundaries, conceptual alignment with LCA categories, and empirical plausibility checks against published case studies and measurement datasets.

\textbf{(1) Structural conformance (executed).}
A structural conformance check verified that the Green AI process model satisfies the syntactic and control-flow constraints of NMDL. 
Using automated inspection combined with manual review, it was confirmed that 
(i)~all XOR gateways contain fully specified, mutually exclusive guard conditions labelled \emph{[meets PCC and PET]} or \emph{[violates PCC or PET]}, 
(ii)~no degenerate single-output XORs occur, ensuring every decision point encodes genuine alternatives, and 
(iii)~each task is connected from start to end events, preventing deadlocks or orphaned nodes. 
These properties ensure that the model is executable without control-flow errors, while preserving traceability of Green AI decisions. 
They also demonstrate consistency with the NMDL notation rules and the PCC/PET logic.

\textbf{(2) Conceptual mapping (constructed).} 
A Level-0 lifecycle overview (Fig.~\ref{fig:GreenAILifecycleL0}) and a detailed phase-to-stage matrix (Table~\ref{tab:LCAMapDetailed}) were produced to evaluate internal consistency. Each of the five lifecycle phases and 33 subphases was positioned against LCA's four stages, making explicit where AI tasks contribute to inventory data, impact levers, or interpretation checkpoints. This makes AI-specific processes auditable against established environmental categories and exposes neglected areas such as hardware provisioning and end-of-life recycling, both recognized in LCA as dominant Scope~3 emission drivers.

Three validation points emerged: 
(i)~\emph{Completeness} - the lifecycle covers provisioning to recycling without definitional gaps; 
(ii)~\emph{Alignment} - decision gates (PCC/PET) map one-to-one onto LCA assessment checkpoints; 
(iii)~\emph{Traceability} - the multi-level view makes it possible to verify NMDL decisions against familiar LCA categories, offering reproducibility and clarity for non-technical stakeholders.
%
Anchoring Green AI phases within LCA makes environmental impacts verifiable and compatible with existing reporting schemes. It ensures that AI-related activities, from workload scheduling to hardware recovery, can be benchmarked using standardized indicators. Beyond regulatory alignment, the framework enables systematic exploration of trade-offs across carbon, water, and material dimensions, extending Green AI from single-focus studies to integrated Green AI assessments.

\textbf{(3) Empirical evaluation (planned).} 
The final validation step will consist of a rigorous empirical assessment in operational Green AI workflows to complement the structural and conceptual checks described above. This stage aims to verify whether the model, when applied in diverse real-world settings and application domains, delivers consistent and measurable Green AI benefits without degrading AI system performance.
Pilot implementations are planned in two representative domains: 
(i)~\emph{industrial AI training pipelines} involving large-scale model development under varying energy/carbon constraints; and 
(ii)~\emph{cloud-based inference services} operating in regions with fluctuating grid carbon intensity. 
In both scenarios, the process model will be fully integrated, including PCC/PET decision gateways and PDCA-driven improvement loops.

Evaluation metrics will include absolute and relative changes in: 
(i)~energy consumption (kWh) and operational energy intensity (J/inference, J/training epoch), 
(ii)~carbon footprint (kg CO\textsubscript{2} emission) with location-specific emission factors and temporal variations, 
(iii)~water footprint (L, WUE) for cooling and fabrication-linked withdrawals where reliably measurable, and 
(iv)~performance indicators such as task-specific accuracy, latency under strict service-level objectives, and throughput in production workloads compared to declared baselines.

These case studies are expected to yield empirical evidence on: 
(i)~the feasibility of integrating the Green AI lifecycle model into contrasting workflows such as cloud-based large-model training and edge-level inference pipelines across diverse application domains; 
(ii)~the quantitative impact of PDCA-driven improvements on concrete Green AI metrics (e.g., reductions in J/inference, CO\textsubscript{2} emission per epoch, or water use per training run) with statistically significant effect sizes; and 
(iii)~phase-specific bottlenecks such as embodied-impact disclosure gaps, inference-time scheduling inefficiencies, or limited end-of-life recovery options, which can inform targeted refinements of the process model and establish reproducible best-practice recommendations for broader adoption.

\begin{table}[ht]
	\centering
	\caption{Mapping of the proposed Green AI lifecycle phases and subphases (Level-1) to the stages of Life Cycle Assessment (LCA, Level-0). 	This Table~\ref{tab:LCAMapDetailed} links the AI-specific process model with established Green AI assessment practice, indicating for each phase the relevant LCA categories: \emph{Goal and scope}, \emph{inventory}, \emph{impact assessment}, and \emph{interpretation}.}
	\label{tab:LCAMapDetailed}
	\scalebox{0.94}{
		\begin{tabular}{p{0.48\linewidth}p{0.52\linewidth}}
			\toprule
			\textbf{Green AI lifecycle phase (Level-1)} & \textbf{LCA stage (Level-0)} \\
			\midrule
			1) Green hardware selection and infrastruc. design & Goal and scope (define PCC/PET); Impact Assessment \\
			\quad 1.1 Responsible material sourcing & Inventory \\
			\quad 1.2 Low-impact manufacturing & Inventory \\
			\quad 1.3 Green logistics & Inventory \\
			\quad 1.4 Energy-aware infrastructure & Inventory \\
			\quad 1.5 Incoming supply chain transparency & Inventory \\
			\addlinespace[2pt]
			2) Green AI development & Goal and scope (constraints, PCC/PET); Impact Assessment \\
			\quad 2.1 Goal setting and business understanding & Inventory \\
			\quad 2.2 AI validation and non-AI alternative check& Inventory \\
			\quad 2.3 Low-carbon data acqu., coll. and prep. & Inventory \\
			\quad 2.4 Green AI design modeling & Inventory \\
			\quad 2.5 Transparent AI documentation & Inventory \\
			\quad 2.6 AI system implementation and evaluation & Inventory \\
			\quad 2.7 Efficient model training and testing & Inventory \\
			\quad 2.8 Efficient hyperparameter tuning & Inventory \\
			\quad 2.9 Green model fine-tuning & Inventory \\
			\quad 2.10 Lifecycle-aware AI system evaluation & Inventory \\
			\quad 2.11 Continuous integration and delivery & Inventory \\
			\quad 2.12 Carbon-aware model compression & Inventory \\
			\quad 2.13 Responsible Green AI & Inventory \\
			\addlinespace[2pt]
			3) Low-footprint AI task realization & Goal and scope (constraints, PCC/PET); Impact Assessment \\
			\quad 3.1 Green deployment & Inventory \\
			\quad 3.2 Energy-aware inference & Inventory \\
			\quad 3.3 Inference optimization & Inventory \\
			\quad 3.4 Green AI operations & Inventory \\
			\quad 3.5 Run-time, sim. and reg. AI sens. assessm. & Inventory \\
			\addlinespace[2pt]
			4) Circular AI maintenance and green administration & Goal and scope (constraints, PCC/PET); Impact Assessment \\
			\quad 4.1 Proactive model monit. and scorecard creat. & Inventory \\
			\quad 4.2 Green compliance & Inventory \\
			\quad 4.3 Benchm. with alter. (AI and non-AI techn.) & Inventory \\
			\quad 4.4 Continuous improvements & Inventory \\
			\addlinespace[2pt]
			5) Green end-of-life and AI circularity & Goal and scope (constraints, PCC/PET); Impact Assessment \\
			\quad 5.1 AI model reutilization & Inventory \\
			\quad 5.2 Resource recovery & Inventory \\
			\quad 5.3 Circular recycling & Inventory \\
			\quad 5.4 AI model archivation & Inventory \\
			\quad 5.5 AI model disposal & Inventory \\
			\quad 5.6 Outgoing supply chain transparency & Inventory \\
			\addlinespace[2pt]
			Decision gateways (per phase) & Impact assessment (check PCC and PET) \\
			Act: adapt (external) / adjust (internal) & Interpretation (apply improvement, re-run) \\
			\bottomrule
		\end{tabular}
	}
\end{table}

\clearpage

\subsection{Green AI hardware}
\label{subsec:GreenAIHardware}

To address RQ3 (``Which hardware components and system architectures enable improved performance per watt and reduced embodied impacts across the lifecycle?''), section~\ref{subsubsec:StateoftheArtofGreenAIHardware} synthesizes hardware used in Green AI research.
Building on this synthesis, a criteria-based selection framework is introduced that remains applicable across hardware generations.
Section~\ref{subsubsec:ContextualDeterminantsofGreenAIHardware} then makes spatial and temporal carbon-intensity factors explicit and defines a standardized reporting guideline.
Finally, section~\ref{subsubsec:IllustrativeHardwareSystems} provides illustrative, non-prescriptive system configurations that reflect common deployment classes.

\subsubsection{State-of-the-art of Green AI hardware}
\label{subsubsec:StateoftheArtofGreenAIHardware}

Energy consumption is a central determinant of overall Green AI performance, 
since training state-of-the-art models often requires megawatt-hours of electricity and sustained use of \textit{Graphics Processing Units} (GPUs) / \textit{Tensor Processing Units} (TPUs) with power draws exceeding hundreds of watts per device. 
Measurements at rack and facility level consistently reveal that training can dominate operational footprints, while inference energy, accumulated over billions of queries, may exceed training in absolute terms. 
Quantifying these demands in J/inference or kWh/epoch is therefore indispensable for comparing hardware classes, optimizing workload placement, and rigorously assessing the long-term environmental trade-offs of architectural choices.

\textit{Edge computing} leverages single-board computers (or rather embedded systems) to process data locally, reducing latency and transmission-related energy costs. 
These platforms combine compact form factors with modest power envelopes, making them suitable for distributed AI in industrial and IoT contexts. 
Typical devices include Raspberry Pi boards, Banana Pi variants, and Arduino-based microcontrollers, each optimized for different workloads. 
Empirical measurements show that a Raspberry Pi~3B with a 1.2~GHz ARM Cortex-A53 CPU consumes about 700~mW during inference, whereas an Arduino Uno R3 operates at roughly 175~mW \cite{albreem_green_2021}. 
Such differences highlight a clear trade-off between computational capability and energy efficiency: while Raspberry Pi devices can support lightweight neural networks or even quantized CNNs, micro-controllers like Arduino excel in ultra-low-power monitoring or control tasks. 
Recent advances in specialized edge accelerators (e.g., Intel Movidius NCS, NVIDIA Jetson Nano/Xavier) further expand the design space, offering GPU-class inference performance at only a few watts, thereby bridging the gap between micro-controllers and cloud-scale hardware \cite{zhu_green_2022}. 

The energy footprint of AI systems is determined not only by GPUs but also by orchestration CPUs, multi-terabyte RAM pools, and high-bandwidth storage arrays that sustain data pipelines \cite{bannour_evaluating_2021}. Empirical profiling shows that \textit{preprocessing} and \textit{feature extraction} can consume up to 30\% of node-level energy, largely invisible in GPU-centric accounts. Access to accelerators is further limited by high acquisition cost and scarce scheduling capacity, creating structural disparities between hyperscale providers and academic clusters. Consequently, \textit{hybrid CPU--GPU infrastructures} dominate HPC deployments, with CPUs managing I/O, checkpointing, and inter-node communication, while GPUs handle matrix-heavy kernels \cite{menghani_efficient_2023}. Coordinated optimization across compute, memory, and storage tiers is therefore essential, since isolated GPU efficiency gains cannot fully mitigate system-level energy demand.

In addition to GPUs, task-specific accelerators such as Google’s TPUs and modern NPUs have become pivotal for ML computation. Their matrix-oriented systolic arrays support high degrees of parallelism, drastically improving throughput and energy efficiency in both training and inference phases relative to general-purpose GPUs \cite{yokoyama_investigating_2023}. The power efficiency of TPUs and NPUs arises from their domain-specific design for matrix multiplications and low-precision arithmetic, enabling significantly higher throughput-per-watt than general-purpose GPUs. When deployed at hyperscale in data centers, TPU/NPU pods can reduce training energy by 2--5$\times$ for transformer-class models, translating into measurable cuts in CO$_2$e emissions per run \cite{jouppi_tpu_2023}. Efficiency gains matter most for transformer-based large language models and recommendation pipelines, where matrix-multiplication kernels (\texttt{GEMM}) dominate both training and inference energy use.

\textit{Low-power accelerators} like NVIDIA's Jetson modules combine ARM cores and CUDA GPUs on compact SoCs, enabling computer vision inference at 30--50 TOPS while staying under 30 W. Intel's Myriad VPU line uses fixed-function units and on-chip caches to minimize DRAM access, cutting the energy footprint of frame-level video analytics by an order of magnitude. For hyperscale deployments, Cloud TPUs offer 100+ TFLOPS per board with demonstrably higher FLOPS-per-watt ratios than flagship GPUs, providing measurable carbon savings in large-scale training runs. Yet, modern high-end GPUs such as H100 or RTX A6000 illustrate the dilemma: their unprecedented throughput requires extensive cooling and interconnect energy, escalating the total footprint of ML infrastructure in research labs and data centers.

The extreme throughput of accelerators is inseparable from high energy requirements for cooling, connections, and storage, pushing training jobs for frontier models into multi-megawatt territory and creating quantifiable Green AI risks for AI datacenters. The hardware classification Table~\ref{tab:HardwareOverview} therefore serves as an analytical baseline to compare CPUs, GPUs, TPUs, NPUs, and edge devices along energy, carbon, and lifecycle attributes, highlighting heterogeneity that directly affects footprint reporting.
It summarizes representative configurations discussed in prior studies, spanning CPUs, GPUs, and specialized AI accelerators (e.g., TPUs). 
Beyond technical specifications (cores, memory, accelerator details, thermal design power), the Table~\ref{tab:HardwareOverview} also aligns configurations with common deployment classes (local CPS level, local cloud, public cloud) to highlight ecosystem diversity rather than prescribe concrete devices.

The Table~\ref{tab:HardwareOverview} provides a systematic overview of hardware configurations frequently mentioned in the Green AI literature. 
It includes conventional CPUs and GPUs as well as domain-specific accelerators such as TPUs or NPUs. 
Beyond technical specifications (core count, memory bandwidth, GPU FLOPS, interconnect bandwidth), the classification considers hardware form factor (desktop workstation, embedded single-board computer, rack-scale accelerator), instruction-set architecture (x86, ARM, RISC-V), and deployment context (edge-level CPS nodes, private enterprise clusters, hyperscale cloud regions). 
This differentiation enables mapping diverse hardware choices directly to their energy profiles and ecological implications across all lifecycle stages.

The classification highlights trade-offs between energy efficiency, compute power, and infrastructure dependency by aligning hardware to a ``recommended level''. 
It thus underlines how different device types, from CPUs to accelerators, map to distinct lifecycle and deployment contexts.
Further, it becomes clear, that ARM-based single-board computers (e.g., Raspberry Pi, NVIDIA Jetson) are suited for low-power, local CPS scenarios, whereas x86-based servers with accelerators (e.g., V100/H100-class GPUs) are used in local or public cloud settings for performance-intensive workloads. This class-based view emphasizes trade-offs between energy efficiency, compute capacity, and infrastructure coupling, without constituting device-level recommendations.

TheTable~\ref{tab:HardwareOverview} reflects an emerging trend: comparative evaluation of hardware platforms with respect to energy tracking \cite{rojahn_conceptual_2025}. 
However, many recommendations are strongly context-dependent, varying with workload type (e.g., training vs.\ inference), deployment scale (edge, enterprise, cloud), and availability of infrastructure such as cooling, renewable supply, or high-bandwidth interconnects. 
Despite this limitation, it provides a structured baseline for comparing hardware options and identifying configurations with lower embodied and operational impacts, thereby guiding more sustainable AI system design.

Since environmental performance depends on regional and infrastructural context, section~\ref{subsec:GreenAIMeasurementTools} extends hardware profiling to whole-system monitoring, including storage arrays, interconnect traffic, and idle baseloads. Context-specific variables - grid mix dynamics, cooling technologies, and workload duty cycles - must be disclosed together with measurement data. This prevents device-level bias and supports reproducibility across sites with different infrastructures or energy mixes.

\textbf{Selection procedure (reportable and reproducible).}
\label{subsubsec:SelectionProcedure}
A device-agnostic, criteria-based framework avoids obsolescence and improves comparability across hardware generations. 
Four dimensions structure the assessment: 
(i)~technical suitability for the workload (throughput, latency, memory and I/O balance under representative load profiles), 
(ii)~platform energy efficiency (performance-per-watt and utilization sensitivity across idle-to-peak), 
(iii)~contextual carbon intensity of electricity (location-dependent, time-varying), and 
(iv)~lifecycle impacts (embodied emissions amortized over service life and utilization). 
This multi-dimensional view aligns with best practices for transparent reporting and comparability in Green AI \cite{henderson_towards_2020, anthony_carbontracker_2020}.

Thus, the following six steps have been realized:
(1)~Profile the workload (compute/memory/I/O characteristics, Service Level Agreements, SLAs; duty cycles). 
(2)~Enumerate candidate \emph{classes} (SBC/edge accelerators, workstations, on-premises servers, public cloud accelerators). 
(3)~Assess energy under representative benchmark runs; combine device telemetry with whole-system measurements (e.g., smart plugs, rack Power Distribution Units, PDUs) to capture additional non-accelerator loads \cite{anthony_carbontracker_2020}. 
(4)~Bind context explicitly: region(s), Power Usage Effectiveness (PUE), water cooling characteristics, and scheduling windows to reflect spatial and temporal variation in grid carbon intensity \cite{henderson_towards_2020}. 
(5)~Include embodied lifecycle impacts of acquisition and replacement cycles \cite{li_making_2023}. 
(6)~Report all methodological assumptions (location, temporal granularity, infrastructure characteristics, measurement method) in a standardized form to ensure reproducibility and cross-study comparability \cite{henderson_towards_2020}.

\subsubsection{Contextual determinants of Green AI hardware}
\label{subsubsec:ContextualDeterminantsofGreenAIHardware}

\noindent
Device-level efficiency alone is insufficient for Green AI assessment. 
Operational emissions depend on \emph{where} and \emph{when} workloads run. 
(1)~Regional electricity mixes can shift carbon intensity by an order of magnitude; location and emission sources must be reported. 
(2)~Temporal variation matters: carbon-aware scheduling aligns execution with renewable availability while respecting SLAs \cite{anthony_carbontracker_2020}. 
(3)~Facility overhead (PUE, cooling) contributes substantially and should be captured via whole-system measurements where feasible. 
(4)~Embodied emissions from production and replacement must complement operational metrics; longer duty cycles and modular, repairable systems reduce amortized impacts \cite{li_making_2023}. 
(5)~Data locality and I/O externalities can shift the greener option between on-premises and cloud deployments.

\textbf{Reporting guideline.}
Each decision report should include: location(s), PUE, carbon-intensity source and temporal granularity, scheduling window, measurement method (device-level vs.\ whole-system), workload profile, and embodied-carbon assumptions \cite{henderson_towards_2020}.
Standardization of these elements enables comparability across studies and preserves validity beyond specific device generations.

\textbf{Exemplary mapping.}
Rather than prescribing individual devices, hardware should be reported at the level of \emph{classes}. This mapping includes: 
(i)~low-power single-board computers or edge accelerators for latency-sensitive inference at the local CPS level; 
(ii)~workstations for mid-scale training or inference where moderate energy budgets and local availability dominate; 
(iii)~on-premises servers when control over PUE and data locality is critical; and 
(iv)~public cloud accelerators for bursty, large-scale training jobs where carbon-aware scheduling across regions can offset higher embodied emissions. The configurations in Tab.~\ref{tab:HardwareOverview} and Fig.~\ref{fig:GreenAIHardware} illustrate such classes but are not intended as prescriptive recommendations.

\textbf{Reporting standardization.} Every hardware decision should therefore include the following: location(s), PUE, carbon-intensity source, temporal granularity, scheduling window, measurement method (device-level vs. whole-system), workload profile, and embodied-carbon assumptions. Standardization of these elements is essential to make results comparable across studies and ensure Green AI assessments remain valid beyond individual hardware generations.

\subsubsection{Illustrative hardware systems}
\label{subsubsec:IllustrativeHardwareSystems}

\begin{figure}[ht]
	\centering
	\includegraphics[width=1\textwidth, angle=0]{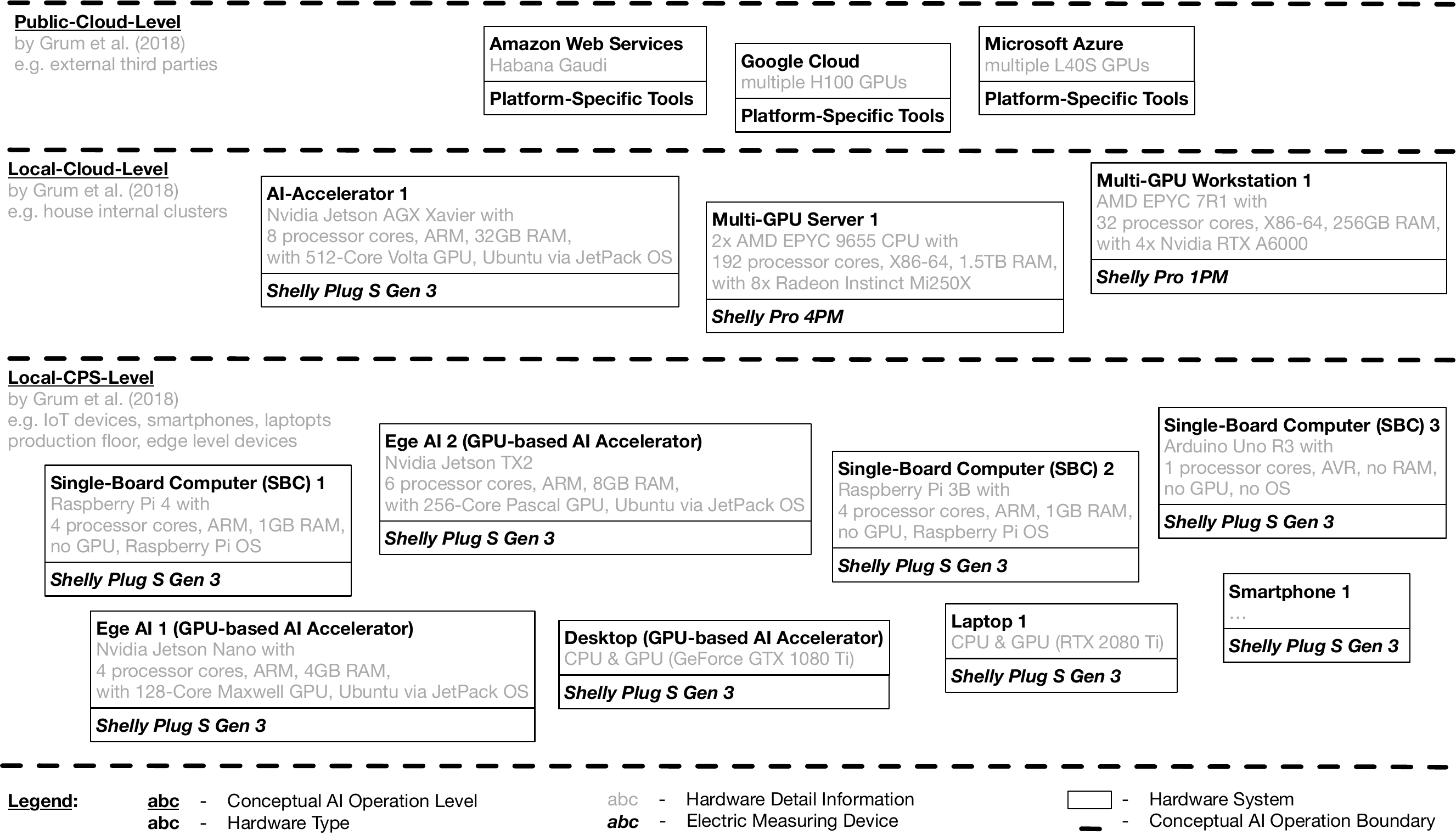}
	\caption{Exemplary hardware systems used in or attractive for Green AI research.}
	\label{fig:GreenAIHardware}
\end{figure}

Based on the Green AI hardware evaluation presented in the previous section, a graduated recommendation for Green AI hardware can be derived, that provides all heterogeneous kinds of devices, which are being researched in Green AI context (presented in Tab.~\ref{tab:HardwareOverview}).
Their arrangement in Fig.~\ref{fig:GreenAIHardware} is oriented to the computing levels of Grum et~al.\ (\citeyear{grum_decision_2018}) and follows identified levels in SLR articles as explicitly stated.

For \textit{edge-centric} or \textit{small-scale AI applications}, energy-frugal hardware platforms are pivotal. ARM-based single-board computers (e.g., Raspberry Pi) and NVIDIA's Jetson accelerators (Nano, Xavier) combine low idle power draw with integrated GPU support, enabling inference and lightweight training at a few watts rather than tens or hundreds. Thanks to their small footprint and independence from data-center cooling, TPUs and NPUs function effectively at the \emph{local CPS level}. Deploying computation near sensors reduces communication overheads and network-induced emissions, fostering device-level carbon optimization. Since consumer and Small and Medium Enterprises (SME) devices (e.g., laptops, smartphones, desktops) share comparable constraints, they can also be classified within this local CPS layer, broadening Green AI deployment beyond hyperscale infrastructures. After all, they are available in large numbers and in every company (even in many everyday situations) and can or should even be used for efficient AI realization,
potentially complemented by low-cost, WiFi smart plugs for direct power monitoring.

For medium requirements, locally operated servers with energy-efficient x86-based CPUs (e.g., Intel Xeon) 
and moderate GPUs (e.g., Tesla P100) offer a practical balance between computing power, energy consumption, operational costs, and maintainability. These can be set up as a \emph{local cloud level}. Further, powerful workstations can also be positioned on this level, as they can combine multiple non-server GPUs that feature active cooling concepts and comparatively moderate requirements on the available electric power supply infrastructure.

High-performance cloud solutions (e.g. with H100 or Habana Gaudi) are typically reserved for compute-intensive training 
if local resources are not sufficient and if their use can be justified by better hardware utilization.
As they are provided by numerous cloud service providers, they can be rented ad-hoc. 
So, they are attractive to be positioned on the \emph{public cloud level}.
Purchasing these kinds of hardware devices can be recommended only 
if you are training and hosting numerous large foundational models
or if you need lowest-latency, high-throughput inference and throughput for AI inference.
However, these server components need a specialized infrastructure to be operated, due to high requirements for active/liquid cooling in HPC and providing adequate massive electric power.
Further, their use demands for specialized software, such as using
\emph{DeepSpeed}, \emph{Megatron-LM}, \emph{TensorRT-LLM}, \emph{FSDP},
\emph{Slurm}, \emph{Kubernetes}, \emph{Airflow} and \emph{Multi-GPU Training} and \emph{Federated Learning}.

\subsection{Green AI measurement tools}
\label{subsec:GreenAIMeasurementTools}

To ensure the successful evaluation of an AI to be green, 
it is vital to have reliable and transparent tools for evaluating Green AI. 
Having the Green AI definition in mind (section~\ref{subsec:DefiningGreenAI}), 
following the Green AI lifecycle proposed (section~\ref{subsec:TransitiontotheGreenAILifecycle}), 
and operating an AI on the Green AI hardware selected (section~\ref{subsec:GreenAIHardware}), 
characterizing a Green AI requires developing robust measurement frameworks. 
Establishing such tools is essential for objectively assessing Green AI systems and guiding future innovations toward more scalable, verifiable, and greener solutions. 

To answer RQ4 (``How can environmental impacts (energy, carbon, water) be measured and reported consistently across the lifecycle, including calibration of indirect estimators with direct measurements?''), the following section contrasts estimator libraries such as \emph{CodeCarbon} with hardware telemetry from performance monitoring counters (PMCs), rack Power Distribution Units (PDUs), and facility dashboards, showing how cross-level calibration reduces systematic bias and narrows uncertainty ranges in reported footprints (section~\ref{subsubsec:StateoftheArtofMeasuringGreenAI}). This evidence base enables measurement strategies that differentiate workloads dominated by GPU compute from I/O-heavy preprocessing, distributed inference (section~\ref{subsubsec:PropositionofaGreenAIMeasurementSelection}), federated learning, composite learning or AI task allocation in distributed computing infrastructures, where trade-offs between granularity, deployment overhead, and replicability directly shape the validity of results (section~\ref{subsubsec:LimitationsoftheGreenAIMeasurementSelectionProposed}).

\subsubsection{State-of-the-art of measuring Green AI}
\label{subsubsec:StateoftheArtofMeasuringGreenAI}

The systematic review reveals that existing measurement approaches form a heterogeneous spectrum ranging from lightweight software estimators to infrastructure-grade metering. As summarized in Table~\ref{tab:MeasurementTools}, \textit{Python-based libraries} such as \emph{CodeCarbon}, \emph{Experiment Impact Tracker}, or \emph{Carbontracker} integrate directly into machine learning workflows, but rely on static emission factors and Thermal Design Power (TDP) assumptions. While these tools offer portability and low integration overhead, empirical studies show deviations of up to 40--60\% when workloads combine GPU-intensive training with I/O-heavy preprocessing or distributed inference across heterogeneous regions \cite{henderson_towards_2020, acun_carbon_2023}. \textit{Cloud-provider dashboards} (e.g., Google, Microsoft, AWS) represent a second class of indirect tools, reporting energy and carbon footprints at service and region level. Their coverage is enriched by provider-side infrastructure data, yet constrained by proprietary scope and limited transparency regarding water withdrawals, allocation rules, and embodied impacts \cite{google_llc_google_2021, microsoft_microsoft_2020}. 
In contrast to estimation approaches, \textit{direct metering methods} provide empirical ground truth across multiple abstraction layers. External meters - such as rack-mounted PDUs or smart plugs - capture total-system energy, including conversion and cooling losses, ensuring robust system-level accuracy but limited component attribution \cite{ligozat_unraveling_2022}. Their independence from specific architectures (Operating system, hardware, or software) enables consistent monitoring of heterogeneous distributed environments. Integrated \textit{performance monitoring counters} (PMCs) and on-board sensors allow fine-grained attribution at millisecond resolution but vary widely across chip types and require privileged access or cross-validation. \textit{Hybrid frameworks} (e.g., \textbf{IrEne} \cite{cao_irene_2021}) combine both paradigms, aligning model-based estimates with measured data to reduce bias. Collectively, these approaches expose the absence of a unified, cross-layer metering framework essential for Green AI assessment.

\textbf{Python packages.}
Tools such as \emph{CodeCarbon} \cite{codecarbon_codecarbon_2021}, \emph{Experiment Impact Tracker} \cite{henderson_towards_2020}, and \emph{Carbontracker} \cite{anthony_carbontracker_2020} are lightweight Python libraries that hook into ML training loops, recording runtime duration, hardware type, and regional grid mix to estimate energy use and associated CO$_2$ emissions.

Their strength lies in portability and low overhead, but methodology is inherently indirect. 
Typically, these packages: 
(i)~obtain hardware utilization (CPU/GPU time, memory) via existing ML frameworks or OS counters, 
(ii)~map this utilization to power draw using static lookup tables or manufacturer-reported TDP values, and 
(iii)~multiply energy consumption with average grid carbon intensities drawn from open datasets (e.g., IEA).

This approach rests on critical assumptions: constant efficiency factors, limited accuracy of manufacturer TDP ratings, and neglect of system-level overhead (e.g., cooling, conversion losses). 
Recent studies show such estimates can deviate significantly (up to 40--60\%) from direct power measurements, especially for heterogeneous or dynamic workloads \cite{acun_carbon_2023, patterson_carbon_2021}. 
Their reliability therefore depends strongly on workload type and local calibration.

\textbf{Online tools.}
Accessible interfaces such as \emph{EnergyVis} \cite{shaikh_energyvis_2021} and the \emph{ML Emissions Calculator} \cite{lacoste_quantifying_2019} target awareness rather than precise profiling. They typically accept model parameters (epochs, dataset size, hardware type) and return coarse-grained estimates using efficiency constants. Their value lies in communication and pedagogy, but they lack the fidelity required for system-level optimization or study-grade reporting.

\textbf{Corporate dashboards.}
Cloud-native dashboards such as Google's Carbon Footprint \cite{google_llc_google_2021} or Microsoft's Emissions Impact Dashboard \cite{microsoft_microsoft_2020} provide richer data than open-source libraries, as they access datacenter-level telemetry (cooling, Power Usage Effectiveness, PUE; and localized grid mix). 
These dashboards report aggregated, auditable emissions, often broken down by region and service class. 
However, they remain proprietary: transparency into methods limited, external validation difficult, and applicability restricted to workloads inside the provider's ecosystem. 
They cannot be generalized to hybrid or on-premises infrastructures, limiting scientific utility for reproducible research \cite{ligozat_unraveling_2022}.

\textbf{Hardware-based measurement.} 
A persistent limitation of many AI-specific tools is their lack of integration with direct hardware instrumentation, such as rack-level power meters, smart plugs (e.g. Shelly), or performance monitoring counters (PMCs) providing ground-truth energy data for CPUs and GPUs. Yet in Green IT and HPC research, external energy meters and integrated PMCs have long been standard. 

External meters (e.g., rack PDUs, smart plugs) capture full-system consumption including overhead, but lack component attribution. Integrated PMCs (used in benchmarks such as \emph{Perun} and the \emph{ML Energy Benchmark} \cite{tornede_towards_2023}) enable fine-grained, per-component analysis with minimal latency, but often require privileged access and careful calibration. 
Both approaches offer higher accuracy than indirect methods, but remain underutilized in AI-specific tooling.

\textbf{Beyond energy: water and other resources.}
Attempts to extend Green AI monitoring to water usage remain experimental. Li et~al.\ \cite{li_making_2023} propose indirect estimation of water consumption based on regional cooling requirements and datacenter energy-water conversion factors. However, these models rely on coarse averages and lack validation against direct measurements. As water becomes a critical constraint in datacenter siting \cite{zhou_opportunities_2023}, the absence of robust, tool-supported methodologies is a major research gap.

\textbf{Critical appraisal.} 
Overall, existing tools either provide accessible but coarse-grained estimates (Python packages, online tools) or detailed yet proprietary dashboards (corporate solutions). High-precision methods from Green IT (external meters, PMCs) are known but rarely integrated into AI libraries. This fragmentation hampers comparability and slows progress toward standardized benchmarks. The state of the art reflects a methodological tension: \emph{indirect but accessible} versus \emph{direct but fragmented}. Bridging these approaches (e.g., hybrid architectures combining software hooks with calibrated hardware telemetry) represents a pressing research frontier.

\textbf{Synthesis of tool coverage.}
Table~\ref{tab:MeasurementTools} further highlights the fragmented nature of the current measurement landscape. Out of 16 tools identified, Python packages dominate in terms of adoption (8/16), but offer only partial coverage of system boundaries. Corporate dashboards and online tools are fewer in number (3 and 3 respectively), yet concentrate unique access to datacenter telemetry and pedagogical communication functions. In contrast, only a small subset integrates hardware-level telemetry or embodied-impact extensions, despite their recognized importance in Green IT research \cite{patterson_carbon_2021, acun_carbon_2023, tornede_towards_2023}. The quantitative distribution therefore confirms the qualitative tension identified above: a trade-off between accessibility and methodological rigor. 

\textbf{Implications for standardization.} 
The imbalance revealed by the tool census suggests that methodological progress in Green AI measurement is currently skewed toward software-level convenience at the expense of physical accuracy. 
This imbalance has two consequences: 

First, the comparability across studies is undermined by inconsistent boundaries (device-only vs.\ whole-system, average vs.\ time-varying carbon intensity). However, since AI infrastructures can contain distributed and heterogeneous computing systems, a system-wide measurement approach is essential.
Second, important Green AI dimensions, such as water or material footprints, remain critically underrepresented \cite{li_making_2023, zhou_opportunities_2023}. 
Bridging this gap requires hybrid approaches that combine lightweight integration with calibrated hardware telemetry, further improved by open protocols for transparently reporting assumptions (location, PUE, temporal granularity). Such a synthesis would enable reproducible, cross-platform benchmarking by mandating calibration of estimator outputs against hardware meters and harmonizing emission-factor provenance, thereby aligning AI-specific tools with the validated measurement protocols long established in HPC and Green IT research.

\subsubsection{Proposal of a Green AI measurement selection}
\label{subsubsec:PropositionofaGreenAIMeasurementSelection}

Building on the critical gaps identified above, a structured selection and integration framework for comprehensive Green AI measurement is proposed. The objective is not to recommend a single tool, but to define \emph{principles and procedures} for systematically combining heterogeneous methods in a reproducible, transparent, and scientifically robust manner.

\begin{figure}[!ht]
	\centering
	\includegraphics[width=1.\textwidth, angle=0]{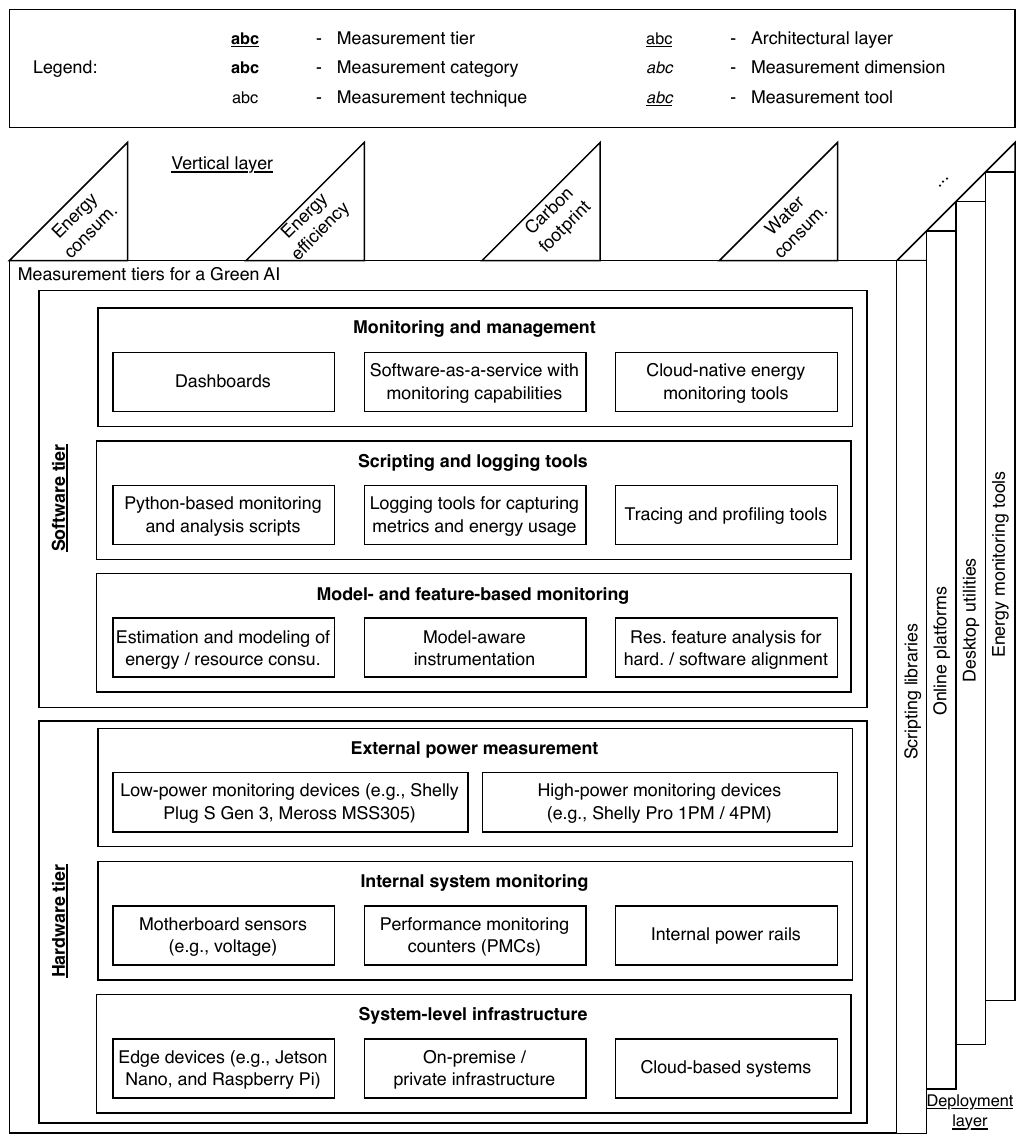}
	\caption{Green AI measurement architecture.}
	\label{fig:GreenAIMeasurementArchitecture}
\end{figure}

\textbf{Dual-layered architecture.} 
Figure~\ref{fig:GreenAIMeasurementArchitecture} illustrates a hybrid measurement architecture that integrates 
(i)~\emph{software-level estimators} for broad coverage and 
(ii)~\emph{hardware-level monitors} for calibration and validation. The software tier provides accessibility, automation, and workflow integration (e.g., CodeCarbon, Carbontracker), while the hardware tier ensures empirical fidelity (e.g., smart plugs, performance monitoring counters, PMCs; rack-level PDUs). This layered approach explicitly acknowledges the trade-off between scope and accuracy and systematically addresses the limitation of relying exclusively on indirect methods.

\textbf{Selection principles.}
A four-step procedure structures the selection and deployment of tools:
In \emph{step~1)~``define workload and scope''}, the AI workload is characterized 
(e.g. training vs.~inference, edge vs.~cloud, real-time vs.~batch) 
and the resolution required is selected (e.g. system-level totals vs.~component-level attribution).

In \emph{step~2)~``select a baseline estimator''},
the Python package or dashboard is applied to ensure coverage and reproducibility. Assumptions must be recorded explicitly (e.g. grid carbon data, emission factors, TDP values).

In \emph{step~3)~``add calibration points''},
direct hardware measurement are introduced 
(e.g. smart plugs for total load, PMCs for per-component) 
to validate and refine baseline estimates. 
When possible, cross-checks against published benchmarks, such as Perun or the ML Energy Benchmark \cite{tornede_towards_2023}, are realized.
In \emph{step~4)~``extend beyond carbon''},
additional metrics are incorporated 
(e.g. water, e-waste proxies, lifecycle amortization, resource depletion) 
where feasible. Although current water-use estimation models remain highly uncertain \cite{li_making_2023}, transparent reporting stimulates broader methodological standardization and cross-domain comparability.

\textbf{Illustrative workflow.} 
Let's assume to have a training job deployed on an on-premises GPU cluster. 
The tool called \textit{CodeCarbon} can be used to capture runtime and estimated emissions, complemented by Shelly smart plugs (low-cost, WiFi-enabled power meters) at the rack level to measure actual power draw. Integrated PMCs provide detailed profiling of GPU versus CPU contributions. Discrepancies between indirect estimates and direct measurements should be reported, and the adjusted emission factor consistently applied to subsequent runs. This workflow combines transparency (via software estimators) with methodological rigor (via calibrated hardware data).
\\
\\
\textbf{Scientific contribution.} 
The proposed framework advances the discussion from tool enumeration to \emph{methodological standardization and integration}. 
Three distinctive contributions are emphasized: 

(i) recognition of indirect-method assumptions (e.g., FLOP- or GPU-hour proxies) and their calibration requirements against empirically metered data, 

(ii) systematic incorporation of direct measurement techniques from Green IT (e.g., RAPL counters, rack-level meters, or facility telemetry) into the AI context, and 

(iii) a pathway toward multi-dimensional metrics that jointly report energy (J/kWh), carbon (kg CO$_2$ emissin with emission-factor provenance), water (L with Water Usage Effectiveness, WUE; disclosure), and embodied impacts (via Environmental Product Declarations, EPDs). 
These elements enable reproducible, comparable, and lifecycle-explicit Green AI reporting practices.

\clearpage

\subsubsection{Limitations of the Green AI measurement selection proposed}
\label{subsubsec:LimitationsoftheGreenAIMeasurementSelectionProposed}

The proposed hybrid framework mitigates fragmentation in Green AI measurement by integrating estimator tools with hardware metering, yet remains constrained by dataset availability, emission-factor heterogeneity, and limited water proxies. Limitations remain as follows.

\textbf{Reliance on indirect methods.} Software-based estimators such as CodeCarbon or Experiment Impact Tracker operate on generalized emission factors and TDP-based assumptions \cite{henderson_towards_2020, codecarbon_codecarbon_2021}. While useful for rapid deployment, their validity decreases under heterogeneous workloads or non-standard infrastructures. Calibration against hardware measurements is necessary, yet rarely performed systematically. 

\textbf{External meters vs.~integrated sensors.} External devices such as smart plugs have been used in IT for over a decade, offering simplicity and full-system coverage but limited attribution. In contrast, integrated PMCs or rack-level PDUs provide fine-grained profiling \cite{tornede_towards_2023}, but are not consistently exposed across hardware. Neither approach alone yields a complete picture.

\textbf{Incomplete metric coverage.} 
Most tools quantify carbon emissions but omit cooling-related water withdrawals, WUE disclosure, and traceable e-waste recovery rates, leaving major lifecycle burdens outside their reporting scope. Recent attempts to estimate water consumption rely on indirect cooling models \cite{li_making_2023}, but lack validation across diverse data centers. This omission constrains the comprehensiveness of current architectures.

\textbf{Fragmentation and comparability.} Outputs often remain incompatible due to inconsistent emission factors, regional datasets, and reporting formats \cite{verdecchia_systematic_2023}. 
Thus, studies are not comparable, undermining cross-benchmarking. Standardized reporting formats and disclosure of assumptions would improve reproducibility.

\textbf{Lifecycle blind spots.} Few tools capture embodied impacts from semiconductor fabrication, assembly, and logistics, or account for reuse and recycling, even though LCA studies show embodied emissions from mining and wafer production can outweigh savings across a system's lifetime \cite{patterson_carbon_2021,luccioni_estimating_2023}. Without harmonized methods for amortizing embodied carbon, reporting risks focusing disproportionately on short-term operational footprints.

In summary, the current measurement landscape remains fragmented, carbon-centric, and assumption-driven. 
Resolving these discrepancies necessitates advances in three complementary areas: first, calibration procedures that align software estimators with hardware-based truth data; second, an expanded indicator framework covering water, e-waste, and lifecycle material dimensions; and third, interoperable reporting formats ensuring cross-system comparability. These measures jointly underpin a reproducible and scalable architecture for comprehensive Green-AI evaluation.

\textbf{Methodological differentiation.} 
Recent comparative studies \cite{verdecchia_systematic_2023} highlight that measurement reliability in Green AI research is not intrinsic to data but contingent upon methodological precision. This dependency reaffirms the centrality of explicit boundary-setting, transparent calibration, and cross-layer verification as foundations of credible Green-AI assessments. Table~\ref{tab:MeasurementMethods} systematizes existing approaches along two dimensions: \emph{indirect vs.\ direct measurement} and \emph{system- vs.\ component-level granularity}.

\begin{table}[ht]
	\centering
	\caption{Comparison of measurement methods for Green AI assessment.}
	\begin{tabular}{L{3,5cm} L{1,7cm} L{1.9cm} L{2.9cm} L{3,9cm}}
		\toprule
		\textbf{Method} & \textbf{Type} & \textbf{Granularity} & \textbf{Strengths} & \textbf{Limitations} \\
		\midrule
		\midrule
		Software-based estimators (CodeCarbon, Experiment Impact Tracker) 
		& Indirect (model-based) 
		& System-level (approx.) 
		& Easy integration; portable; minimal setup		
		& Relies on emission factors; low accuracy under heterogeneous loads \cite{henderson_towards_2020, codecarbon_codecarbon_2021} \\
		\midrule
		Cloud dashboards (Google, Microsoft, Amazon) 
		& Indirect (provider-reported) 
		& Data center-/region-level 
		& Access to provider infrastructure data; automated aggregation 
		& Proprietary scope; limited transparency; inapplicable beyond providers \cite{google_llc_google_2021, microsoft_microsoft_2020} \\
		\midrule
		External meters (smart plugs, rack PDUs) 
		& Direct (whole-system) 
		& Device-/rack-level 
		& Tradition in Green IT; captures total system incl. cooling.		
		& Cannot isolate components; susceptible to conversion inefficiencies \\
		\midrule
		Integrated sensors (PMCs, motherboard sensors) 
		& Direct (component-level) 
		& CPU/GPU, memory, accelerator 
		& High temporal resolution; fine-grained attribution 
		& Requires technical expertise; inconsistent availability across hardware \cite{tornede_towards_2023} \\
		\midrule
		Hybrid approaches (software and external/internal meters) 
		& Combined 
		& Multi-tier (system and component) 
		& Balances portability and precision; enables method calibration.		
		& Rarely standardized; higher setup effort; limited tool support \cite{cao_irene_2021, verdecchia_systematic_2023} \\
		\bottomrule
	\end{tabular}
	\label{tab:MeasurementMethods}
\end{table}

Python-based or rather \textit{software-based packages} (e.g., \emph{CodeCarbon}, \emph{Experiment Impact Tracker}) represent indirect, model-driven estimation. They integrate seamlessly into machine learning pipelines but rely on generalized emission factors and hardware assumptions, leading to potential inaccuracies under heterogeneous workloads \cite{henderson_towards_2020, codecarbon_codecarbon_2021}. Cloud-provider dashboards (e.g., Google, Microsoft, AWS) similarly rely on indirect reporting, though enriched by infrastructure-level data; however, they remain constrained to proprietary ecosystems \cite{google_llc_google_2021, microsoft_microsoft_2020}. 

In contrast, \textit{direct measurement} has a long tradition in Green IT. External devices such as smart plugs or rack-mounted PDUs capture whole-system consumption including conversion losses and cooling overhead, providing robust upper bounds but lacking component-level attribution. Integrated sensors such as PMCs or motherboard energy probes enable fine-grained attribution to CPUs, GPUs, and memory, though their availability and calibration differ significantly across hardware generations \cite{tornede_towards_2023}. 

\textit{Hybrid methods} combining indirect estimators with external or internal direct meters offer a promising compromise by calibrating software-based models against physical measurements. Such hybridization has been demonstrated in frameworks like IrEne \cite{cao_irene_2021}, where software estimators are continuously calibrated with rack-level meters. Recent reviews \cite{verdecchia_systematic_2023} stress, however, that the absence of shared calibration protocols and emission-factor disclosure keeps these methods difficult to compare across studies.

This methodological differentiation shows that portability favors lightweight estimators, precision requires hardware metering, and transparency depends on disclosure of emission factors and calibration errors, making tool choice inseparable from granularity and deployment context.

\begin{figure}[ht]
	\centering
	\includegraphics[width=1.\textwidth]{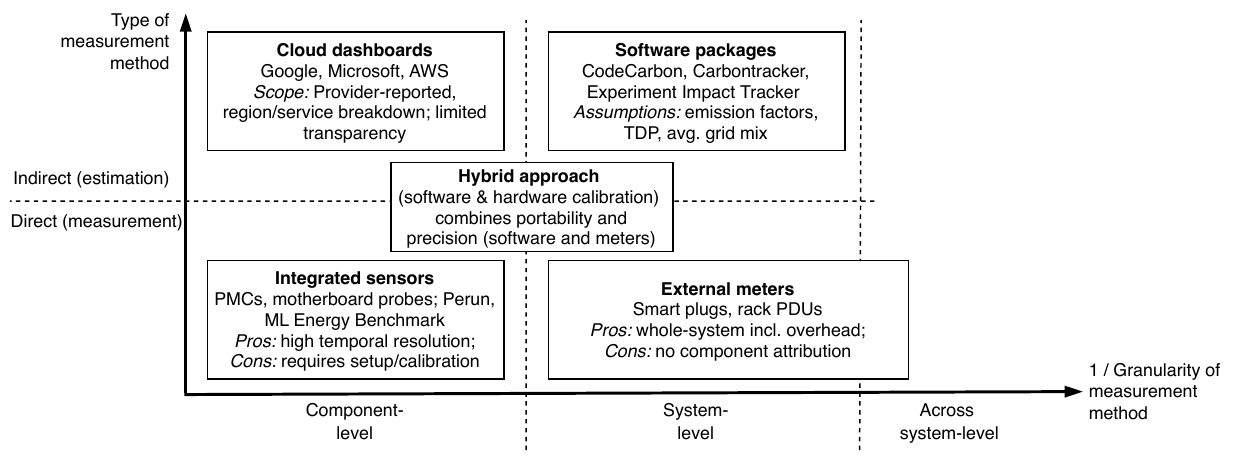}
	\caption{Conceptual mapping of Green AI measurement along two axes: 
	(indirect vs.\ direct) and (system- vs.\ component-level granularity). 
	Software packages (e.g., \emph{CodeCarbon}, \emph{Carbontracker}) rely on emission factors and grid averages \cite{henderson_towards_2020, anthony_carbontracker_2020}. 
	Cloud dashboards (Google, Microsoft, Amazon) report region- and service-specific data with limited transparency \cite{google_llc_google_2021, microsoft_microsoft_2020}. 
	External meters (e.g., smart plugs, rack PDUs) capture whole-system power but lack attribution. 
	Integrated sensors (PMCs, benchmarks like \emph{Perun}, \emph{ML Energy Benchmark}) provide fine-grained data but require calibration \cite{tornede_towards_2023}. 
	Hybrid approaches combine software with calibrated hardware telemetry, balancing portability and precision \cite{cao_irene_2021, verdecchia_systematic_2023}.}
	\label{fig:GreenAIMeasurementApproaches}
\end{figure}

As summarized in Fig.~\ref{fig:GreenAIMeasurementApproaches}, these methods can be systematically categorized along two axes: indirect vs.\ direct measurement, and component-level vs. \ system-/across-system-level granularity. It so provides a conceptual mapping of representative tools, highlighting their respective assumptions, data sources, and calibration requirements.

While the Fig.~\ref{fig:GreenAIMeasurementApproaches} provides valuable structural clarity, their practical implications go considerably further. 
\textit{Indirect system-level estimators} (e.g., \emph{CodeCarbon}, \emph{Experiment Impact Tracker}) often underestimate cooling and data-movement overheads, leading to 40--60\% deviation from metered values in heterogeneous workloads \cite{acun_carbon_2023, henderson_towards_2020}. This discrepancy is most evident when workloads combine GPU-dominated phases with I/O-heavy preprocessing or distributed training across regions with differing grid mixes \cite{patterson_carbon_2021, schwartz_green_2020}. Without systematic calibration against ground-truth data, such tools risk underestimating non-compute overheads (cooling, conversion losses) that facility-level audits consistently flag as major Scope~2 contributors \cite{anthony_carbontracker_2020}.

\textit{Direct measurements} provide higher fidelity but entail their own challenges. 
\textit{External meters}, such as rack PDUs or smart plugs (\emph{Shelly}, APC), capture end-to-end consumption, including cooling and idle overhead, thereby aligning with Green IT practice \cite{ligozat_unraveling_2022}. 
However, their granularity is insufficient for attributing consumption to CPUs, GPUs, or accelerators. 
Conversely, integrated PMCs and motherboard sensors allow fine-grained attribution at millisecond resolution, as demonstrated by benchmarks like \emph{Perun} and the \emph{ML Energy Benchmark} \cite{tornede_towards_2023}. Integrated sensors expose heterogeneous registers and sampling rates across x86, ARM, and accelerator platforms, with vendor-specific APIs (e.g., Intel RAPL, NVIDIA NVML, ARM PMU) lacking interoperability. The absence of unified access protocols or reference calibration datasets prevents reproducible measurements across HPC, edge, and cloud deployments \cite{verdecchia_systematic_2023}. A pronounced research gap persists: no unified direct or indirect measurement approach currently spans multiple system levels. Due to their system-level orientation, external meters are particularly attractive for heterogeneous, distributed infrastructures, where they offer architecture-independent energy data collection. Yet, the absence of a cross-layer integration model prevents full traceability between component, node, and facility levels, impeding comprehensive lifecycle accounting.

\textit{Hybrid approaches}, such as the Frameworks IrEne \cite{cao_irene_2021}, show how estimator outputs can be corrected with hardware telemetry, reducing systematic bias while retaining portability. Recent proposals for cross-layer integration - e.g., combining Python hooks with rack-level smart plugs and grid-aware scheduling - show calibration can cut error margins below 10\% while preserving workflow usability \cite{rojahn_conceptual_2025}. Yet no consensus exists on protocols for calibration frequency, emission-factor provenance, or uncertainty disclosure. This gap highlights the need for open benchmarks that systematically evaluate estimator-meter combinations across contexts (training, inference, preprocessing, storage), coupled with transparent reporting of assumptions such as PUE, WUE, and embodied-carbon amortization \cite{li_making_2023, zhou_opportunities_2023}.

\textbf{Attempts for coherentism.}
Taken together, the methodological differentiation reveals that portability, precision, and transparency cannot be optimized simultaneously: software-only tools favor accessibility but lack rigor, hardware-only approaches maximize accuracy but limit comparability, and hybrid solutions remain under-standardized. 
Advancing Green AI measurement thus requires moving beyond isolated tools toward integrated, auditable pipelines that combine estimator scalability with metered calibration. 
Only then can reported metrics achieve the reproducibility and cross-platform comparability needed for lifecycle-explicit Green AI assessments.

\section{Challenges in implementing Green AI and future research directions}
\label{sec:ChallengesInImplementingGreenAIAndFutureResearchDirections}

This section synthesizes the challenges and research gaps identified in the systematic review, extending the conceptual structure developed in the previous chapter. In line with RQ5 (``Which open challenges and research priorities (benchmarks, reporting schemas, uncertainty handling, circularity, governance) are needed to advance lifecycle integration of Green AI?''), the focus lies on identifying methodological blind spots - such as calibration protocols, embodied-impact accounting, and multi-metric integration - and on outlining concrete pathways for advancing reproducible, lifecycle-explicit Green AI measurement.

\subsection{Challenges and research gaps in Green AI}
\label{subsec:ChallengesandResearchGapsinGreenAIConcepts}

\noindent 

\textbf{Green approaches to AI task allocation.} 
Following the \emph{Green AI definition} (section~\ref{subsec:DefiningGreenAI}), Green AI requires not only efficiency but adaptive task distribution across diverse infrastructures. Edge and on-device deployments mitigate network-transfer emissions yet remain bounded by limited accelerator memory and inconsistent power-monitoring capabilities \cite{verdecchia_systematic_2023, jouppi_tpu_2023}. Advancing Green AI therefore entails intelligent, multi-criteria workload orchestration - balancing environmental cost, latency, and resource utilization - to realize verifiable lifecycle optimization across distributed systems \cite{grum_decision_2018}.

\textbf{Flexible resource allocations.} 
Considering the \emph{Green AI lifecycle} (section~\ref{subsec:TransitiontotheGreenAILifecycle}), 
future AI systems must become more responsive to operational variability, dynamically adjusting their energy usage and emissions output as demand changes. Many existing systems still lack this kind of adaptive resource management. Future research should target algorithmic solutions that allow AI to modulate its computing load (such as intelligently migrating processes to energy-efficient platforms or activating scaling mechanisms based on task demand). Given the varied nature of AI applications, such strategies must also accommodate different urgency levels and resource intensities. Real-time analytics may demand immediate processing, whereas other tasks could be scheduled for off-peak execution on green hardware \cite{grum_decision_2018}. This kind of adaptive approach must carefully balance ecological objectives with business imperatives. As noted in \cite{grum_meaningfulness_2024}, practical Green AI must reconcile Green AI goals with real-world constraints on performance and availability. Improving resource efficiency requires reducing redundant gradient computations during training, minimizing checkpoint overhead, and leveraging mixed-precision arithmetic, which collectively cut energy per epoch without degrading accuracy.

\textbf{Heterogeneous (computing) networks.}
The diversification of platforms - from general-purpose CPUs and GPUs to TPUs and low-power edge accelerators - offers concrete levers for energy reduction by aligning workloads with architectural strengths. CPUs handle control and I/O-heavy preprocessing, GPUs accelerate dense linear algebra, while TPUs and NPUs excel in high-throughput tensor operations. Edge devices, in turn, reduce data-movement overhead when inference is performed locally. What remains missing is a scheduling framework that exploits these complementarities at runtime, coordinating placement decisions across architectures to minimize joules per task and reduce total embodied emissions \cite{anzt_high_2022, zhu_green_2022}.
Particularly in regard with unused standard computing systems, as these are available in SME and everyday situations, such as laptops, desktops, smartphones and smart home devices, one can assume a high potential for establishing innovative Green AI approaches.

\textbf{Carbon- and energy-aware distribution.} 
Although real-time energy monitoring via mainboard sensors and performance monitoring counter (PMC)-based integration has been demonstrated, systematic validation against rack-level meters is missing, leaving large error margins in heterogeneous workloads unaddressed. In direct continuity with the methodological gaps identified for Green AI measurement (section~\ref{subsec:GreenAIMeasurementTools}), the research frontier now shifts toward scalable, adaptive frameworks that allocate AI workloads not only by efficiency, but also by \emph{carbon intensity of the energy source}. This includes directing computations to data centers powered by renewable energy or offloading tasks to energy-efficient edge devices \cite{acun_carbon_2023}. A key underexplored strategy is \emph{carbon-aware scheduling} - deferring or relocating tasks according to temporal and geographic variations in grid emissions \cite{radovanovic_carbon-aware_2023,acun_carbon_2023,henderson_towards_2020}. Such scheduling mechanisms are well established in Green IT and HPC research, where job schedulers optimize workloads based on energy availability and cost signals \cite{anzt_high_2022, li_towards_2024}. Their systematic integration into AI workflows would bridge the current gap between Green IT practices and Green AI-specific efforts. Beyond electricity, future extensions must systematically account for water and material footprints, explicitly incorporate environmental, social, and technical trade-offs, aligning workload allocation with broader Green AI constraints \cite{li_making_2023, zhou_opportunities_2023}. Beyond carbon, region-specific water stress indicators and embodied-impact amortization can be added in allocation, prioritizing regions with lower cooling water intensity and longer hardware duty cycles \cite{li_making_2023, zhou_opportunities_2023}.

\textbf{Operationalization of carbon-aware scheduling.}
Effective carbon-aware scheduling requires (i)~short-term carbon-intensity forecasts at the selected regions, 
(ii)~explicit deferral windows derived from Service Level Agreements (SLAs), 
and (iii)~migration policies that minimize data movement.
A pragmatic baseline aligns non-urgent training with low-intensity windows and pins latency-sensitive inference 
to regions with consistently cleaner grids, while reporting the allowed time-shift and regional scope.
Such operational guardrails make carbon-aware scheduling auditable and reproducible \cite{henderson_towards_2020}. In practice, relocation policies must account for data gravity and compliance constraints 
(e.g., egress costs, data residency), and report any non-migratable workloads 
and privacy-driven pinning decisions. A 12-24\,h deferral window aligned with day-ahead forecasts unlocks 
double-digit percentage emission reductions for non-urgent training, without affecting 
latency-bound inference.

\textbf{Reporting checklist for carbon-aware studies.}
Each study should disclose: regions and data sources for carbon intensity (including temporal resolution), 
explicit deferral windows and associated SLA constraints (e.g., latency bounds, maximum postponement), 
migration policies and data-locality assumptions, 
telemetry sources (device vs.\ whole-system) and calibration error, 
as well as the multi-objective weights (carbon, water, embodied impacts).
Standardized disclosure enables cross-study comparability and improves the transfer of Green IT practices 
to AI-specific settings \cite{henderson_towards_2020}. Studies should further disclose whether a single weighted objective or a Pareto analysis was used to trade off carbon, water, and embodied impacts.

\textbf{Bridging to Green IT and HPC scheduling.}
Mature techniques from Green IT/HPC (such as power capping (e.g., Running Average Power Limit, RAPL), dynamic voltage and frequency scaling, DVFS-based throttling, 
energy-aware job placement, and queue-level energy budgets) are directly transferable to AI pipelines.
Adapted to AI, these mechanisms inform cluster-level placement (CPU/GPU/TPU bins), 
coordinate batch sizes and activation checkpointing under power caps, 
and co-schedule data locality with accelerator availability to reduce both energy and egress overhead 
\cite{anzt_high_2022, li_towards_2024}. Preemption and periodic checkpointing enable safe deferral within carbon windows, 
bounding rollback overhead while minimizing energy, carbon, and resource impacts and preserving SLA commitments.

\textbf{Closed-loop scheduling.}
Direct telemetry (PMCs, rack Power Distribution Units, PDUs) should calibrate software estimators at deployment time and feed back into 
the scheduler to update placement and deferral policies. Reporting the pre/post calibration error - for instance, the deviation between CodeCarbon estimates and rack-level power meter readings - and the resulting change in workload allocation or siting decisions aligns AI-specific practice with established Green IT validation routines, thus improving transparency, reproducibility, and decision quality \cite{tornede_towards_2023,henderson_towards_2020}.

\textbf{Critical research gap.}
Existing studies remain fragmented: FLOP- or GPU-hour proxies are used without validating against metered baselines, telemetry data lacks harmonized calibration protocols, and carbon-intensity traces are seldom integrated into workload schedulers. What is missing is a benchmarked framework that couples workload profiling, ground-truth metering, and carbon-aware dispatch into reproducible allocation strategies across cloud, edge, and on-premises infrastructures.Building this bridge is a central research challenge for advancing Green AI from conceptual discussions to operational practice.

\subsection{Operational pathways toward lifecycle-explicit Green AI}
\label{subsec:OperationalPathwaysTowardLifecycle-ExplicitGreenAI}

Despite obstacles outlined earlier, significant potential remains for advancing scalable implementation of Green AI. This review demonstrates how (i)~instrumentation at rack and facility level (e.g., PDUs, PMCs, smart plugs) can be directly coupled with AI-specific workload managers using DVFS and RAPL-based capping, (ii)~multi-objective benchmarks can be established that jointly report carbon intensity, cooling-water withdrawals, and embodied hardware emissions with declared provenance, and (iii)~closed-loop calibration protocols can align estimator libraries with ground-truth meters, enabling auditable siting and scheduling across heterogeneous infrastructures.

\textbf{Extend lifecycle footprint analyses to multi-dimensional indicators.} 
Future studies should broaden footprint analyses beyond training to encompass the entire AI pipeline, 
including data collection, preprocessing, deployment, inference, and eventual model retirement 
\cite{castellanos-nieves_improving_2023, barbierato_toward_2024, salehi_data-centric_2024}. 
In contrast to existing reviews, our synthesis foregrounds the integration of \emph{embodied emissions} from hardware production, packaging, and disposal \cite{bannour_evaluating_2021}, together with full Scope~1--3 inventories that capture supply-chain logistics and indirect energy use \cite{dodge_measuring_2022}. 
It should be stressed that multi-dimensional indicators - such as water withdrawals in cooling-intensive regions, 
Water Usage Effectiveness (WUE) in hyperscale data centers, and material scarcity metrics for rare earths - must complement carbon-centric accounts 
to reveal overlooked burdens and trade-offs \cite{li_making_2023, zhou_opportunities_2023}. 
This expansion turns lifecycle analyses into actionable baselines for siting, procurement, and workload-management decisions.

\textbf{Developing comprehensive monitoring tools.} 
User-friendly and integrable monitoring tools remain scarce. Future research must go beyond GPU/CPU/DRAM counters to systematically capture networking, storage, and cooling overheads that often dominate datacenter footprints \cite{anthony_carbontracker_2020}. What is missing today are \emph{end-to-end toolchains} that not only meter resource use but also link results to actionable levers - for instance, flagging energy-intensive data loaders, suggesting batch-size or precision adjustments, or recommending carbon-aware region scheduling across cloud, edge, and on-device systems. Such integration would transform monitoring from passive reporting into active decision support for sustainable AI design and deployment.

\textbf{Bridging to Green IT and HPC scheduling.} 
Earlier Green AI reviews have largely concentrated on footprint measurement and reporting, whereas systematic transfer of established HPC scheduling methods into AI workflows has received little attention. Mechanisms such as power capping (RAPL), dynamic voltage and frequency scaling (DVFS)-based throttling, energy-aware job placement, queue-level energy budgets, preemption, and checkpointing are well established in HPC \cite{anzt_high_2022, li_towards_2024}. Adapting them to AI would enable cluster-level placement (CPU/GPU/TPU bins), coordinated batch sizing under power caps, and co-scheduling of data locality with accelerator availability to reduce both energy and egress overhead. These concrete pathways provide operational levers for reducing AI's carbon footprint beyond the descriptive level emphasized in earlier surveys. 

\textbf{Toward multi-objective resource allocation.} 
Most current studies (including reviews) remain carbon-centric. Future research should explicitly address multi-objective trade-offs, e.g., integrating carbon intensity, water stress indicators, and embodied hardware impacts into workload scheduling decisions. This enables holistic allocation frameworks that align AI with broader Green AI goals rather than single-metric optimization.

\textbf{Closed-loop scheduling and telemetry integration.} 
Another neglected area is the use of direct telemetry. Future AI infrastructures should integrate real-time telemetry from PMCs, rack-level PDUs, and energy-aware mainboards directly into scheduling mechanisms. This continuous feedback enables adaptive workload placement, throttling, and carbon-aware task deferral \cite{tornede_towards_2023, henderson_towards_2020}. Reporting calibration errors alongside corrective actions - such as dynamic power capping - would ensure transparency, auditability, and reproducibility in Green-AI governance. This tight integration of measurement and control - long established in HPC power management - is still absent in AI pipelines, leaving a major gap for research and practice.

\textbf{Critical research gap.} 
Taken together, the literature lacks integrative frameworks that couple workload profiling, hardware-level telemetry (PMCs, PDUs), and region-specific carbon-intensity forecasts into actionable allocation and scheduling policies. Whereas Green IT and HPC already apply energy-aware job scheduling and DVFS-based throttling, equivalent control mechanisms remain largely absent from AI training and inference workflows. This omission leaves scheduling - an established efficiency lever in conventional computing - underexploited within Green AI, despite its potential to substantially reduce operational energy intensity and embodied-carbon impact. Building this bridge constitutes a central and novel pathway for future research, moving Green AI beyond conceptual discussions toward operational practice.
	
\section{Discussion} 
\label{sec:Discussion}
Modern AI application systems consolidate heterogeneous workloads - spanning training, inference, and data management - under stringent service-level and latency constraints. This integration drives peak power draw, amplifying Scope~2 emissions and accelerating hardware obsolescence, thereby embedding environmental costs directly into infrastructural design and operation \cite{robbins_our_2022}. This review addresses these challenges by systematically synthesizing 103 articles and introducing a lifecycle-oriented taxonomy that extends beyond prior reviews \cite{wu_sustainable_2022, verdecchia_systematic_2023, kaack_aligning_2022}. Whereas earlier surveys have mapped high-level implications \cite{wu_sustainable_2022}, aligned AI with climate policy goals \cite{kaack_aligning_2022}, or focused on Green Software Engineering tools and practices \cite{verdecchia_systematic_2023}, the novelty of this article lies in advancing an \emph{operational perspective} on Green AI. Specifically, the contribution consists of: 
(i)~integrating LCA logic with AI-specific phases, including embodied and end-of-life impacts, 
(ii)~linking Green AI explicitly to mature methods from Green IT and HPC, 
(iii)~highlighting neglected multi-objective trade-offs beyond carbon (e.g., water, material scarcity), and (iv)~emphasizing closed-loop integration of measurement and scheduling, where estimator errors are continuously corrected with metered data to document reproducibility and ensure audit-ready reporting. Together, these mechanisms link lifecycle concepts directly to verifiable implementation practices.

\subsection{Theoretical and practical implications} 
\label{subsec:TheoreticalImplications}

From a theoretical perspective, the review underscores that Green AI cannot be reduced to efficiency gains during training alone. Prior perspectives (e.g., \cite{kaack_aligning_2022, wu_sustainable_2022}) have emphasized policy alignment, broad Green AI challenges, or mitigation potential. In contrast, this article contributes a structured \emph{lifecycle model} that explicitly maps all phases of AI systems to LCA categories. This reveals hidden environmental burdens such as Scope~3 emissions, embodied carbon, water use, and resource depletion that are overlooked in carbon-centric accounts.

Practically, the findings advance the state of the art by identifying concrete pathways for operationalizing Green AI. Unlike Verdecchia et~al. \cite{verdecchia_systematic_2023}, who focus on tools for energy transparency, our synthesis highlights the transfer of mature HPC techniques - such as dynamic voltage and frequency scaling (DVFS)-based throttling, Running Average Power Limit, RAPL power capping, queue-level energy budgets, and preemptive checkpointing - into AI pipelines. Adapted to AI, these methods inform workload placement across CPUs/GPUs/TPUs, enable batch-size coordination under power caps, and integrate locality-aware scheduling to minimize both energy and data-movement overhead. This operational bridging between Green IT/HPC and Green AI has not yet been systematically articulated in prior reviews.

Most current approaches remain carbon-centric, neglecting other critical impact vectors. Multi-objective allocation extends workload scheduling by integrating water-use coefficients from facility telemetry and embodied-carbon amortization of hardware refresh cycles alongside grid-intensity signals. Closed-loop scheduling operationalizes this integration: direct performance monitoring counter (PMC) streams and rack-level Power Distribution Unit (PDU) data continuously recalibrate CodeCarbon-like estimators, with deviations fed back into Kubernetes or Slurm plugins to dynamically shift jobs. This allows placement to adapt to real-time regional grid mix and cooling-water stress, ensuring scheduling reflects the full environmental envelope rather than only compute throughput. Explicitly documenting calibration deltas - for example, a 25\% deviation between CodeCarbon estimates and rack-level power meters - and showing how this alters task placement or deferral thresholds allows reproducibility tests and systematic cross-study comparisons.

Finally, the analysis highlights how Green AIs constrained by material and infrastructural lock-ins. Semiconductor supply chains rely on energy-intensive fabrication in a few regions, while short hardware refresh cycles amplify embodied emissions and waste. Studies show that without governance interventions, these structural patterns outweigh efficiency gains at the software level, binding AI progress to unsustainable material and energy flows \cite{robbins_our_2022}. By explicitly embedding end-of-life, resource recovery, and circularity into the Green AI lifecycle, the proposed framework makes these lock-in trajectories visible and provides a basis for evaluating alternative policy interventions (e.g., right-to-repair rules, renewable procurement mandates, or disclosure requirements for embodied carbon).

\subsection{Limitations} 
\label{subsec:Limitations}

Despite its comprehensive scope, this review has limitations. First, the database-driven search, while systematic, may have missed relevant contributions outside computer science venues, particularly in engineering, sustainability science, and political economy. 
Second, the integration of LCA with AI lifecycle phases remains conceptual and requires empirical validation through real-world case studies. Third, although the taxonomy spans 103 articles, continuous revision is necessary as standards, tools, and policies expand. Beyond technical metrics, the next research wave must examine how disclosure regulations, renewable-energy siting, and extended producer responsibility affect global equity. Particular attention is required for regions where material extraction, manufacturing, and e-waste processing occur, since these sites absorb disproportionate ecological burdens from AI infrastructures \cite{robbins_our_2022, li_making_2023, zhou_opportunities_2023}. 

Overall, by integrating lifecycle mapping with operational mechanisms from Green IT and HPS, this article advances prior surveys into a reproducible methodology. It specifies phase-explicit measurement protocols linked to inventory and impact assessment stages, enforces calibration of estimator tools against performance monitoring counters (PMCs) and rack-level meters, and embeds embodied-carbon accounting into hardware-selection criteria. Together with levers such as DVFS, RAPL-based capping, and carbon-aware dispatch, these elements translate Green AI targets into auditable engineering practice.
	
\section{Conclusion} 
\label{sec:Conclusion}

This article consolidates the foundations of \emph{Green AI} and advances the field from concept to practice. A lifecycle-oriented perspective is combined with a class-based hardware selection framework and a differentiated measurement architecture that integrates indirect software estimators with direct, hardware-level telemetry. The synthesis emphasizes operational levers (carbon-aware scheduling, transfer of mature Green IT/high-performance computing (HPC) techniques, and closed-loop calibration) that enable auditable Green AI improvements beyond training-only efficiency gains.

\textbf{Practice and policy implications.}
Future AI studies should report environmental metrics alongside standard performance results and adopt standardized disclosure of assumptions (region-specific carbon intensity and temporal granularity, Power Usage Effectiveness, PUE; deferral windows/SLA constraints, migration and data-locality policies, telemetry sources and calibration error, and weighting across carbon, water, and embodied impacts). Studie of negative results, dataset/code release for Green AI experiments, and integration of Green AI criteria into funding evaluations can materially reduce redundant computation and accelerate methodological convergence \cite{kaack_aligning_2022, wu_sustainable_2022}. 

\textbf{Contributions.}
This article provides a structured and operational contribution to Green AI by: 
(i)~articulating a precise conceptual basis and operational definition of Green AI, 
(ii)~proposing a \emph{five-phase} lifecycle model with \emph{33} concrete realization, adaptation, and adjustment tasks mapped to LCA categories, 
(iii)~introducing a \emph{class-level} hardware selection framework that links contextual determinants (regional grid carbon intensity, datacenter PUE/cooling efficiency, scheduling and deferral windows) to reproducible audit trails, 
(iv)~systematizing measurement methods across indirect estimators and direct approaches, including performance monitoring counters (PMCs), RAPL telemetry, and rack-level Power Distribution Units (PDUs), while exposing limitations of water-use and embodied-emission proxies, and 
(v)~deriving actionable pathways that integrate carbon-aware scheduling, dynamic voltage and frequency scaling, DVFS/power-capping, and HPC-style resource orchestration into AI pipelines through closed-loop telemetry and PCC/PET decision gates.

\textbf{Research answers.}
Table~\ref{tab:SummaryofPrincipalFindingsperResearchQuestion} consolidates the synthesized evidence for RQ1--RQ5, linking each research question to its corresponding findings, methodological steps, and lifecycle phases addressed. The five research questions structure both the analysis and the proposed contributions. 

\begin{table}[t]
	\centering
	\caption{Summary of principal findings per research question.}
	\label{tab:SummaryofPrincipalFindingsperResearchQuestion}
	\begin{tabular}{p{5cm}|p{10cm}}
		\hline
		\textbf{Research question} & \textbf{Principal findings} \\
		\hline
		1) How should \emph{Green AI} be defined and delimited relative to adjacent concepts, and which principles govern its lifecycle-wide application? & 
		$\bullet$ Unified, lifecycle-explicit definition anchored in Life Cycle Assessment (LCA) logic \newline
		$\bullet$ Beyond training efficiency: multi-impact framing (carbon, water, embodied resources) \newline
		$\bullet$ Transparent, standardized reporting established as first-class principle \\
		\hline
		2) Which phases and subphases constitute the \emph{Green AI lifecycle}, and how do they map to Life Cycle Assessment (LCA) stages? & 
		$\bullet$ Five-phase, 33 task process model specifying realization, adaptation, and adjustment actions \newline
		$\bullet$ Structured traceability and comparability of Green AI measures across the full AI system lifespan \newline
		$\bullet$ Lifecycle framing enables improvement of Green AI practices \\
		\hline
		3) Which hardware components and system architectures enable improved performance per watt and reduced embodied impacts across the lifecycle? & 
		$\bullet$ Class-based hardware and infrastructure mapping (edge, workstation, on-prem, cloud accelerators) \newline
		$\bullet$ Explicit context binding (region, Power Usage Effectiveness, PUE; temporal scheduling) for reproducibility and comparability \newline
		$\bullet$ Shift from device prescriptions to architectural principles enabling energy-efficient deployment \\
		\hline
		4) How can environmental impacts (energy, carbon, water) be measured and reported consistently across the lifecycle, including calibration of indirect estimators with direct measurements? & 
		$\bullet$ Hybrid measurement architecture: software estimators (CodeCarbon, Carbontracker) \emph{calibrated} by direct methods (performance monitoring counters (PMCs), rack Power Distribution Units (PDUs), smart plugs) \newline
		$\bullet$ Corporate dashboards useful but ecosystem-bound; water-use estimation remains uncertain and must be qualified \\
		\hline
		5) Which open challenges and research priorities (benchmarks, reporting schemas, uncertainty handling, circularity, governance) are needed to advance lifecycle integration of Green AI? & 
		$\bullet$ Operational carbon-aware scheduling across regions/times; explicit Service Level Agreement (SLA)/deferral reporting \newline
		$\bullet$ Transfer of HPC techniques (dynamic voltage and frequency scaling (DVFS), Running Average Power Limit (RAPL) power caps, energy-aware placement, checkpointing) to AI workflows \newline
		$\bullet$ Closed-loop telemetry and multi-objective allocation (carbon, water, embodied) with auditable calibration \\
		\hline
	\end{tabular}
\end{table}

For RQ1 (``How should \emph{Green AI} be defined and delimited relative to adjacent concepts, and which principles govern its lifecycle-wide application?''), the review establishes a unified definition that distinguishes Green AI from Sustainable AI, anchors it in measurable lifecycle impacts (energy, carbon, water, embodied materials), and sets explicit boundaries for Scope~1--3 reporting. It extends beyond training efficiency to a multi-impact framing (carbon, water, embodied resources) and positions transparent, standardized reporting as a core principle. 

RQ2 (``Which phases and subphases constitute the \emph{Green AI lifecycle}, and how do they map to Life Cycle Assessment (LCA) stages?'') is addressed with a five-phase, 33 subphase process model explicitly aligned to the four LCA stages, ensuring that provisioning, operation, and end-of-life are all systematically represented. It specifies realization, adaptation, and adjustment actions, thereby enabling structured traceability and continuous improvement across the full AI system lifespan. 

For RQ3 (``Which hardware components and system architectures enable improved performance per watt and reduced embodied impacts across the lifecycle?''), the review develops a class-based mapping of edge, workstation, on-premises, and cloud accelerators. By binding hardware choice to contextual factors (region, PUE, scheduling), the focus shifts from device prescriptions to reproducible architectural principles for energy-efficient deployment. 

RQ4 (``How can environmental impacts (energy, carbon, water) be measured and reported consistently across the lifecycle, including calibration of indirect estimators with direct measurements?'') is answered with a hybrid architecture where estimator outputs are continuously benchmarked against direct telemetry from PMCs, rack PDUs, and smart plugs, enabling quantifiable error bounds and lifecycle-consistent reporting. While cloud dashboards provide aggregated transparency at corporate level, their ecosystem lock-in limits reproducibility across heterogeneous infrastructures. Water-use assessment, in particular, remains hindered by reliance on unvalidated cooling models and scarce facility-level disclosures, requiring standardized reporting protocols to ensure scientific comparability \cite{li_making_2023, zhou_opportunities_2023}. 

RQ5 (``Which open challenges and research priorities are needed to advance lifecycle integration of Green AI?'') highlights that water-use assessment remains constrained by reliance on coarse cooling-water coefficients and the absence of validation against metered withdrawals at data centers. This prevents robust cross-facility comparison and hinders integration into lifecycle inventories. Progress requires facility-level disclosure of withdrawal and consumption data, coupled with standardized calibration protocols that explicitly link water usage effectiveness (WUE) estimates to measured cooling performance \cite{li_making_2023, zhou_opportunities_2023}.

\textbf{Outlook.}
The path forward requires 
(i)~hybrid measurement frameworks that report calibration errors of $<$10\% against metered ground truth, 
(ii)~benchmark suites coupling representative AI workloads with temporally resolved carbon intensity, regional WUE factors, and embodied-carbon amortization rules, and 
(iii)~schedulers that make Service Level Agreement (SLA) trade-offs explicit by publishing deferral windows, rollback costs, and environmental deltas per job. 
Embedding Green AI in these operational practices enables reproducible cross-study comparisons, reduces long-term risks of carbon lock-in in data center infrastructures, and grounds technical advances in accountable Green AI trajectories rather than narrow efficiency gains \cite{robbins_our_2022, verdecchia_systematic_2023, kaack_aligning_2022}.

\section{Acknowledgments}
This research was funded by Federal Ministry for Economic Affairs and Climate Action based on a resolution of the German Bundestag (funding code: KK5272103MS3) and the Gebauer GmbH bringing Green AI to ERP systems.
\section{References}

\bibliographystyle{ACM-Reference-Format}
\bibliography{bibliography_RojahnAndGrum}

\appendix
\clearpage

\section{}

\subsection{Term consolidation and synonym grouping results}

\begin{table}[htbp]
	\centering
	\caption{Excerpt from the consolidated term list}
	\resizebox{1\textwidth}{!}{ 
	\begin{filecontents*}{./FiguresAndTables/Term_Consolidation_and_Synonym_Grouping_Results.csv}
Source;Year;development;inference;production;distribution;power consumption;deployment;model training;ai model;manufacturing;efficiency;transparency;transport;privacy;accountability;reuse;benchmarking;pruning;architecture search;energy efficiency;neural architecture search;quantization;sparsity;transfer learning;data collection;hyperparameter tuning;documentation;extraction;life cycle assessment;lighting;maintenance;refine;model distillation;fine tuning;disposal;early stopping;wire;data processing;data storage;edge computing;labeling;cooling system;model compression;end of life;raw material;reproducibility;model deployment;ai ethics;use of ai;federated learning;model design;data augmentation;carbon accounting;recycle;archive;explainable ai;carbon footprint estimation;ai hardware;active learning;residual connection;data acquisition;reduce model size;screening;idea generation;dvfs;model monitoring;environmental impact assessment;cloud sustainability;load balancing;re tuning;hardware acceleration;dynamic voltage and frequency scaling;low rank factorization;data preparation;model implementation;model exploration;carbon aware computing;ai design;real time inference;process improvement;operational use;data sourcing;data usage;cleaning;tokenization;ressource;data provenance;circular economy;algorithmic efficiency;incremental learning;server utilization;technical implementation;adaptive computation;model shrinking;feature engineering;large scale experiments;model comparison;browsing;problem design;business understanding;spare parts;de duplicating;running simulations;crawling;code feature extraction;refactoring detection;data production;feasibility study;model requirements;on device learning;green data centers;environmental standards;dismantle;disposal stage;model documentation;mutual information maximization;continuous integration;test configurations;small experiments;efficient coding;appropriate scaling;model experimentation;educative training;usage monitoring;simulation and analysis;continuous delivery;idle use;dynamic use;license selection;initialisation;low precision arithmetic operations;adaptive sampling;remanufacturing;continual learning;model interpretability;green data center design;fog computing;autoscaling;federated optimization;approximate computing;eco design;modular hardware design;hardware reuse;e waste management;thermal management;chip efficiency;end of life management;data minimization;data lifecycle management;privacy preserving machine learning;model checkpointing;component sourcing;supply chain transparency;conflict minerals;compute aware scheduling;lazy evaluation;resource aware algorithms;sustainability reporting;software containerization;renewable energy integration;model lifecycle management;supply chain audits;environmental kpis;sustainable procurement;regulatory compliance
Gossart;2015;1;0;1;0;1;1;0;0;1;1;0;1;1;0;0;0;0;0;1;0;0;0;0;0;0;0;0;1;1;0;0;0;0;1;0;1;0;1;0;0;0;0;0;1;0;0;0;0;0;0;0;0;0;1;0;0;0;0;0;0;0;0;0;0;0;1;0;0;0;0;0;0;0;0;0;0;0;0;0;0;0;0;0;0;0;0;0;0;0;0;0;0;0;0;0;0;0;0;0;0;0;0;0;0;0;0;0;0;0;0;0;0;0;0;0;0;0;0;0;0;0;0;0;0;0;0;0;0;0;0;0;0;0;0;0;0;0;0;0;0;0;0;0;0;0;0;0;0;0;0;0;0;0;0;0;0;0;0;0;0;0;0;0;0
Batra et al.;2018;1;1;0;0;1;0;0;1;0;1;0;0;0;0;0;0;0;0;0;0;0;0;0;0;0;0;0;0;0;0;1;0;0;0;0;0;0;0;1;0;0;0;0;0;0;0;0;0;0;0;0;0;0;0;0;0;1;0;0;0;0;0;0;0;0;0;0;0;0;0;0;0;0;0;0;0;0;0;0;0;0;0;0;0;0;0;0;0;0;0;0;0;0;0;0;0;0;0;0;0;0;0;0;0;0;0;0;0;0;0;0;0;0;0;0;0;0;0;0;0;0;0;0;0;0;0;0;0;0;0;0;0;0;0;0;0;0;0;0;0;0;0;0;0;0;0;0;0;0;0;0;0;0;0;0;0;0;0;0;0;0;0;0;0
Lacoste et al.;2019;0;1;1;1;1;0;1;0;0;1;1;0;1;0;0;0;0;0;0;0;0;0;0;0;1;0;0;0;0;0;0;0;0;0;0;0;0;0;0;0;0;0;0;0;0;0;0;0;0;0;0;0;0;0;0;0;0;0;0;0;0;0;0;0;0;0;0;1;0;0;0;0;0;0;0;0;0;0;0;0;0;0;0;0;1;0;0;0;0;0;0;0;0;0;0;0;0;0;0;0;0;0;0;0;0;0;0;0;0;0;0;0;0;0;0;0;0;0;0;0;0;0;0;0;0;0;0;0;0;0;0;0;0;0;0;0;0;0;0;0;0;0;0;0;0;0;0;0;0;0;0;0;0;0;0;0;0;0;0;0;0;0;0;0
Wu et al.;2019;1;1;1;1;1;1;0;0;0;0;0;0;0;0;1;1;1;1;0;1;1;1;0;0;0;1;0;0;0;0;0;0;0;0;0;0;0;0;0;0;0;0;0;0;0;1;0;0;0;0;0;0;0;0;0;0;0;0;0;0;0;0;0;0;0;0;0;0;0;0;0;0;0;0;0;0;0;1;0;0;0;0;0;0;0;0;0;0;0;0;0;0;0;0;0;0;0;0;0;0;0;0;0;0;0;0;0;0;0;0;0;0;0;0;0;0;0;0;0;0;0;0;0;0;0;0;0;0;0;0;0;0;0;0;0;0;0;0;0;0;0;0;0;0;0;0;0;0;0;0;0;0;0;0;0;0;0;0;0;0;0;0;0;0
Anthony et al.;2020;1;1;1;0;1;0;1;0;1;1;1;0;0;0;0;1;1;0;0;0;1;0;0;0;1;0;1;0;0;0;0;0;0;0;1;0;0;0;0;0;0;0;0;0;1;0;0;0;0;0;1;0;0;0;0;0;0;0;0;0;0;1;0;1;0;0;0;0;0;0;1;0;0;0;0;0;0;0;0;0;0;0;0;0;0;0;0;0;0;0;0;0;0;0;0;0;0;0;0;0;0;0;0;0;0;0;0;0;0;0;0;0;0;0;0;0;0;0;0;0;0;0;0;0;0;0;0;0;0;0;0;0;0;0;0;0;0;0;0;0;0;0;0;0;0;0;0;0;0;0;0;0;0;0;0;0;0;0;0;0;0;0;0;0
Strubell et al.;2020;1;1;0;1;1;0;0;1;0;0;0;0;0;0;0;0;0;1;0;1;0;0;0;0;1;0;0;0;0;0;0;0;0;0;0;0;0;0;0;1;0;0;0;0;0;0;0;0;0;0;0;0;0;0;0;0;0;0;0;0;0;0;0;0;0;0;0;0;0;0;0;0;0;0;0;0;0;0;0;0;0;0;0;0;0;0;0;0;0;0;0;0;0;0;0;0;0;0;0;0;0;0;0;0;0;0;0;0;0;0;0;0;0;0;0;0;0;0;0;0;0;0;0;0;0;0;0;0;0;0;0;0;0;0;0;0;0;0;0;0;0;0;0;0;0;0;0;0;0;0;0;0;0;0;0;0;0;0;0;0;0;0;0;0
Schwartz et al.;2020;1;1;1;0;0;0;0;1;0;1;1;0;0;0;1;0;1;1;0;1;0;0;1;0;1;0;0;0;1;0;0;0;1;0;1;0;0;0;0;0;0;0;0;0;0;0;0;0;0;0;1;0;0;0;0;0;0;0;0;0;0;0;0;0;0;0;0;0;0;0;0;0;0;1;0;0;0;0;0;0;0;0;0;0;0;0;0;0;0;0;0;0;0;0;0;0;0;0;0;0;0;0;0;0;0;0;0;0;0;0;0;0;0;0;0;0;0;0;0;0;0;0;0;0;0;0;0;0;0;0;0;0;0;0;0;0;0;0;0;0;0;0;0;0;0;0;0;0;0;0;0;0;0;0;0;0;0;0;0;0;0;0;0;0
Hernandez and Brown;2020;0;1;0;0;0;0;0;1;0;0;0;0;0;0;0;0;0;1;0;0;0;1;1;0;0;0;0;0;0;0;0;0;0;0;0;0;0;0;0;0;0;0;0;0;0;0;0;0;0;0;0;0;0;0;0;0;0;0;1;0;0;0;0;0;0;0;0;0;0;0;0;0;0;0;0;0;0;0;0;0;0;0;0;0;0;0;0;0;0;0;0;0;0;0;0;0;0;0;0;0;0;0;0;0;0;0;0;0;0;0;0;0;0;0;0;0;0;0;0;1;0;0;0;0;0;0;0;0;0;0;0;0;0;0;0;0;0;0;0;0;0;0;0;0;0;0;0;0;0;0;0;0;0;0;0;0;0;0;0;0;0;0;0;0
Henderson et al.;2020;1;1;1;1;1;1;1;0;1;0;1;1;0;1;0;1;0;0;0;0;0;0;0;0;0;1;0;1;1;0;0;0;0;1;0;0;0;0;0;1;0;0;0;0;1;0;0;0;0;0;0;1;0;0;0;0;0;0;0;0;0;0;0;0;0;0;0;0;0;0;0;0;0;0;0;0;0;0;0;0;0;0;0;0;0;0;0;0;0;0;0;0;0;0;0;0;0;0;0;0;0;0;0;0;0;0;0;0;0;0;0;0;0;0;0;0;0;0;0;0;0;0;0;0;0;0;0;0;0;0;0;0;0;0;0;0;0;0;0;0;0;0;0;0;0;0;0;0;0;0;0;0;0;0;0;0;0;0;0;0;0;0;0;0
Trébaol;2020;1;0;1;1;1;1;0;0;1;0;1;1;1;0;0;1;0;0;0;0;0;0;0;0;0;1;0;0;0;0;0;0;0;0;0;1;0;0;0;0;0;0;0;0;0;1;0;1;0;0;0;0;0;0;0;1;0;0;0;0;0;1;0;0;0;0;0;0;0;0;0;0;0;1;0;0;0;0;0;0;0;0;0;0;0;0;0;0;0;0;0;0;0;0;0;0;1;0;0;0;0;0;0;0;0;0;0;0;0;0;0;0;0;0;0;0;0;0;0;0;0;0;0;0;0;0;0;0;0;0;0;0;0;0;0;0;0;0;0;0;0;0;0;0;0;0;0;0;0;0;0;0;0;0;0;0;0;0;0;0;0;0;0;0
Shaikh et al.;2021;0;0;1;0;0;1;1;0;0;0;1;0;0;0;0;0;0;1;0;1;0;0;0;0;0;0;0;0;1;0;0;0;0;0;1;0;0;0;0;0;0;0;0;0;0;0;0;0;0;0;0;0;0;0;0;0;0;0;0;0;0;0;0;0;0;0;0;0;0;0;0;0;0;0;0;0;0;0;0;0;0;0;0;0;0;0;0;0;0;0;0;0;0;0;0;0;0;0;0;0;0;0;0;0;0;0;0;0;0;0;0;0;0;0;0;0;0;0;0;0;0;0;0;0;0;0;0;0;0;0;0;0;0;0;0;0;0;0;0;0;0;0;0;0;0;0;0;0;0;0;0;0;0;0;0;0;0;0;0;0;0;0;0;0
Xu et al.;2021;1;1;1;1;1;1;1;1;0;0;0;0;1;0;1;0;1;1;0;1;1;1;1;0;0;0;1;0;0;0;0;1;1;0;1;0;0;0;1;1;0;1;0;0;0;0;0;0;0;1;0;0;0;0;0;0;0;1;1;0;1;0;0;0;0;0;0;0;1;0;0;1;0;0;0;0;0;0;0;0;0;1;0;0;0;0;0;0;0;0;0;1;1;0;0;0;0;0;0;0;0;0;0;0;0;0;0;0;0;0;0;0;0;0;0;0;0;0;0;0;0;0;0;0;0;0;0;0;0;0;0;0;0;0;0;0;0;0;0;0;0;0;0;0;0;0;0;0;0;0;0;0;0;0;0;0;0;0;0;0;0;0;0;0
Haakman et al.;2021;1;1;1;1;0;1;1;1;0;0;1;0;1;1;1;0;0;0;0;0;0;0;0;1;0;1;0;0;0;0;1;0;0;1;0;0;1;0;0;1;0;0;1;0;1;1;0;0;0;1;0;0;0;0;1;0;0;0;0;1;0;0;0;0;1;0;0;0;0;0;0;0;1;0;0;0;0;0;1;0;0;0;1;0;0;0;0;0;0;0;0;0;0;1;0;0;0;1;1;0;0;0;0;0;0;0;1;1;0;0;0;0;1;0;0;1;0;0;0;0;0;0;0;0;1;0;0;0;0;0;0;0;0;0;0;0;0;0;0;0;0;0;0;0;0;0;0;0;0;0;0;0;0;0;0;0;0;0;0;0;0;0;0;0
Bannour et al.;2021;1;1;1;1;0;1;0;0;1;0;0;0;0;1;0;1;0;0;0;0;0;0;0;0;0;1;1;0;0;0;0;0;0;0;0;0;0;0;0;1;0;0;1;0;0;0;0;0;0;0;0;0;0;0;0;0;0;0;0;0;0;0;0;0;0;0;0;0;0;0;0;0;0;0;0;0;0;0;0;0;0;0;0;0;0;0;0;0;0;0;0;0;0;0;0;0;0;0;0;0;0;0;0;0;0;0;0;0;0;0;0;0;0;0;0;0;0;0;0;0;0;0;0;0;0;1;1;0;0;0;0;0;0;0;0;0;0;0;0;0;0;0;0;0;0;0;0;0;0;0;0;0;0;0;0;0;0;0;0;0;0;0;0;0
Li et al.;2021;1;1;0;1;1;1;1;1;0;0;0;0;0;0;0;1;1;1;0;1;1;0;1;0;0;0;0;0;0;0;0;1;0;0;0;0;0;0;0;0;1;1;0;0;0;1;0;0;0;0;0;0;0;0;0;0;0;0;0;0;0;0;0;0;0;0;0;0;0;0;0;0;0;0;0;0;0;0;0;0;0;0;0;0;0;0;0;0;0;0;0;0;0;0;0;0;0;0;0;0;0;0;0;0;0;0;0;0;0;0;0;0;0;0;0;0;0;0;0;0;0;0;0;0;0;0;0;0;0;0;0;0;0;0;0;0;0;0;0;0;0;0;0;0;0;0;0;0;0;0;0;0;0;0;0;0;0;0;0;0;0;0;0;0
Patterson et al.;2021;1;1;1;1;1;0;1;0;0;1;0;0;0;0;1;1;1;1;1;1;1;1;1;0;0;0;0;0;0;0;1;1;1;0;1;1;0;0;0;0;0;0;0;0;0;0;0;0;0;0;0;0;0;0;0;0;0;0;0;0;0;0;0;1;0;0;0;0;0;0;0;0;0;0;0;0;0;0;0;0;0;0;0;0;0;1;0;1;0;1;0;0;0;0;0;0;0;0;0;0;0;0;0;0;0;0;0;0;0;0;0;0;0;0;0;0;0;0;1;0;0;0;0;0;0;0;0;0;0;0;0;0;0;0;0;0;0;0;0;0;0;0;0;0;0;0;0;0;0;0;0;0;0;0;0;0;0;0;0;0;0;0;0;0
Patterson et al.;2021;1;1;1;0;1;1;1;0;1;1;1;0;0;1;1;0;0;1;1;1;0;1;0;0;0;0;0;0;0;0;0;0;0;0;0;0;0;0;0;0;0;0;0;0;0;0;0;0;1;0;0;1;0;0;0;0;0;0;0;0;0;0;0;0;0;0;0;0;0;0;0;0;0;0;0;0;0;0;0;0;0;0;0;0;0;0;0;0;0;0;0;0;0;0;0;0;0;0;0;0;0;0;0;0;0;0;0;0;0;0;0;0;0;0;0;0;0;0;0;0;0;0;0;0;0;0;0;0;0;0;0;0;0;0;0;0;0;0;0;0;0;0;0;0;0;0;0;0;0;0;0;0;0;0;0;0;0;0;0;0;0;0;0;0
Lannelongue et al.;2021;1;0;1;1;1;0;1;0;1;0;0;1;0;0;1;0;0;0;0;0;0;0;0;0;0;0;0;1;1;1;0;0;0;0;0;0;1;0;0;0;0;0;0;0;0;0;0;0;0;0;0;0;0;1;0;0;0;0;0;0;0;0;0;0;0;0;0;0;0;0;0;0;0;0;0;0;0;0;0;0;0;0;0;0;0;0;0;0;0;0;0;0;0;0;0;0;0;0;0;0;0;0;0;0;0;0;0;0;0;0;0;0;0;0;0;0;0;0;0;0;0;0;0;0;0;0;0;0;0;0;0;0;0;0;0;0;0;0;0;0;0;0;0;0;0;0;0;0;0;0;0;0;0;0;0;0;0;0;0;0;0;0;0;0
Cao et al.;2021;0;1;0;1;1;1;0;0;0;0;0;0;0;0;0;0;0;0;0;0;0;1;0;0;1;0;0;0;1;0;0;0;0;0;0;0;0;0;0;0;0;0;0;0;0;0;0;0;0;0;0;0;0;0;0;0;0;0;0;0;0;0;0;0;0;0;0;0;0;0;0;0;0;0;0;0;0;0;0;0;0;0;0;0;0;0;0;0;0;0;0;0;0;0;0;0;0;0;0;0;0;0;0;0;0;0;0;0;0;0;0;0;0;0;0;0;0;0;0;0;0;0;0;0;0;0;0;0;0;0;0;0;0;0;0;0;0;0;0;0;0;0;0;0;0;0;0;0;0;0;0;0;0;0;0;0;0;0;0;0;0;0;0;0
Van Wynsberghe;2021;1;0;1;1;0;0;1;1;1;0;0;0;1;1;0;0;0;1;0;1;0;0;0;0;0;0;0;0;0;0;0;0;0;0;0;0;0;0;0;0;1;0;0;0;0;0;1;1;0;0;0;0;0;0;0;0;0;0;0;0;0;0;1;0;0;0;0;0;1;0;0;0;0;0;0;0;0;0;0;0;0;0;0;0;0;0;0;0;0;0;0;0;0;0;0;0;0;0;0;0;0;0;0;0;0;0;0;0;0;0;0;0;0;0;0;0;0;0;0;0;0;0;0;0;0;0;0;0;0;0;0;0;0;0;0;0;0;0;0;0;0;0;0;0;0;0;0;0;0;0;0;0;0;0;0;0;0;0;0;0;0;0;0;0
Robbins and Van Wynsberghe;2022;1;0;1;0;1;0;0;1;1;0;1;1;1;0;0;0;0;0;0;0;0;0;0;1;0;0;0;0;0;0;0;0;0;1;0;1;1;0;1;1;0;0;0;0;0;0;1;1;0;0;0;0;0;0;1;0;0;0;0;0;0;0;1;0;0;0;0;0;1;0;0;0;0;0;0;0;0;0;0;0;0;0;0;0;0;0;0;0;0;0;0;0;0;0;0;0;0;0;0;0;0;0;0;0;0;0;0;0;0;0;0;0;0;0;0;0;0;0;0;0;0;0;0;0;0;0;0;0;0;0;0;0;0;0;0;0;0;0;0;0;0;0;0;0;0;0;0;0;0;0;0;0;0;0;0;0;0;0;0;0;0;0;0;0
Ligozat et al.;2022;1;1;1;1;0;1;0;1;1;0;0;1;0;0;1;0;0;0;0;0;0;1;0;0;0;0;1;1;0;0;0;0;0;0;0;0;0;1;0;0;0;0;1;1;0;0;0;0;0;0;0;0;1;1;0;0;0;0;0;1;0;0;0;0;0;1;0;0;0;0;0;0;0;0;0;0;0;0;0;0;0;0;0;0;0;0;0;0;0;0;0;0;0;0;0;0;0;0;0;0;0;0;0;0;0;1;0;0;0;0;0;1;0;0;0;0;0;0;0;0;0;0;0;0;0;0;0;0;0;0;0;0;0;0;0;0;0;0;0;0;0;0;0;0;0;0;0;0;0;0;0;0;0;0;0;0;0;0;0;0;0;0;0;0
Dodge et al.;2022;1;1;1;0;1;1;1;1;1;1;1;1;0;1;0;0;0;1;1;1;0;0;1;0;1;0;1;0;0;1;1;0;0;1;0;0;0;0;0;0;0;0;0;1;0;0;0;0;1;0;0;1;0;0;0;0;0;0;0;0;0;0;0;0;0;1;1;0;0;0;0;0;0;0;0;0;0;0;0;0;0;0;0;0;0;0;0;0;0;0;0;0;0;0;0;0;0;0;0;0;0;0;0;0;0;0;0;0;0;0;0;0;0;0;0;0;0;0;0;0;0;0;0;0;0;0;0;0;0;0;0;0;0;0;0;0;0;0;0;0;0;0;0;0;0;0;0;0;0;0;0;0;0;0;0;0;0;0;0;0;0;0;0;0
Kaack et al.;2022;1;1;1;0;0;1;1;0;1;1;1;1;1;1;1;0;0;0;1;0;0;0;1;1;1;0;1;1;0;1;0;0;1;0;0;0;0;0;0;0;1;0;1;1;0;0;1;0;0;0;0;0;0;0;0;0;0;0;0;0;0;0;0;0;0;0;0;0;0;0;0;0;0;0;0;0;0;0;0;0;0;0;0;0;0;0;0;0;0;0;0;0;0;0;0;0;0;0;0;0;0;0;0;0;0;0;0;0;0;0;0;0;0;0;0;0;0;0;0;0;0;0;0;0;0;0;0;0;0;0;0;0;0;0;0;0;0;0;0;0;0;0;0;0;0;0;0;0;0;0;0;0;0;0;0;0;0;0;0;0;0;0;0;0
Georgiou et al.;2022;1;1;0;1;1;0;1;0;1;1;0;1;0;0;0;0;1;0;1;0;0;0;0;0;0;1;0;0;0;1;0;0;0;0;0;0;0;0;1;0;0;0;0;0;1;0;0;0;0;0;0;0;0;0;0;0;0;0;0;0;0;0;0;0;0;0;0;0;0;0;0;0;0;0;0;0;0;0;0;0;0;0;0;0;0;0;0;0;0;0;0;0;0;0;1;0;0;0;0;0;0;0;0;0;0;0;0;0;0;0;0;0;0;0;0;0;1;1;0;0;0;0;0;0;0;0;0;0;1;0;0;0;0;0;0;0;0;0;0;0;0;0;0;0;0;0;0;0;0;0;0;0;0;0;0;0;0;0;0;0;0;0;0;0
Wu et al.;2022;1;1;1;1;1;1;1;1;1;1;1;1;1;1;0;1;1;1;0;1;1;1;0;1;1;0;0;0;0;0;0;0;0;0;0;1;1;1;0;0;0;0;0;0;0;0;0;1;1;0;0;1;0;0;0;0;1;0;0;0;0;0;0;0;0;0;0;0;0;0;0;0;0;0;1;1;0;0;0;1;0;0;0;0;0;0;1;0;0;1;0;0;0;0;0;0;0;0;0;0;0;0;0;0;0;0;0;0;1;0;0;0;0;0;0;0;0;0;0;0;1;0;0;0;0;0;0;0;0;0;0;0;0;0;0;0;0;0;0;0;0;0;0;0;0;0;0;0;0;0;0;0;0;0;0;0;0;0;0;0;0;0;0;0
Gupta et al.;2022;1;1;1;0;1;0;0;0;1;1;0;1;0;0;1;0;0;0;1;0;0;0;0;0;0;0;0;1;0;0;0;0;0;0;0;0;0;0;0;0;0;0;1;1;0;0;0;0;0;0;0;1;1;0;0;0;0;0;0;0;0;0;0;1;0;0;0;0;0;0;0;0;0;0;0;1;0;0;0;1;0;0;0;0;0;0;0;0;0;0;0;0;0;0;0;0;0;0;0;0;0;0;0;0;0;0;0;0;0;0;0;0;0;0;0;0;0;0;0;0;0;0;0;0;0;0;0;0;0;0;0;0;0;0;0;0;0;0;0;0;0;0;0;0;0;0;0;0;0;0;0;0;0;0;0;0;0;0;0;0;0;0;0;0
Menghani;2023;1;1;1;1;1;1;1;0;0;1;0;0;1;0;1;1;1;1;0;1;1;1;0;0;1;1;0;0;0;0;1;1;1;1;1;0;0;0;0;1;0;1;0;0;0;0;0;0;0;0;1;0;0;0;0;0;0;1;0;0;0;0;0;0;0;0;0;0;0;0;0;1;0;0;0;0;0;0;0;0;0;0;0;0;0;0;0;0;0;0;0;0;0;0;0;0;0;0;0;0;0;0;0;0;0;0;0;0;0;0;0;0;0;0;0;0;0;0;0;0;0;0;0;0;0;0;0;0;0;0;0;0;0;0;0;0;0;0;0;0;0;0;0;0;0;0;0;0;0;0;0;0;0;0;0;0;0;0;0;0;0;0;0;0
Eilam et al.;2023;1;1;0;1;1;1;1;1;1;1;0;1;0;0;0;0;0;0;1;0;0;0;1;1;0;0;1;1;0;1;0;1;1;0;0;0;0;0;0;0;0;0;0;0;0;0;0;0;0;1;0;0;0;0;0;0;0;0;0;1;0;0;0;0;0;0;0;0;0;0;1;0;1;0;1;0;0;0;0;0;0;0;0;0;0;0;0;0;0;0;0;0;0;0;0;0;0;0;0;0;1;1;1;0;0;0;0;0;0;0;0;0;0;0;0;0;0;0;0;0;0;0;0;0;0;0;0;0;0;0;0;0;0;0;0;0;0;0;1;0;0;0;0;0;0;0;0;0;0;0;0;0;0;0;0;0;0;0;0;0;0;0;0;0
Verdecchia et al.;2023;1;1;1;1;1;1;1;1;0;1;1;0;1;1;1;1;1;0;1;0;1;0;0;1;1;1;1;1;1;0;1;1;0;0;0;1;0;0;1;0;0;0;0;0;1;0;0;1;1;0;0;0;1;0;0;0;0;0;0;0;0;0;0;0;1;0;0;0;0;0;0;0;0;0;0;0;0;0;0;0;0;0;0;0;0;0;1;0;0;0;0;0;0;0;0;1;0;0;0;0;0;0;0;0;0;0;0;0;0;0;0;0;0;0;0;0;0;0;0;0;0;0;0;0;0;0;0;0;0;0;0;0;0;0;0;0;0;0;0;0;0;0;0;0;0;0;0;0;0;0;0;0;0;0;0;0;0;0;0;0;0;0;0;0
Chen et al.;2023;1;1;0;1;0;1;1;1;0;0;1;1;1;1;1;0;1;0;0;0;1;1;1;1;0;0;0;0;1;0;0;1;0;0;0;1;0;0;1;0;0;1;0;0;0;0;1;0;1;1;0;0;0;0;1;0;0;1;0;1;1;0;0;0;0;0;0;0;0;0;0;0;0;0;0;0;1;1;0;0;0;0;0;0;0;0;0;0;0;0;0;0;0;0;0;0;0;0;0;0;0;0;0;0;0;0;0;0;0;0;0;0;0;0;0;0;0;0;0;0;0;0;0;0;0;0;0;0;0;0;0;0;0;0;0;0;0;1;0;0;0;0;0;0;0;0;0;0;0;0;0;0;0;0;0;0;0;0;0;0;0;0;0;0
Grum et al.;2023;1;0;1;1;0;0;0;0;0;0;0;1;0;0;1;0;0;0;0;0;0;0;0;1;0;0;0;0;0;0;0;0;0;0;0;0;0;0;0;0;0;0;0;0;0;0;0;0;0;0;0;0;0;0;0;0;0;0;0;0;0;0;0;0;0;0;0;0;0;0;0;0;0;0;0;0;0;0;0;0;0;0;0;0;0;0;0;0;0;0;1;0;0;0;0;0;0;0;0;0;0;0;0;0;0;0;0;0;0;0;0;0;0;0;0;0;0;0;0;0;0;1;0;1;0;0;0;0;0;0;0;0;0;0;0;0;0;0;0;0;0;0;0;0;0;0;0;0;0;0;0;0;0;0;0;0;0;0;0;0;0;0;0;0
Tornede et al.;2023;1;1;1;1;1;0;0;0;0;0;1;0;0;1;0;1;0;1;0;1;0;0;1;0;1;0;0;0;0;1;0;0;0;0;1;0;0;0;0;0;0;1;0;1;1;0;1;0;0;0;0;0;0;0;0;0;0;0;0;0;0;0;0;0;0;0;0;0;0;0;0;0;0;0;0;0;0;0;0;0;0;0;0;0;0;0;0;0;0;0;0;0;0;0;1;0;0;0;0;1;0;0;0;0;0;0;0;0;0;0;0;0;0;0;0;0;0;0;0;0;0;0;0;0;0;0;0;0;0;0;0;0;0;0;0;0;0;0;0;0;0;0;0;0;0;0;0;0;0;0;0;0;0;0;0;0;0;0;0;0;0;0;0;0
Bouza et al.;2023;1;0;1;1;1;1;0;1;1;0;1;1;0;1;0;0;0;0;0;0;0;0;0;1;0;1;0;1;0;0;0;0;0;0;0;0;0;1;0;0;0;0;1;0;0;0;0;0;0;0;0;0;0;0;0;1;0;0;0;0;0;0;0;0;0;0;0;0;0;0;0;0;0;0;0;0;0;0;0;0;0;0;0;0;0;0;0;0;0;0;0;0;0;0;0;0;0;0;0;0;0;0;0;0;0;0;0;0;0;0;0;0;0;0;0;0;0;0;0;0;0;0;0;0;0;0;0;0;0;0;0;0;0;0;0;0;0;0;0;0;0;0;0;0;0;0;0;0;0;0;0;0;0;0;0;0;0;0;0;0;0;0;0;0
Luccioni et al.;2023;1;1;1;1;1;1;1;0;1;0;1;1;0;0;0;1;0;0;0;0;0;0;0;0;0;1;1;1;0;1;0;0;0;1;0;1;1;0;0;0;1;0;0;1;0;1;0;0;0;0;0;0;0;0;0;1;0;0;0;0;0;0;0;0;0;0;0;0;0;0;0;0;0;0;0;0;0;0;0;0;1;0;0;1;1;0;0;0;0;0;0;0;0;0;0;0;0;0;0;0;0;0;0;0;0;0;0;0;0;0;0;0;0;0;0;0;0;0;0;0;0;0;0;0;0;0;0;0;0;0;0;0;0;0;0;0;0;0;0;0;0;0;0;0;0;0;0;0;0;0;0;0;0;0;0;0;0;0;0;0;0;0;0;0
Li et al.;2023;1;1;0;1;1;0;1;1;1;1;1;0;1;0;1;0;1;0;1;0;0;0;0;0;0;0;0;0;0;0;0;0;0;0;0;0;0;1;0;0;1;0;0;0;0;0;0;0;0;0;0;0;0;0;0;0;0;0;0;0;0;0;0;0;0;0;1;1;0;0;0;0;0;0;0;0;0;0;0;0;0;0;0;0;0;0;0;0;0;0;0;0;0;0;0;0;0;0;0;0;0;0;0;0;0;0;0;0;0;0;0;0;0;0;0;0;0;0;0;0;0;0;0;0;0;0;0;0;0;0;0;0;0;0;1;0;0;0;0;0;0;0;0;0;0;0;0;0;0;0;0;0;0;0;0;0;0;0;0;0;0;0;0;0
Zhou et al.;2023;1;1;1;1;1;1;1;1;0;1;0;1;1;0;1;0;1;1;1;1;1;1;1;1;1;0;1;0;1;1;1;1;1;1;1;1;1;1;1;1;0;1;0;0;0;0;0;0;0;1;1;0;1;1;0;0;0;1;1;0;0;1;0;0;0;0;0;0;0;1;0;1;0;0;1;0;0;0;0;0;0;1;0;0;0;0;0;0;1;0;0;1;0;1;0;0;0;0;0;0;0;0;0;1;1;0;0;0;0;0;0;0;0;0;0;0;0;0;0;0;0;0;0;0;0;0;0;0;0;0;1;0;1;1;0;0;1;0;0;0;0;0;0;0;0;0;0;0;0;0;0;0;0;0;0;0;0;0;0;0;0;0;0;0
Alzoubi and Mishra;2024;1;1;1;1;1;1;1;1;1;1;1;1;1;1;0;1;1;0;1;0;1;1;0;1;0;1;1;1;1;1;1;1;1;1;1;0;1;1;1;0;1;1;0;0;0;1;0;1;0;0;0;0;0;0;0;1;1;0;0;0;1;0;0;0;0;0;1;0;0;1;1;0;0;0;0;0;0;0;0;0;0;0;0;0;0;0;0;0;0;0;0;0;0;0;0;0;0;0;0;0;0;0;0;0;0;0;0;0;0;0;1;0;0;0;0;0;0;0;0;0;0;0;0;0;0;0;0;0;0;0;0;0;0;0;0;1;0;0;0;0;0;0;0;0;0;0;0;0;0;0;0;0;0;0;0;0;0;0;0;0;0;0;0;0
Longpre et al.;2024;1;1;0;1;0;1;1;1;1;1;1;0;1;1;1;1;0;0;0;0;0;0;1;1;0;1;0;0;0;1;1;0;1;0;0;0;1;1;0;0;0;0;0;0;1;1;0;0;1;0;0;0;0;1;0;0;0;0;0;0;0;0;0;0;0;0;0;0;0;0;0;0;1;0;0;0;0;0;0;0;1;0;1;1;0;1;0;0;0;0;0;0;0;0;0;0;0;0;0;0;0;0;0;0;0;0;0;0;0;0;0;0;0;1;0;0;0;0;0;0;0;0;1;0;0;0;0;1;0;0;0;1;0;0;0;0;0;0;0;0;0;0;0;0;0;0;0;0;0;0;0;0;0;0;0;0;0;0;0;0;0;0;0;0
Bolón-Canedo et al.;2024;1;1;1;1;1;1;0;1;1;1;1;1;1;1;0;0;1;0;1;0;1;1;0;1;1;0;0;0;1;1;0;1;1;0;0;1;1;0;1;0;1;0;0;1;0;0;0;1;1;1;1;0;0;0;1;0;1;0;0;0;0;0;0;0;0;0;0;1;0;0;0;0;0;0;0;0;0;0;0;0;0;0;0;0;0;0;0;0;0;0;0;0;1;0;0;0;0;0;0;0;0;0;0;0;0;0;0;0;0;0;0;0;0;0;1;0;0;0;0;0;0;0;0;0;0;0;0;0;0;1;0;0;0;0;0;0;0;0;0;0;0;0;0;0;0;0;0;0;0;0;0;0;0;0;0;0;0;0;0;0;0;0;0;0
Grum and Gronau;2024;1;1;1;1;0;0;0;1;1;0;0;0;0;1;0;1;0;0;0;0;0;0;1;1;0;1;0;0;0;1;1;0;0;0;0;0;0;0;0;0;0;0;0;0;0;0;0;0;0;0;0;0;0;0;0;0;0;0;0;0;0;0;0;0;0;0;0;0;0;0;0;0;0;1;0;0;1;0;1;0;0;0;0;0;0;0;0;0;0;0;1;0;0;0;0;0;0;0;0;0;0;0;0;0;0;0;0;0;0;0;0;0;0;0;0;0;0;0;0;0;0;0;0;0;0;0;0;0;0;0;0;0;0;0;0;0;0;0;0;0;0;0;0;0;0;0;0;0;0;0;0;0;0;0;0;0;0;0;0;0;0;0;0;0
Barbierato and Gatti;2024;1;1;1;1;1;1;1;1;0;1;0;1;0;0;0;1;1;0;1;0;1;1;1;0;0;0;1;0;1;0;1;1;1;1;1;0;0;1;0;0;1;1;1;0;0;0;1;0;0;0;1;0;0;0;0;0;0;0;1;0;0;0;1;0;1;0;0;0;0;1;0;0;0;0;0;0;0;0;0;0;0;0;0;0;0;0;0;0;1;0;0;0;0;0;0;1;0;0;0;0;0;0;0;0;0;0;0;0;0;1;0;0;0;0;0;0;0;0;0;0;0;0;0;0;0;0;0;0;0;0;0;0;0;0;0;0;0;0;0;0;0;0;0;0;0;0;0;0;0;0;0;0;0;0;0;0;0;0;0;0;0;0;0;0
Wright et al.;2025;1;1;1;1;0;1;1;1;1;1;1;1;1;1;0;1;0;1;1;1;1;0;0;0;0;0;0;1;0;0;0;0;0;1;0;0;0;0;0;0;0;0;1;0;0;0;1;0;0;0;0;0;1;0;0;0;0;0;0;0;0;0;0;0;0;0;0;0;0;0;0;0;0;0;0;0;0;0;0;0;0;0;0;0;0;0;0;1;0;0;0;0;0;0;0;0;0;0;0;0;0;0;0;0;0;0;0;0;0;0;0;0;0;0;0;0;0;0;0;0;0;0;0;0;0;0;0;0;0;0;0;0;0;0;0;0;0;0;0;0;0;0;0;0;0;0;0;0;0;0;0;0;0;0;0;0;0;0;0;0;0;0;0;0
;;39;35;33;32;30;27;26;24;24;22;22;21;18;17;17;17;16;16;15;15;14;14;14;14;13;13;12;12;12;12;11;11;11;11;10;10;9;9;9;8;8;8;8;8;7;7;7;7;7;6;6;5;5;5;4;4;4;4;4;4;3;3;3;3;3;3;3;3;3;3;3;3;3;3;3;2;2;2;2;2;2;2;2;2;2;2;2;2;2;2;2;2;2;2;2;2;1;1;1;1;1;1;1;1;1;1;1;1;1;1;1;1;1;1;1;1;1;1;1;1;1;1;1;1;1;1;1;1;1;1;1;1;1;1;1;1;1;1;1;0;0;0;0;0;0;0;0;0;0;0;0;0;0;0;0;0;0;0;0;0;0;0;0;0
\end{filecontents*}
\csvreader[
	separator=semicolon,
	tabular=|l|l|l|l|l|l|l|l|,
	table head=\hline Source & Year & development & inference & production & distribution & power consumption & deployment \\ \hline,
	late after line=\\\hline
	]{./FiguresAndTables/Term_Consolidation_and_Synonym_Grouping_Results.csv}{}
	{\csvcoli & \csvcolii & \csvcoliii & \csvcoliv & \csvcolv & \csvcolvi & \csvcolvii & \csvcolviii}
	}
	\label{tab:ConsolidatedTermList}
\end{table}

\clearpage

\subsection{Bar charts of lifecycle phases}

\subsubsection{Terms occurring more than eight times (upper panel)}

\begin{figure}[ht]
	\centering
	\includegraphics[width=0.7\textheight, angle=270]{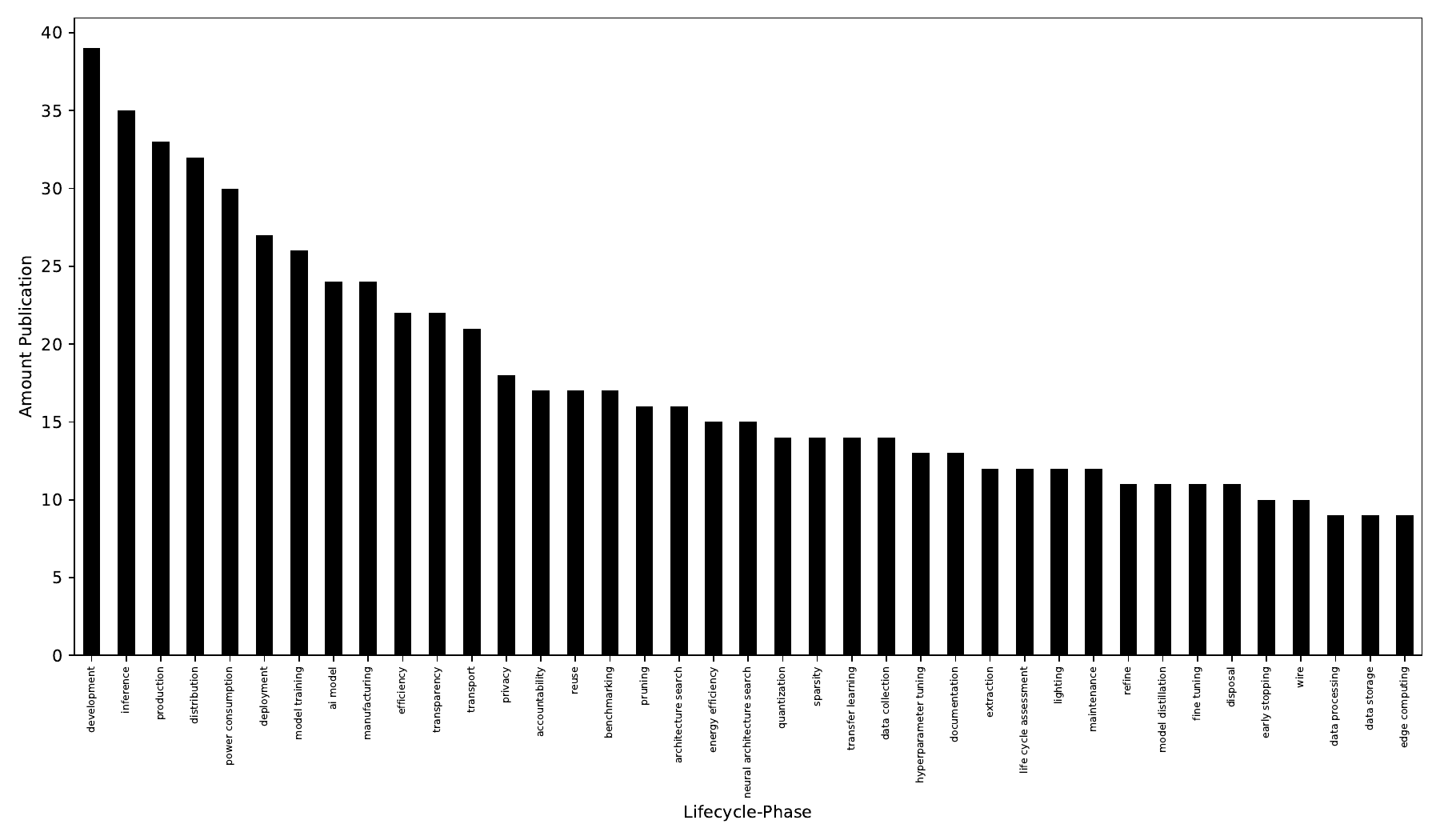}
	\caption{Bar chart of lifecycle phases - \textbf{upper panel} showing terms occurring more than eight times across the corpus. These high-frequency ``head'' terms represent dominant lifecycle stages such as \emph{model training}, \emph{deployment}, \emph{inference}, and \emph{power consumption}, which collectively account for the majority of observed salience. Their prominence reflects a research focus on compute-intensive phases, as quantified in section~\ref{subsubsec:StateoftheArtGreenAILifecycleStages}.}
	\label{fig:BarchartLifeCyclePhasesOver8}
\end{figure}

\clearpage

\subsubsection{Terms occurring up to eight times (lower panel)}

\begin{figure}[ht]
	\centering
	\includegraphics[width=0.7\textheight, angle=270]{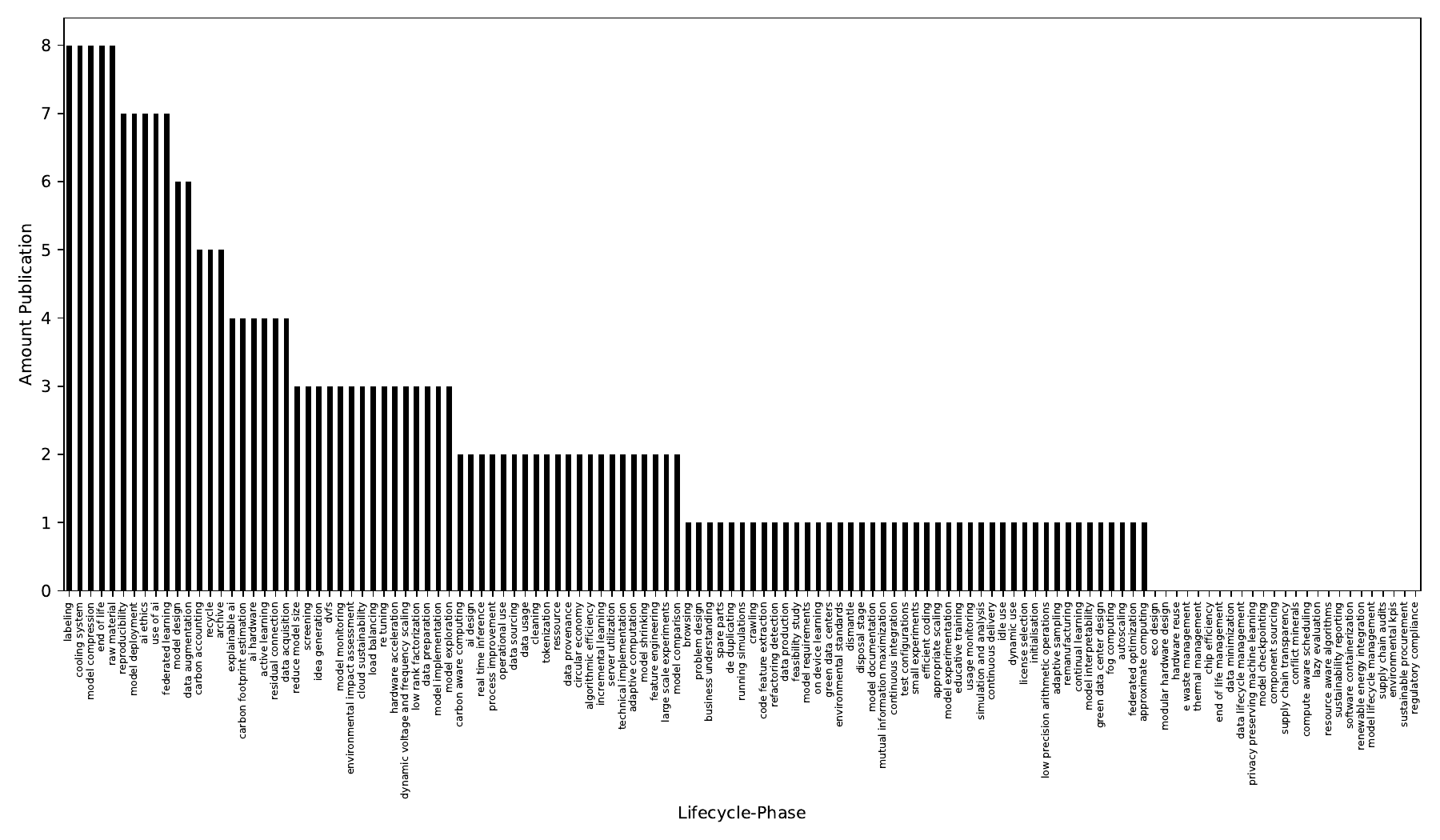}
	\caption{Bar chart of lifecycle phases - \textbf{lower panel} showing terms occurring less than nine times across the corpus. These low-frequency terms represent dominant lifecycle stages such as \emph{model training}, \emph{deployment}, \emph{inference}, and \emph{power consumption}, which collectively account for the majority of observed salience. Their prominence reflects a research focus on compute-intensive phases, as quantified in section~\ref{subsubsec:StateoftheArtGreenAILifecycleStages}.}
	\label{fig:BarchartLifeCyclePhasesUnder9}
\end{figure}

\clearpage

\subsection{Heatmap of the 34 most frequent lifecycle-related terms across the reviewed articles.}
\begin{table}[htbp]
	\centering
	\caption{Heatmap of the 34 most frequent lifecycle-related terms across the reviewed articles. The heatmap ranges from blue-white-red. A filled cell indicates that the term (column) is explicitly addressed in the corresponding article (row). Terms are ordered according to the dendrogram leaf arrangement (Fig.~\ref{fig:DendrogramAILifecycle}) to ensure comparability across figures. This Table~\ref{tab:Heatmap} forms the basis for aggregating terms into harmonized subphases and high-level phases in the phase-level analysis (Table~\ref{tab:Lifecycle}).}
	\scalebox{0.52}{
		\begin{tabular}{@{} |l | *{34}{>{\centering\arraybackslash}m{0.29cm}} | m{0.45cm} | @{}}
			\toprule
			\textbf{Synthesis} & \multicolumn{34}{c|}{\textbf{Terms}} & \\
			\cmidrule(lr){1-1}\cmidrule(l){2-35}\cmidrule(lr){36-36}
			Source & \rotatebox{90}{development} 
			& \rotatebox{90}{inference} 
			& \rotatebox{90}{production} 
			& \rotatebox{90}{distribution} 
			& \rotatebox{90}{power consumption} 
			& \rotatebox{90}{deployment} 
			& \rotatebox{90}{model training} 
			& \rotatebox{90}{ai model} 
			& \rotatebox{90}{manufacturing} 
			& \rotatebox{90}{efficiency} 
			& \rotatebox{90}{transparency} 
			& \rotatebox{90}{transport} 
			& \rotatebox{90}{privacy} 
			& \rotatebox{90}{accountability} 
			& \rotatebox{90}{reuse} 
			& \rotatebox{90}{benchmarking} 
			& \rotatebox{90}{pruning} 
			& \rotatebox{90}{architecture search} 
			& \rotatebox{90}{energy efficiency} 
			& \rotatebox{90}{neural architecture search} 
			& \rotatebox{90}{quantization} 
			& \rotatebox{90}{sparsity} 
			& \rotatebox{90}{transfer learning} 
			& \rotatebox{90}{data collection} 
			& \rotatebox{90}{hyperparameter tuning} 
			& \rotatebox{90}{documentation} 
			& \rotatebox{90}{extraction} 
			& \rotatebox{90}{lifecycle assessment} 
			& \rotatebox{90}{lighting} 
			& \rotatebox{90}{maintenance} 
			& \rotatebox{90}{refine} 
			& \rotatebox{90}{model distillation} 
			& \rotatebox{90}{fine tuning} 
			& \rotatebox{90}{disposal} & $\sum$ \\
			\midrule
			Gossart \citeyear{hilty_rebound_2015} \cite{hilty_rebound_2015}  & \checkmark &  & \checkmark &  & \checkmark & \checkmark &  &  & \checkmark & \checkmark &  & \checkmark & \checkmark &  &  &  &  &  & \checkmark &  &  &  &  &  &  &  &  & \checkmark & \checkmark &  &  &  &  & \checkmark & \cellcolor[HTML]{9595FF}12 \\
			Batra et~al. \citeyear{batra_artificial-intelligence_2018} \cite{batra_artificial-intelligence_2018} & \checkmark & \checkmark &  &  & \checkmark &  &  & \checkmark &  & \checkmark &  &  &  &  &  &  &  &  &  &  &  &  &  &  &  &  &  &  &  &  & \checkmark &  &  &  & \cellcolor[HTML]{9595FF}6 \\
			Wu et~al. \citeyear{wu_machine_2019} \cite{wu_machine_2019} & \checkmark & \checkmark & \checkmark & \checkmark & \checkmark & \checkmark &  &  &  &  &  &  &  &  & \checkmark & \checkmark & \checkmark & \checkmark &  & \checkmark & \checkmark & \checkmark &  &  &  & \checkmark &  &  &  &  &  &  &  &  & \cellcolor[HTML]{B8B8FF}14 \\
			Lacoste et~al. \citeyear{lacoste_quantifying_2019} \cite{lacoste_quantifying_2019} &  & \checkmark & \checkmark & \checkmark & \checkmark &  & \checkmark &  &  & \checkmark & \checkmark &  & \checkmark &  &  &  &  &  &  &  &  &  &  &  & \checkmark &  &  &  &  &  &  &  &  &  &  \cellcolor[HTML]{9595FF}9 \\
			Schwartz et~al. \citeyear{schwartz_green_2020} \cite{schwartz_green_2020} & \checkmark & \checkmark & \checkmark &  &  &  &  & \checkmark &  & \checkmark & \checkmark &  &  &  & \checkmark &  & \checkmark & \checkmark &  & \checkmark &  &  & \checkmark &  & \checkmark &  &  &  & \checkmark &  &  &  & \checkmark &  & \cellcolor[HTML]{B8B8FF}14 \\
			Henderson et~al. \citeyear{henderson_towards_2020} \cite{henderson_towards_2020} & \checkmark & \checkmark & \checkmark & \checkmark & \checkmark & \checkmark & \checkmark &  & \checkmark &  & \checkmark & \checkmark &  & \checkmark &  & \checkmark &  &  &  &  &  &  &  &  &  & \checkmark &  & \checkmark & \checkmark &  &  &  &  & \checkmark & \cellcolor[HTML]{B8B8FF}14 \\
			Anthony et~al. \citeyear{anthony_carbontracker_2020} \cite{anthony_carbontracker_2020} & \checkmark & \checkmark & \checkmark &  & \checkmark &  & \checkmark &  & \checkmark & \checkmark & \checkmark &  &  &  &  & \checkmark & \checkmark &  &  &  & \checkmark &  &  &  & \checkmark &  & \checkmark &  &  &  &  &  &  &  & \cellcolor[HTML]{A3A3FF}13 \\
			Tr\'ebaol \citeyear{trebaol_ecole_2020} \cite{trebaol_ecole_2020} & \checkmark &  & \checkmark & \checkmark & \checkmark & \checkmark &  &  & \checkmark &  & \checkmark & \checkmark & \checkmark &  &  & \checkmark &  &  &  &  &  &  &  &  &  & \checkmark &  &  &  &  &  &  &  &  & \cellcolor[HTML]{9595FF}11 \\
			Strubell et~al. \citeyear{strubell_energy_2020} \cite{strubell_energy_2020} & \checkmark & \checkmark &  & \checkmark & \checkmark &  &  & \checkmark &  &  &  &  &  &  &  &  &  & \checkmark &  & \checkmark &  &  &  &  & \checkmark &  &  &  &  &  &  &  &  &  & \cellcolor[HTML]{9595FF}8 \\
			Hernandez and Brown \citeyear{hernandez_measuring_2020} \cite{hernandez_measuring_2020} &  & \checkmark &  &  &  &  &  & \checkmark &  &  &  &  &  &  &  &  &  & \checkmark &  &  &  & \checkmark & \checkmark &  &  &  &  &  &  &  &  &  &  &  & \cellcolor[HTML]{9595FF}5 \\
			Patterson et~al. \citeyear{patterson_carbon_2021} \cite{patterson_carbon_2021} & \checkmark & \checkmark & \checkmark & \checkmark & \checkmark &  & \checkmark &  &  & \checkmark &  &  &  &  & \checkmark & \checkmark & \checkmark & \checkmark & \checkmark & \checkmark & \checkmark & \checkmark & \checkmark &  &  &  &  &  &  &  & \checkmark & \checkmark & \checkmark &  & \cellcolor[HTML]{FFE0E0}19 \\
			Xu et~al. \citeyear{xu_survey_2021} \cite{xu_survey_2021} & \checkmark & \checkmark & \checkmark & \checkmark & \checkmark & \checkmark & \checkmark & \checkmark &  &  &  &  & \checkmark &  & \checkmark &  & \checkmark & \checkmark &  & \checkmark & \checkmark & \checkmark & \checkmark &  &  &  & \checkmark &  &  &  &  & \checkmark & \checkmark &  & \cellcolor[HTML]{FFE0E0}19 \\
			Haakman et~al. \citeyear{haakman_ai_2021} \cite{haakman_ai_2021} & \checkmark & \checkmark & \checkmark & \checkmark &  & \checkmark & \checkmark & \checkmark &  &  & \checkmark &  & \checkmark & \checkmark & \checkmark &  &  &  &  &  &  &  &  & \checkmark &  & \checkmark &  &  &  &  & \checkmark &  &  & \checkmark & \cellcolor[HTML]{CCCCFF}15 \\
			Li et~al. \citeyear{li_full-cycle_2021} \cite{li_full-cycle_2021} & \checkmark & \checkmark &  & \checkmark & \checkmark & \checkmark & \checkmark & \checkmark &  &  &  &  &  &  &  & \checkmark & \checkmark & \checkmark &  & \checkmark & \checkmark &  & \checkmark &  &  &  &  &  &  &  &  & \checkmark &  &  & \cellcolor[HTML]{B8B8FF}14 \\
			Lannelongue et~al. \citeyear{lannelongue_green_2021} \cite{lannelongue_green_2021} & \checkmark &  & \checkmark & \checkmark & \checkmark &  & \checkmark &  & \checkmark &  &  & \checkmark &  &  & \checkmark &  &  &  &  &  &  &  &  &  &  &  &  & \checkmark & \checkmark & \checkmark &  &  &  &  & \cellcolor[HTML]{9595FF}11 \\
			Van Wynsberghe \citeyear{van_wynsberghe_sustainable_2021} \cite{van_wynsberghe_sustainable_2021} & \checkmark &  & \checkmark & \checkmark &  &  & \checkmark & \checkmark & \checkmark &  &  &  & \checkmark & \checkmark &  &  &  & \checkmark &  & \checkmark &  &  &  &  &  &  &  &  &  &  &  &  &  &  & \cellcolor[HTML]{9595FF}10 \\
			Bannour et~al. \citeyear{bannour_evaluating_2021} \cite{bannour_evaluating_2021} & \checkmark & \checkmark & \checkmark & \checkmark &  & \checkmark &  &  & \checkmark &  &  &  &  & \checkmark &  & \checkmark &  &  &  &  &  &  &  &  &  & \checkmark & \checkmark &  &  &  &  &  &  &  & \cellcolor[HTML]{9595FF}10 \\
			Cao et~al. \citeyear{cao_irene_2021} \cite{cao_irene_2021} &  & \checkmark &  & \checkmark & \checkmark & \checkmark &  &  &  &  &  &  &  &  &  &  &  &  &  &  &  & \checkmark &  &  & \checkmark &  &  &  & \checkmark &  &  &  &  &  & \cellcolor[HTML]{9595FF}7 \\
			Shaikh et~al. \citeyear{shaikh_energyvis_2021} \cite{shaikh_energyvis_2021} &  &  & \checkmark &  &  & \checkmark & \checkmark &  &  &  & \checkmark &  &  &  &  &  &  & \checkmark &  & \checkmark &  &  &  &  &  &  &  &  & \checkmark &  &  &  &  &  & \cellcolor[HTML]{9595FF}7 \\
			Wu et~al. \citeyear{wu_sustainable_2022} \cite{wu_sustainable_2022} & \checkmark & \checkmark & \checkmark & \checkmark & \checkmark & \checkmark & \checkmark & \checkmark & \checkmark & \checkmark & \checkmark & \checkmark & \checkmark & \checkmark &  & \checkmark & \checkmark & \checkmark &  & \checkmark & \checkmark & \checkmark &  & \checkmark & \checkmark &  &  &  &  &  &  &  &  &  & \cellcolor[HTML]{FFA3A3}22 \\
			Dodge et~al. \citeyear{dodge_measuring_2022} \cite{dodge_measuring_2022} & \checkmark & \checkmark & \checkmark &  & \checkmark & \checkmark & \checkmark & \checkmark & \checkmark & \checkmark & \checkmark & \checkmark &  & \checkmark &  &  &  & \checkmark & \checkmark & \checkmark &  &  & \checkmark &  & \checkmark &  & \checkmark &  &  & \checkmark & \checkmark &  &  & \checkmark & \cellcolor[HTML]{FF9595}21 \\
			Kaack et~al. \citeyear{kaack_aligning_2022} \cite{kaack_aligning_2022} & \checkmark & \checkmark & \checkmark &  &  & \checkmark & \checkmark &  & \checkmark & \checkmark & \checkmark & \checkmark & \checkmark & \checkmark & \checkmark &  &  &  & \checkmark &  &  &  & \checkmark & \checkmark & \checkmark &  & \checkmark & \checkmark &  & \checkmark &  &  & \checkmark &  & \cellcolor[HTML]{FF8787}20 \\
			Patterson et~al. \citeyear{patterson_carbon_2022} \cite{patterson_carbon_2022} & \checkmark & \checkmark & \checkmark &  & \checkmark & \checkmark & \checkmark &  & \checkmark & \checkmark & \checkmark &  &  & \checkmark & \checkmark &  &  & \checkmark & \checkmark & \checkmark &  & \checkmark &  &  &  &  &  &  &  &  &  &  &  &  & \cellcolor[HTML]{CCCCFF}15 \\
			Georgiou et~al. \citeyear{georgiou_green_2022} \cite{georgiou_green_2022} & \checkmark & \checkmark &  & \checkmark & \checkmark &  & \checkmark &  & \checkmark & \checkmark &  & \checkmark &  &  &  &  & \checkmark &  & \checkmark &  &  &  &  &  &  & \checkmark &  &  &  & \checkmark &  &  &  &  & \cellcolor[HTML]{9595FF}12 \\
			Ligozat et~al. \citeyear{ligozat_unraveling_2022} \cite{ligozat_unraveling_2022} & \checkmark & \checkmark & \checkmark & \checkmark &  & \checkmark &  & \checkmark & \checkmark &  &  & \checkmark &  &  & \checkmark &  &  &  &  &  &  & \checkmark &  &  &  &  & \checkmark & \checkmark &  &  &  &  &  &  & \cellcolor[HTML]{9595FF}12 \\
			Robbins et al. \citeyear{robbins_our_2022} \cite{robbins_our_2022} & \checkmark &  & \checkmark &  & \checkmark &  &  & \checkmark & \checkmark &  & \checkmark & \checkmark & \checkmark &  &  &  &  &  &  &  &  &  &  & \checkmark &  &  &  &  &  &  &  &  &  & \checkmark & \cellcolor[HTML]{9595FF}10 \\
			Gupta et~al. \citeyear{gupta_act_2022} \cite{gupta_act_2022} & \checkmark & \checkmark & \checkmark &  & \checkmark &  &  &  & \checkmark & \checkmark &  & \checkmark &  &  & \checkmark &  &  &  & \checkmark &  &  &  &  &  &  &  &  & \checkmark &  &  &  &  &  &  & \cellcolor[HTML]{9595FF}10 \\
			Zhou et~al. \citeyear{zhou_opportunities_2023} \cite{zhou_opportunities_2023} & \checkmark & \checkmark & \checkmark & \checkmark & \checkmark & \checkmark & \checkmark & \checkmark &  & \checkmark &  & \checkmark & \checkmark &  & \checkmark &  & \checkmark & \checkmark & \checkmark & \checkmark & \checkmark & \checkmark & \checkmark & \checkmark & \checkmark &  & \checkmark &  & \checkmark & \checkmark & \checkmark & \checkmark & \checkmark & \checkmark & \cellcolor[HTML]{FF7474}30 \\
			Verdecchia et~al. \citeyear{verdecchia_systematic_2023} \cite{verdecchia_systematic_2023} & \checkmark & \checkmark & \checkmark & \checkmark & \checkmark & \checkmark & \checkmark & \checkmark &  & \checkmark & \checkmark &  & \checkmark & \checkmark & \checkmark & \checkmark & \checkmark &  & \checkmark &  & \checkmark &  &  & \checkmark & \checkmark & \checkmark & \checkmark & \checkmark & \checkmark &  & \checkmark & \checkmark &  &  & \cellcolor[HTML]{FFBABA}25 \\
			Menghani \citeyear{menghani_efficient_2023} \cite{menghani_efficient_2023} & \checkmark & \checkmark & \checkmark & \checkmark & \checkmark & \checkmark & \checkmark &  &  & \checkmark &  &  & \checkmark &  & \checkmark & \checkmark & \checkmark & \checkmark &  & \checkmark & \checkmark & \checkmark &  &  & \checkmark & \checkmark &  &  &  &  & \checkmark & \checkmark & \checkmark & \checkmark & \cellcolor[HTML]{FFA3A3}22 \\
			Chen et~al. \citeyear{chen_survey_2023} \cite{chen_survey_2023} & \checkmark & \checkmark &  & \checkmark &  & \checkmark & \checkmark & \checkmark &  &  & \checkmark & \checkmark & \checkmark & \checkmark & \checkmark &  & \checkmark &  &  &  & \checkmark & \checkmark & \checkmark & \checkmark &  &  &  &  & \checkmark &  &  & \checkmark &  &  & \cellcolor[HTML]{FFDBDB}18 \\
			Eilam et~al. \citeyear{eilam_towards_2023} \cite{eilam_towards_2023} & \checkmark & \checkmark &  & \checkmark & \checkmark & \checkmark & \checkmark & \checkmark & \checkmark & \checkmark &  & \checkmark &  &  &  &  &  &  & \checkmark &  &  &  & \checkmark & \checkmark &  &  & \checkmark & \checkmark &  & \checkmark &  & \checkmark & \checkmark &  & \cellcolor[HTML]{FFDBDB}18 \\
			Luccioni et~al. \citeyear{luccioni_estimating_2023} \cite{luccioni_estimating_2023} & \checkmark & \checkmark & \checkmark & \checkmark & \checkmark & \checkmark & \checkmark &  & \checkmark &  & \checkmark & \checkmark &  &  &  & \checkmark &  &  &  &  &  &  &  &  &  & \checkmark & \checkmark & \checkmark &  & \checkmark &  &  &  & \checkmark & \cellcolor[HTML]{F2F2FF}16 \\
			Tornede et~al. \citeyear{tornede_towards_2023} \cite{tornede_towards_2023} & \checkmark & \checkmark & \checkmark & \checkmark & \checkmark &  &  &  &  &  & \checkmark &  &  & \checkmark &  & \checkmark &  & \checkmark &  & \checkmark &  &  & \checkmark &  & \checkmark &  &  &  &  & \checkmark &  &  &  &  & \cellcolor[HTML]{A3A3FF}13 \\
			Li et~al. \citeyear{li_making_2023} \cite{li_making_2023} & \checkmark & \checkmark &  & \checkmark & \checkmark &  & \checkmark & \checkmark & \checkmark & \checkmark & \checkmark &  & \checkmark &  & \checkmark &  & \checkmark &  & \checkmark &  &  &  &  &  &  &  &  &  &  &  &  &  &  &  & \cellcolor[HTML]{A3A3FF}13 \\
			Bouza et~al. \citeyear{bouza_how_2023} \cite{bouza_how_2023} & \checkmark &  & \checkmark & \checkmark & \checkmark & \checkmark &  & \checkmark & \checkmark &  & \checkmark & \checkmark &  & \checkmark &  &  &  &  &  &  &  &  &  & \checkmark &  & \checkmark &  & \checkmark &  &  &  &  &  &  & \cellcolor[HTML]{A3A3FF}13 \\
			Grum et~al. \citeyear{masrour_ai_2023} \cite{masrour_ai_2023} & \checkmark &  & \checkmark & \checkmark &  &  &  &  &  &  &  & \checkmark &  &  & \checkmark &  &  &  &  &  &  &  &  & \checkmark &  &  &  &  &  &  &  &  &  &  &  \cellcolor[HTML]{9595FF}6 \\
			Alzoubi and Mishra \citeyear{alzoubi_green_2024} \cite{alzoubi_green_2024} & \checkmark & \checkmark & \checkmark & \checkmark & \checkmark & \checkmark & \checkmark & \checkmark & \checkmark & \checkmark & \checkmark & \checkmark & \checkmark & \checkmark &  & \checkmark & \checkmark &  & \checkmark &  & \checkmark & \checkmark &  & \checkmark &  & \checkmark & \checkmark & \checkmark & \checkmark & \checkmark & \checkmark & \checkmark & \checkmark & \checkmark & \cellcolor[HTML]{FF7575}29 \\
			Bol\'on Canedo et~al. \citeyear{bolon-canedo_review_2024} \cite{bolon-canedo_review_2024} & \checkmark & \checkmark & \checkmark & \checkmark & \checkmark & \checkmark &  & \checkmark & \checkmark & \checkmark & \checkmark & \checkmark & \checkmark & \checkmark &  &  & \checkmark &  & \checkmark &  & \checkmark & \checkmark &  & \checkmark & \checkmark &  &  &  & \checkmark & \checkmark &  & \checkmark & \checkmark &  & \cellcolor[HTML]{FF8D8D}23 \\
			Barbierato and Gatti \citeyear{barbierato_toward_2024} \cite{barbierato_toward_2024} & \checkmark & \checkmark & \checkmark & \checkmark & \checkmark & \checkmark & \checkmark & \checkmark &  & \checkmark &  & \checkmark &  &  &  & \checkmark & \checkmark &  & \checkmark &  & \checkmark & \checkmark & \checkmark &  &  &  & \checkmark &  & \checkmark &  & \checkmark & \checkmark & \checkmark & \checkmark & \cellcolor[HTML]{FFA3A3}22 \\
			Longpre et~al. \citeyear{longpre_responsible_2024} \cite{longpre_responsible_2024} & \checkmark & \checkmark &  & \checkmark &  & \checkmark & \checkmark & \checkmark & \checkmark & \checkmark & \checkmark &  & \checkmark & \checkmark & \checkmark & \checkmark &  &  &  &  &  &  & \checkmark & \checkmark &  & \checkmark &  &  &  & \checkmark & \checkmark &  & \checkmark &  & \cellcolor[HTML]{FFE0E0}19 \\
			Grum and Gronau \citeyear{grum_meaningfulness_2024} \cite{grum_meaningfulness_2024} & \checkmark & \checkmark & \checkmark & \checkmark &  &  &  & \checkmark & \checkmark &  &  &  &  & \checkmark &  & \checkmark &  &  &  &  &  &  & \checkmark & \checkmark &  & \checkmark &  &  &  & \checkmark & \checkmark &  &  &  & \cellcolor[HTML]{A3A3FF}13 \\
			Wright et~al. \citeyear{wright_efficiency_2025} \cite{wright_efficiency_2025} & \checkmark & \checkmark & \checkmark & \checkmark &  & \checkmark & \checkmark & \checkmark & \checkmark & \checkmark & \checkmark & \checkmark & \checkmark & \checkmark &  & \checkmark &  & \checkmark & \checkmark & \checkmark & \checkmark &  &  &  &  &  &  & \checkmark &  &  &  &  &  & \checkmark & \cellcolor[HTML]{FF8787}20 \\
			\midrule
			$\sum$ 
			& \cellcolor[HTML]{FF7474} 39 
			& \cellcolor[HTML]{FF7474} 35 
			& \cellcolor[HTML]{FF7474} 33 
			& \cellcolor[HTML]{FF8080} 32 
			& \cellcolor[HTML]{FFA4A4} 30 
			& \cellcolor[HTML]{FFDBDB} 27 
			& \cellcolor[HTML]{FFEDED} 26 
			& \cellcolor[HTML]{EDEDFF} 24 
			& \cellcolor[HTML]{EDEDFF} 24 
			& \cellcolor[HTML]{C8C8FF} 22 
			& \cellcolor[HTML]{C8C8FF} 22 
			& \cellcolor[HTML]{B6B6FF} 21 
			& \cellcolor[HTML]{9595FF} 18 
			& \cellcolor[HTML]{9595FF} 17 
			& \cellcolor[HTML]{9595FF} 17 
			& \cellcolor[HTML]{9595FF} 17 
			& \cellcolor[HTML]{9595FF} 16 
			& \cellcolor[HTML]{9595FF} 16 
			& \cellcolor[HTML]{9595FF} 15 
			& \cellcolor[HTML]{9595FF} 15 
			& \cellcolor[HTML]{9595FF} 14 
			& \cellcolor[HTML]{9595FF} 14 
			& \cellcolor[HTML]{9595FF} 14 
			& \cellcolor[HTML]{9595FF} 14 
			& \cellcolor[HTML]{9595FF} 13 
			& \cellcolor[HTML]{9595FF} 13 
			& \cellcolor[HTML]{9595FF} 12 
			& \cellcolor[HTML]{9595FF} 12 
			& \cellcolor[HTML]{9595FF} 12 
			& \cellcolor[HTML]{9595FF} 12 
			& \cellcolor[HTML]{9595FF} 11 
			& \cellcolor[HTML]{9595FF} 11 
			& \cellcolor[HTML]{9595FF} 11 
			& \cellcolor[HTML]{9595FF} 11 
			& 640\\
			\bottomrule
		\end{tabular}
	}
	\label{tab:Heatmap}
\end{table}

\clearpage

\subsection{NMDL micro-example}
\begin{figure}[ht]
	\centering
	\includegraphics[width=0.6\textheight, angle=0]{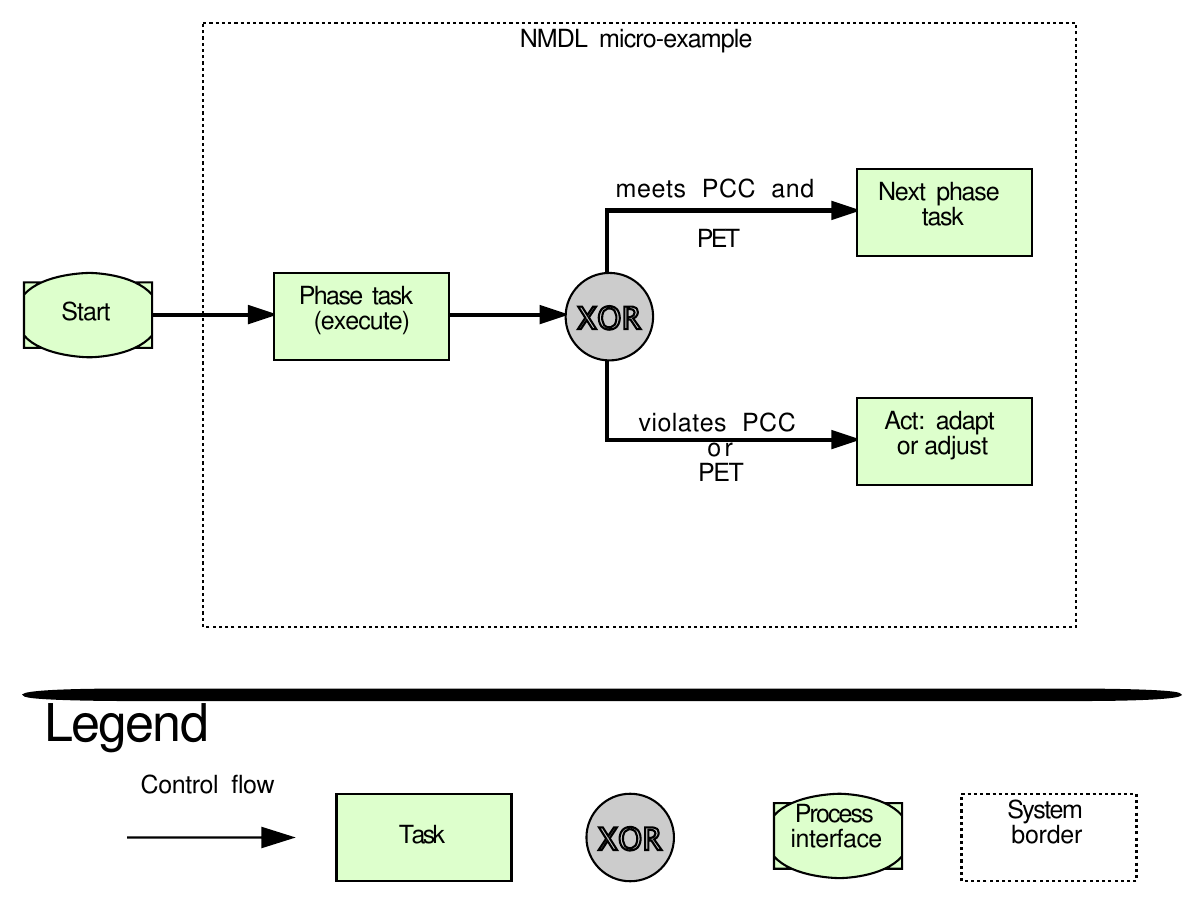}
	\caption{NMDL micro-example: a task is followed by an XOR gateway that evaluates PCC first and PET second. If both are satisfied, flow continues; otherwise an improvement action (\emph{adapt} or \emph{adjust}) is taken before re-checking.}
	\label{fig:NMDLBasics}
\end{figure}

\clearpage

\subsection{Green AI lifecycle - Process models}

\subsubsection{Green AI lifecycle - Realize green hardware selection and infrastructure design}

\begin{figure}[ht]
	\centering
	\includegraphics[width=0.75\textheight,angle=90]{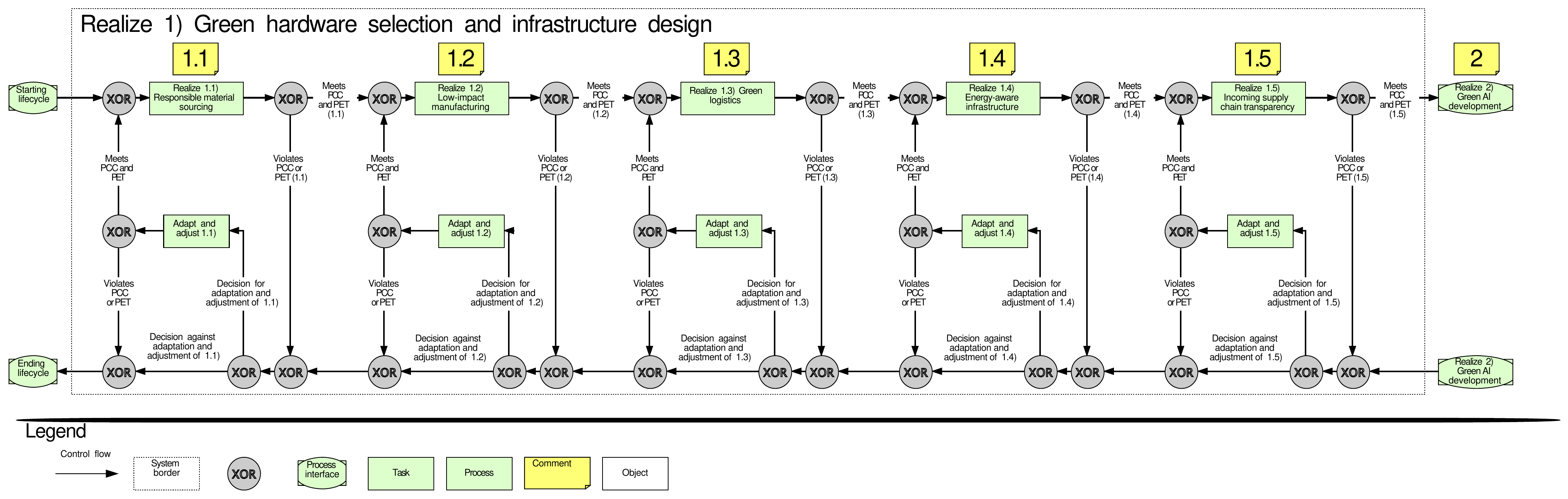}
	\caption{Realize green hardware selection and infrastructure design.}
	\label{fig:RealizeGreenHardwareSelectionAndInfrastructureDesign}
\end{figure}

\clearpage

\subsubsection{Green AI lifecycle - Realize Green AI development part 1}

\begin{figure}[ht]
	\centering
	\includegraphics[width=0.75\textheight, angle=90]{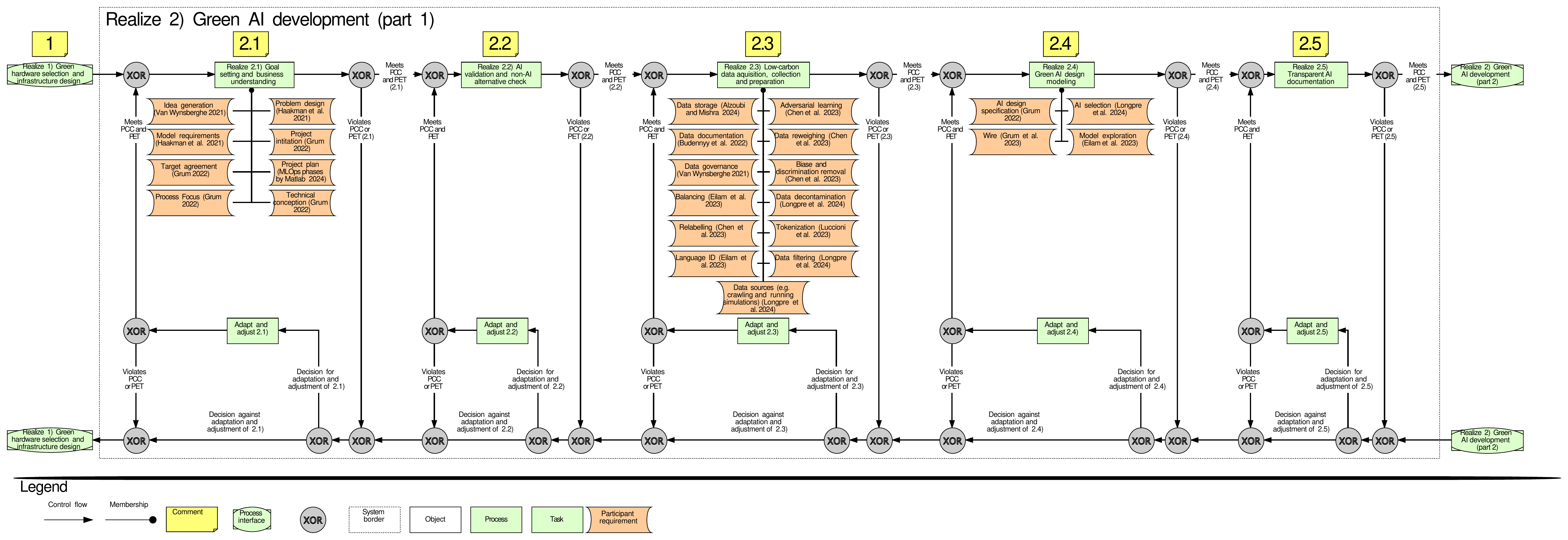}
	\caption{Realize Green AI development part 1.}
	\label{fig:RealizeGreenAIDevelopment1}
\end{figure}

\clearpage

\subsubsection{Green AI lifecycle - Realize Green AI development part 2}

\begin{figure}[ht]
	\centering
	\includegraphics[width=0.75\textheight, angle=90]{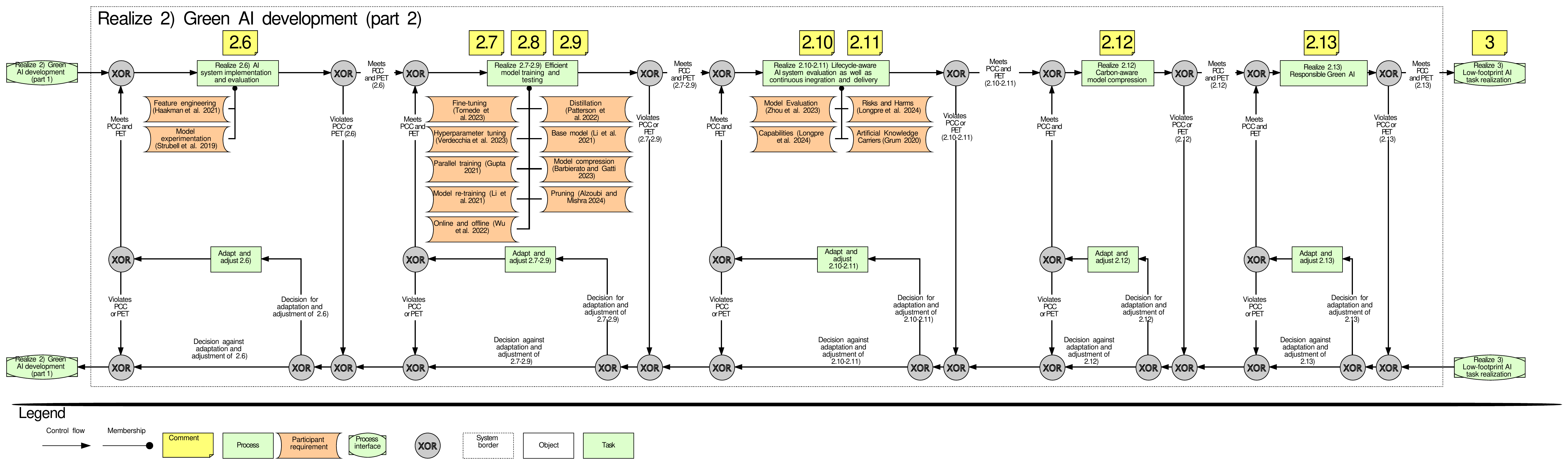}
	\caption{Realize Green AI development part 2.}
	\label{fig:RealizeGreenAIDevelopment2}
\end{figure}

\clearpage

\subsubsection{Green AI lifecycle - Realize low-footprint AI task realization}

\begin{figure}[ht]
	\centering
	\includegraphics[width=0.75\textheight, angle=90]{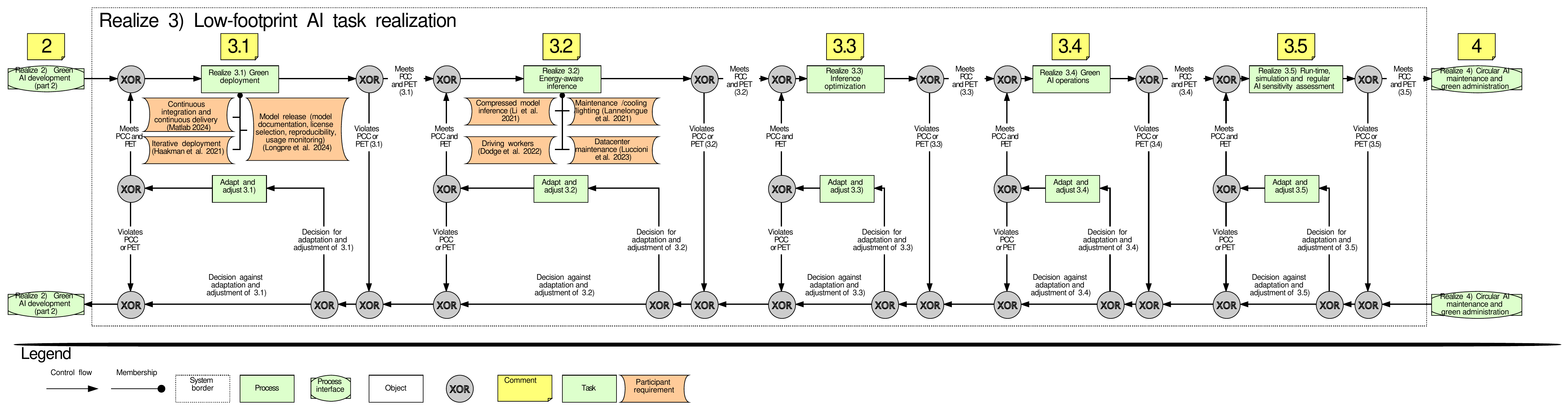}
	\caption{Realize low-footprint AI task realization.}
	\label{fig:RealizeLowFootprintAITaskRealization}
\end{figure}

\clearpage

\subsubsection{Green AI lifecycle - Realize circular AI maintenance and green administration}

\begin{figure}[ht]
	\centering
	\includegraphics[width=0.75\textheight, angle=90]{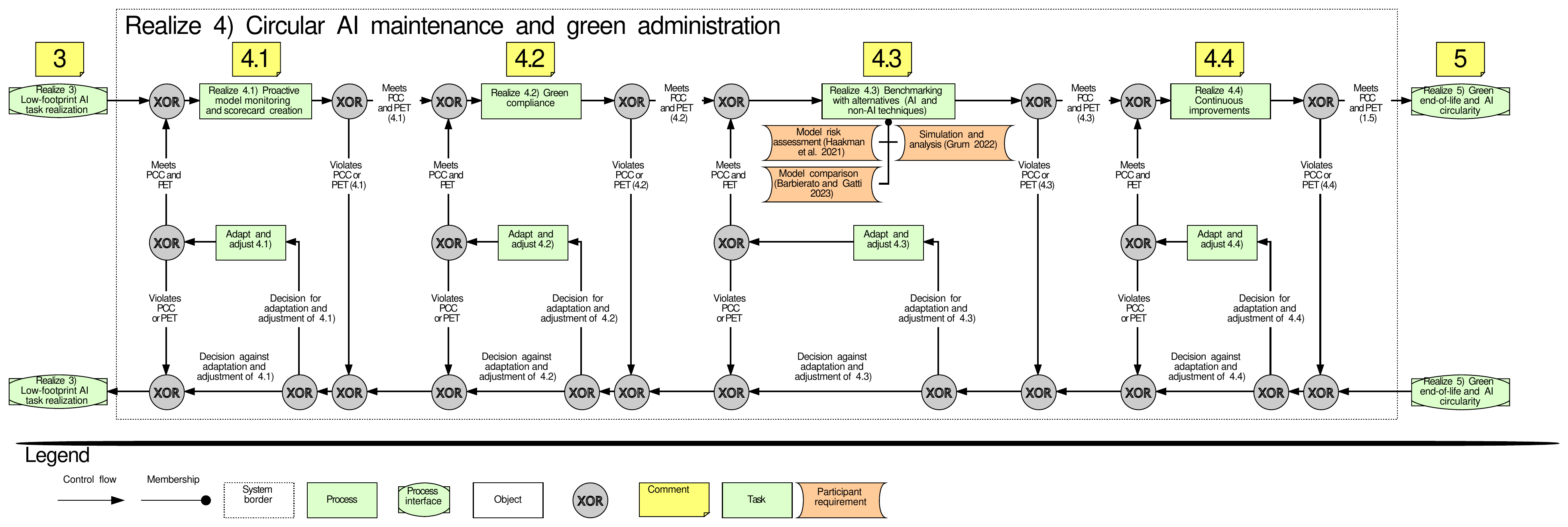}
	\caption{Realize circular AI maintenance and green administration.}
	\label{fig:RealizeCircularAIMaintenanceAndGreenAdministration}
\end{figure}

\clearpage

\subsubsection{Green AI lifecycle - Realize green end-of-life and AI circularity}

\begin{figure}[ht]
	\centering
	\includegraphics[width=0.75\textheight, angle=90]{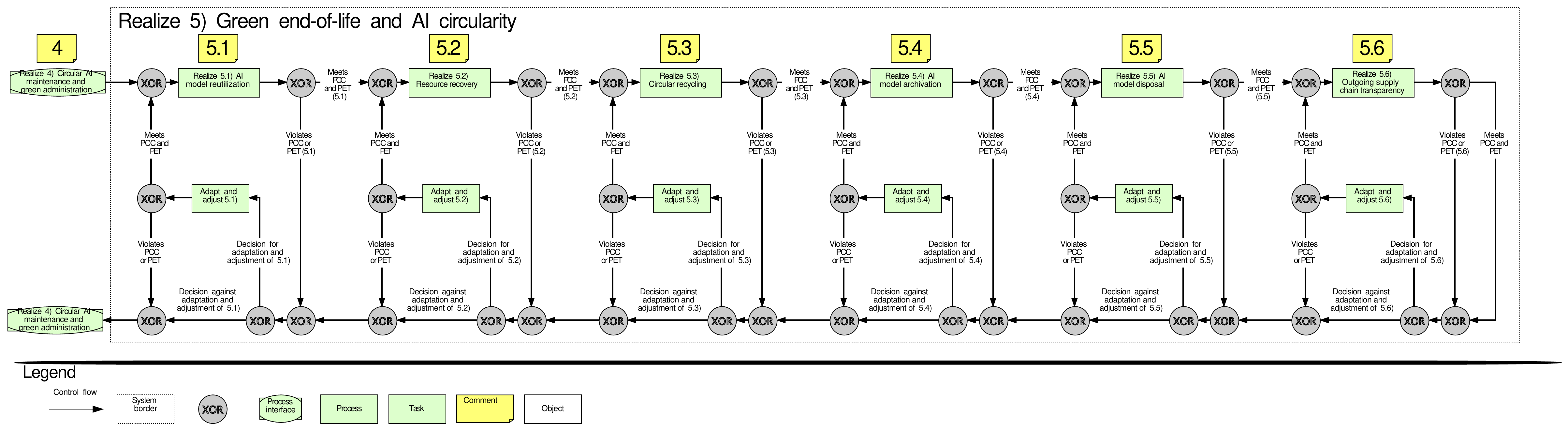}
	\caption{Realize green end-of-life and AI circularity.}
	\label{fig:RealizeGreenEndoflifeAndAICircularity}
\end{figure}

\clearpage

\subsection{Overview of Green AI hardware}
\begin{table}[ht]
	\centering
	\caption{Overview of hardware \emph{instances} related to Green AI (illustrative, non-prescriptive).}
	\scalebox{0.65}{\rotatebox{90}{
			\begin{tabular}{|l|llllllllll|}
				\toprule
				\rotatebox{90}{\shortstack{Source}} & \rotatebox{90}{\shortstack{Core \\ Type}} & \rotatebox{90}{\shortstack{CPU}} & \rotatebox{90}{\shortstack{CPU \\ Cores}} & \rotatebox{90}{\shortstack{RAM}} & \rotatebox{90}{\shortstack{GPU}} & \rotatebox{90}{\shortstack{GPU \\ RAM}} & \rotatebox{90}{\shortstack{GPU \\ Cores}} & \rotatebox{90}{\shortstack{Hardware type}} & \rotatebox{90}{\shortstack{Hardware architecture}} & \rotatebox{90}{\shortstack{Recommended level}} \\
				\midrule\midrule
				\cite{albreem_green_2021} & CPU & Arduino Uno R3 & 1 C & - & - & - & - & Single-board comp. (SBC) & AVR & Local CPS \\
				\cite{albreem_green_2021} & CPU & Arduino Mega 2560 & 1 C & - & - & - & - & Single-board computer & AVR & Local CPS \\
				\cite{albreem_green_2021} & CPU & Banana Pi & 2 C & 1 GB & - & - & - & Single-board computer & ARM & Local CPS \\
				\cite{albreem_green_2021} & CPU & Raspberry Pi 3B & 4 C & 1 GB & - & - & - & Single-board computer & ARM & Local CPS \\
				\cite{bannour_evaluating_2021} & GPU & - & - & - & GeForce GTX 1080 Ti & 11 GB & 3584 & Desktop GPU & x86 & Local CPS \\
				\cite{bannour_evaluating_2021} & GPU & - & - & - & NVIDIA Tesla V100 & 16 GB & 5120 & SBC / AI Accel. & x86 & Local CPS \\
				\cite{zhu_green_2022} & Both & Jetson Xavier NX & 6 C & 8 GB & NVIDIA A100 & 80 GB & 6912 & {SBC/ AI Accelerator} & ARM & Local CPS \\
				\cite{yokoyama_investigating_2023} & Both & Denver 2, 2 GHz & 18 C & 24 GB & Pascal @ 1300 MHz & - & 768 C & CPU & ARM & Local CPS \\
				\cite{yokoyama_investigating_2023} & Both & ARM v8.2 64-bit & 8 C & 32 GB & NVIDIA Volta & - & 512 & Server GPU / AI Accel. & x86 & Local CPS \\
				\cite{yokoyama_investigating_2023} & GPU & i7 8700 @ 3.2 GHz & 6 C & 64 GB & - & - & - & Desktop CPU & x86 & Local CPS \\
				\cite{yokoyama_investigating_2023} & Both & i7 8700 @ 3.2 GHz & 6 C & 64 GB & RTX 2080 Ti, 1545 MHz & 11 GB & 544 & Desktop (CPU and GPU) & x86 & Local CPS \\
				\cite{yokoyama_investigating_2023} & CPU & NVIDIA Jetson Nano & 4 C & 4 GB & 128-Core Maxwell GPU & - & 128 & AI Accelerator & ARM & Local CPS \\
				\cite{yokoyama_investigating_2023} & CPU & NVIDIA Jetson TX2 & 6 C & 8 GB & 256-Core Pascal GPU & - & 256 & AI Accelerator & ARM & Local CPS \\
				\cite{yokoyama_investigating_2023} & CPU & Raspberry Pi 4 & 4 C & 1 GB & - & - & - & Single-board computer & ARM & Local CPS \\
				\cite{sprind_composite_2024} & GPU & - & - & - & RTX 4000 SFF & 20 GB & 6144 & Desktop GPU & x86 & Local CPS \\
				\midrule
				\cite{sprind_composite_2024} & GPU & - & - & - & RTX A6000 & 48 GB & 10752 & Workstation GPU & x86 & Local cloud \\
				\cite{sprind_composite_2024} & CPU & EPYC 7R13 & 64 C & - & - & - & - & Server CPU & x86 & Local cloud \\
				\cite{sprind_composite_2024} & CPU & Intel Xeon 8175 & 28 C & - & - & - & - & Server CPU & x86 & Local cloud \\
				\cite{yokoyama_investigating_2023} & CPU & NVIDIA Jetson AGX Xavier & 8 C & 32 GB & 512-Core Volta GPU & - & 512 & AI Accelerator & ARM & Local cloud \\
				\cite{sprind_composite_2024} & AI TP. & - & - & - & Radeon Instinct Mi250X & 128 GB & 0 & Server GPU / AI Accel. & CDNA2 / x86 & Local cloud \\
				\cite{xu_energy_2023} & GPU & - & - & - & Tesla P100-PCIE-16GB & 16 GB & 3584 & AI Accelerator & x86 & Local cloud \\
				\cite{xu_energy_2023} & Both & E5-2698 v4, 2.2 GHz & 20 C & - & Tesla V100-SXM2-32GB & 32 GB & 5120 & AI Accelerator & x86 & Local cloud \\
				\cite{qiu_first_2023} & Both & Gold 6152, 2.1 GHz & 22 C & - & Tesla V100 PCIe 32 GB & 32 GB & 5120 & Server GPU / AI Accel. and CPU & x86 & Local cloud \\
				\cite{sprind_composite_2024} & GPU & - & - & - & L40S & 48 GB & 18176 & Server GPU / AI Accel. & x86 & Local cloud \\
				\midrule
				\cite{sprind_composite_2024} & AI TP. & - & - & - & Habana Gaudi & 128 GB & 0 & AI Accelerator & x86 & Public cloud \\
				\cite{sprind_composite_2024} & GPU & - & - & - & NVIDIA H100 & 94 GB & 14592 & Server GPU / AI Accel. & x86 & Public cloud \\
				\bottomrule
			\end{tabular} 
	}}
	\label{tab:HardwareOverview}
\end{table}

\clearpage

\subsection{Summary of Green AI measurement tools}
\begin{table}[ht]
	\centering
	\caption{Summary of Green AI measurement tools (fulfilled: \checkmark; not fulfilled: empty) as heatmap. The heatmap ranges from blue-white-red.}
	\scalebox{1}{\rotatebox{90}{
			\centering
			\scriptsize
					\begin{tabular}{|l|lccccccc|ccc|ccc|c|c|c|}
					\toprule
					{Unit of Analysis / Source} & \multicolumn{8}{l|}{ } & \multicolumn{3}{l|}{ } & \multicolumn{3}{l|}{ } & \multicolumn{1}{l|}{ } & \multicolumn{1}{l|}{ } & \multicolumn{1}{l|}{ } \\
					& \multicolumn{8}{l|}{Python Packages} & \multicolumn{3}{l|}{Online Tools} & \multicolumn{3}{l|}{Dashboards} & Cloud & Water & \\
					\cmidrule(lr){2-9}\cmidrule(lr){10-12}\cmidrule(lr){13-15}\cmidrule(lr){16-16}\cmidrule(lr){17-17}\cmidrule(lr){18-18}
					&
					\rotatebox{90}{\shortstack{EOECD.AI EIT \cite{oecdai_experiment_2023}}} &
					\rotatebox{90}{\shortstack{Eco2AI \cite{budennyy_eco2ai_2022}}} &
					\rotatebox{90}{\shortstack{IrEne \cite{cao_irene_2021}}} &
					\rotatebox{90}{\shortstack{CodeCarbon (MLCO\textsubscript{2}) \cite{codecarbon_codecarbon_2021}}} &
					\rotatebox{90}{\shortstack{Carbon Tracker \cite{anthony_carbontracker_2020}}} &
					\rotatebox{90}{\shortstack{Experiment Impact Tracker \cite{henderson_towards_2020}}} &
					\rotatebox{90}{\shortstack{CUMULAOR \cite{trebaol_ecole_2020}}} &
					\rotatebox{90}{\shortstack{Energy Usage \cite{lottick_energy_2019}}} &
					\rotatebox{90}{\shortstack{Green Algorithms \cite{lannelongue_green_2021}}} &
					\rotatebox{90}{\shortstack{ML Emissions Calculator \cite{lacoste_quantifying_2019}}} &
					\rotatebox{90}{\shortstack{EnergyVis \cite{shaikh_energyvis_2021}}} &
					\rotatebox{90}{\shortstack{Carbon Footprint Tool \cite{aws_aws_2022}}} &
					\rotatebox{90}{\shortstack{Carbon Footprint \cite{google_llc_google_2021}}} &
					\rotatebox{90}{\shortstack{Emissions Impact Dashboard \cite{microsoft_microsoft_2020}}} &
					\rotatebox{90}{\shortstack{Cloud Carbon Footprint \cite{thoughtworks_cloud_2021}}} &
					\rotatebox{90}{\shortstack{Water Estimation Tool \cite{li_making_2023}}} & $\sum$ \\
					\midrule
					Lottick et~al. \citeyear{lottick_energy_2019} \cite{lottick_energy_2019} &   &   &   &   &   &   &   & \checkmark &   &   &   &   &   &   &   &   & \cellcolor[HTML]{9595FF}1\\
					Lacoste et~al. \citeyear{lacoste_quantifying_2019} \cite{lacoste_quantifying_2019} &   &   &   &   &   &   &   &   &   & \checkmark &   &   &   &   &   &   & \cellcolor[HTML]{9595FF}1\\
					Anthony et~al. \citeyear{anthony_carbontracker_2020} \cite{anthony_carbontracker_2020} &   &   &   &   & \checkmark & \checkmark &   &   &   & \checkmark &   &   &   &   &   &   & \cellcolor[HTML]{9595FF}3\\
					Tr\'ebaol \citeyear{trebaol_ecole_2020} \cite{trebaol_ecole_2020} &   &   &   &   & \checkmark & \checkmark &   &   & \checkmark &   &   &   &   &   &   &   & \cellcolor[HTML]{9595FF}3\\
					Henderson et~al. \citeyear{henderson_towards_2020} \cite{henderson_towards_2020} &   &   &   &   &   & \checkmark &   &   &   &   &   &   &   &   &   &   & \cellcolor[HTML]{9595FF}1\\
					Bannour et~al. \citeyear{bannour_evaluating_2021} \cite{bannour_evaluating_2021}  &   &   &   & \checkmark & \checkmark & \checkmark & \checkmark & \checkmark & \checkmark & \checkmark &   &   &   &   &   &   & \cellcolor[HTML]{FF7474}7\\
					Patterson et~al. \citeyear{patterson_carbon_2021} \cite{patterson_carbon_2021} &   &   &   & \checkmark &   & \checkmark &   &   & \checkmark & \checkmark &   &   &   &   &   &   & \cellcolor[HTML]{DBDBFF}4\\
					Shaikh et~al. \citeyear{shaikh_energyvis_2021} \cite{shaikh_energyvis_2021} &   &   &   & \checkmark & \checkmark &   &   &   &   &   & \checkmark &   &   &   &   &   & \cellcolor[HTML]{9595FF}3\\
					Van Wynsberghe \citeyear{van_wynsberghe_sustainable_2021} \cite{van_wynsberghe_sustainable_2021} &   &   &   &   & \checkmark & \checkmark &   &   &   & \checkmark &   &   &   &   &   &   & \cellcolor[HTML]{9595FF}3\\
					Cao et~al. \citeyear{cao_irene_2021} \cite{cao_irene_2021} &   &   & \checkmark &   &   & \checkmark &   &   &   &   &   &   &   &   &   &   & \cellcolor[HTML]{9595FF}2\\
					Parcollet and Ravanelli \citeyear{parcollet_energy_2021} \cite{parcollet_energy_2021} &   &   &   &   & \checkmark &   &   &   &   &   &   &   &   &   &   &   & \cellcolor[HTML]{9595FF}1\\
					Lannelongue et~al. \citeyear{lannelongue_green_2021} \cite{lannelongue_green_2021} &   &   &   &   &   &   &   &   & \checkmark &   &   &   &   &   &   &   & \cellcolor[HTML]{9595FF}1\\
					Ligozat et~al. \citeyear{ligozat_unraveling_2022} \cite{ligozat_unraveling_2022} &   &   &   & \checkmark & \checkmark & \checkmark &   &   & \checkmark & \checkmark &   &   &   &   &   &   & \cellcolor[HTML]{FFDBDB}5\\
					Budennyy et~al. \citeyear{budennyy_eco2ai_2022} \cite{budennyy_eco2ai_2022} &   & \checkmark &   & \checkmark & \checkmark & \checkmark &   &   & \checkmark &   &   &   &   &   & \checkmark &   & \cellcolor[HTML]{FF9292}6\\
					Dodge et~al. \citeyear{dodge_measuring_2022} \cite{dodge_measuring_2022} &   &   &   & \checkmark & \checkmark &   &   &   &   &   &   &   &   &   &   &   & \cellcolor[HTML]{9595FF}2\\
					Patterson et~al. \citeyear{patterson_carbon_2022} \cite{patterson_carbon_2022} &   &   &   &   &   &   &   &   &   & \checkmark &   &   &   &   &   &   & \cellcolor[HTML]{9595FF}1\\
					Bouza et~al. \citeyear{bouza_how_2023} \cite{bouza_how_2023} &   & \checkmark &   & \checkmark & \checkmark & \checkmark & \checkmark & \checkmark & \checkmark & \checkmark &   &   &   &   &   &   & \cellcolor[HTML]{FF7474}8\\
					Tornede et~al. \citeyear{tornede_towards_2023} \cite{tornede_towards_2023} &   &   & \checkmark & \checkmark & \checkmark &   &   &   & \checkmark & \checkmark & \checkmark &   &   &   &   &   & \cellcolor[HTML]{FF9292}6\\
					Yokoyama et~al. \citeyear{yokoyama_investigating_2023} \cite{yokoyama_investigating_2023} &   &   &   & \checkmark & \checkmark & \checkmark &   &   & \checkmark & \checkmark &   &   &   &   &   &   & \cellcolor[HTML]{FFDBDB}5\\
					Zhou et~al. \citeyear{zhou_opportunities_2023} \cite{zhou_opportunities_2023} &   &   &   & \checkmark & \checkmark &   &   &   & \checkmark & \checkmark &   &   &   &   &   &   & \cellcolor[HTML]{DBDBFF}4\\
					Mart\'inez-Fern\'andez et~al. \citeyear{martinez-fernandez_towards_2023} \cite{martinez-fernandez_towards_2023} &   &   &   &   &   &   &   &   & \checkmark &   & \checkmark & \checkmark & \checkmark &   &   &   & \cellcolor[HTML]{DBDBFF}4\\
					Luccioni et~al. \citeyear{luccioni_estimating_2023} \cite{luccioni_estimating_2023} &   &   &   & \checkmark & \checkmark &   &   &   &   & \checkmark &   &   &   &   &   &   & \cellcolor[HTML]{9595FF}3\\
					Li et~al. \citeyear{li_making_2023} \cite{li_making_2023} &   &   &   &   &   & \checkmark &   &   &   &   &   &   &   &   &   & \checkmark & \cellcolor[HTML]{9595FF}2\\
					Verdecchia et~al. \citeyear{verdecchia_systematic_2023} \cite{verdecchia_systematic_2023} &   &   &   &   & \checkmark &   &   &   &   &   &   &   &   &   &   &   & \cellcolor[HTML]{9595FF}1\\
					Xu et~al. \citeyear{xu_energy_2023} \cite{xu_energy_2023} &   &   &   & \checkmark &   &   &   &   &   &   &   &   &   &   &   &   & \cellcolor[HTML]{9595FF}1\\
					Alzoubi and Mishra \citeyear{alzoubi_green_2024} \cite{alzoubi_green_2024}  & \checkmark & \checkmark &   & \checkmark & \checkmark & \checkmark & \checkmark &   & \checkmark & \checkmark &   &   &   &   &   &   & \cellcolor[HTML]{FF7474}8\\
					Longpre et~al. \citeyear{longpre_responsible_2024} \cite{longpre_responsible_2024} &   &   &   & \checkmark & \checkmark & \checkmark &   &   &   & \checkmark &   & \checkmark & \checkmark & \checkmark &   & \checkmark & \cellcolor[HTML]{FF7474}8\\
					Bol\'on Canedo et~al.. \citeyear{bolon-canedo_review_2024} \cite{bolon-canedo_review_2024} &   &   &   & \checkmark & \checkmark &   &   &   & \checkmark &   &   &   &   &   &   &   & \cellcolor[HTML]{9595FF}3\\
					\midrule
					\textbf{Total per Tool}
					& \cellcolor[HTML]{9595FF} 1 
					& \cellcolor[HTML]{9595FF} 3 
					& \cellcolor[HTML]{9595FF} 2 
					& \cellcolor[HTML]{FF7474} 15 
					& \cellcolor[HTML]{FF7474} 18 
					& \cellcolor[HTML]{FF7878} 14 
					& \cellcolor[HTML]{9595FF} 3 
					& \cellcolor[HTML]{9595FF} 3 
					& \cellcolor[HTML]{FF9696} 13 
					& \cellcolor[HTML]{FF7878} 14 
					& \cellcolor[HTML]{9595FF} 3 
					& \cellcolor[HTML]{9595FF} 2 
					& \cellcolor[HTML]{9595FF} 2 
					& \cellcolor[HTML]{9595FF} 1 
					& \cellcolor[HTML]{9595FF} 1 
					& \cellcolor[HTML]{9595FF} 2
					& \textbf{98}\\
					\bottomrule					
				\end{tabular} 
		}}
		\label{tab:MeasurementTools}
	\end{table}

\clearpage

\end{document}